\pdfoutput=1
\documentclass[journal]{IEEEtai}

\usepackage{array}
\usepackage{graphicx}
\usepackage{balance}
\usepackage{xurl}
\usepackage{multirow}
\usepackage[table,xcdraw,dvipsnames]{xcolor}
\usepackage[normalem]{ulem}
\useunder{\uline}{\ul}{}
\usepackage{rotating}
\usepackage{subcaption}
\usepackage{boldline} 
\usepackage{tcolorbox}
\usepackage{listings}
\usepackage{placeins}
\usepackage{amssymb}
\usepackage{amsmath}
\usepackage[font=small,skip=0pt]{caption}
\usepackage[colorlinks,urlcolor=blue,linkcolor=blue,citecolor=blue]{hyperref}
\hypersetup{
  pdftitle={Efficiency vs. Alignment: Investigating Safety and Fairness Risks in Parameter-Efficient Fine-Tuning of LLMs},
  pdfauthor={Mina Taraghi, Yann Pequignot, Amin Nikanjam, Mohamed Amine Merzouk, and Foutse Khomh}
}
\definecolor{bg}{RGB}{248,248,248}
\definecolor{frame}{RGB}{210,210,210}
\definecolor{kw}{HTML}{1750EB}
\definecolor{str}{HTML}{067D17}
\newcommand{\kw}[1]{\textcolor{kw}{#1}}
\lstdefinestyle{promptstyle}{
  language=Python,
  basicstyle=\ttfamily\small\bfseries,
  keywordstyle=\color{kw}\bfseries,
  stringstyle=\color{str},
  commentstyle=\color{gray!70},
  showstringspaces=false,
  breaklines=true,
  columns=fullflexible,
  keepspaces=true,
  frame=single,
  framerule=0.5pt,
  rulecolor=\color{frame},
  backgroundcolor=\color{bg},
  xleftmargin=6pt,
  xrightmargin=6pt,
  captionpos=b,
  escapeinside={(*@}{@*)}
}

\begin{document}

\title{Efficiency vs. Alignment: Investigating Safety and Fairness Risks in Parameter-Efficient Fine-Tuning of LLMs}

\author{Mina Taraghi, Yann Pequignot, Amin Nikanjam, Mohamed Amine Merzouk, and Foutse Khomh
\thanks{This work was supported by Fonds de Recherche du Qu\'ebec (FRQ), the Canadian Institute for Advanced Research (CIFAR), and the DEEL project CRDPJ 537462-18 funded by the Natural Sciences and Engineering Research Council of Canada (NSERC) and the Consortium for Research and Innovation in Aerospace in Qu\'ebec (CRIAQ), together with its industrial partners Thales Canada Inc., Bell Textron Canada Limited, CAE Inc., and Bombardier Inc.}
\thanks{Mina Taraghi is with Polytechnique Montr\'eal, Montreal, QC H3T 1J4, Canada (e-mail: mina.taraghi@polymtl.ca).}
\thanks{Yann Pequignot is with IID, Laval University, Quebec City, QC G1V 0A6, Canada (e-mail: yann.pequignot@iid.ulaval.ca).}
\thanks{Amin Nikanjam was with Polytechnique Montr\'eal, Montreal, QC H3T 1J4, Canada. He is currently with Huawei Distributed Scheduling and Data Engine Lab, Toronto, Canada (e-mail: amin.nikanjam@h-partners.com).}
\thanks{Mohamed Amine Merzouk was with Polytechnique Montr\'eal, Montreal, QC H3T 1J4, Canada. He is currently with McGill University and Mila--Quebec AI Institute, Montreal, QC H2S 3H1, Canada (e-mail: mohamed-amine.merzouk@mila.quebec).}
\thanks{Foutse Khomh is with Polytechnique Montr\'eal, Montreal, QC H3T 1J4, Canada (e-mail: foutse.khomh@polymtl.ca).}}

\maketitle
\IEEEaftertitletext{\vspace{-1.2\baselineskip}} 

\begin{abstract}
Organizations increasingly adapt Large Language Models (LLMs) from public repositories such as HuggingFace to downstream tasks. Prior work shows that even fine-tuning on benign datasets can weaken safety alignment, raising a practical question: does benign parameter-efficient fine-tuning (PEFT) also affect safety and fairness? We present the first large-scale, systematic study showing that benign PEFT can significantly alter both. We fine-tune four instruction-tuned model families (Meta-Llama-3-8B, Qwen2.5-7B, Mistral-7B, and Gemma-7B) with four widely used PEFT methods: LoRA, IA3, Prompt-Tuning, and P-Tuning. In total, we evaluate 235 conversationally fine-tuned variants across eleven safety hazard categories and nine fairness dimensions. We assess generalization beyond conversational tuning by incorporating a compact extension focused on coding tasks, involving 96 additional fine-tuned models. Results show that benign PEFT can induce detrimental alignment shifts. Adapter-based methods (LoRA, IA3) are generally safer and less disruptive to fairness, whereas prompt-based methods more often reduce safety and worsen fairness accuracy. Base model choice strongly moderates these effects: LLaMA is comparatively stable, Qwen shows modest gains, Gemma exhibits the steepest safety decline, and Mistral is the most variable. The coding-task extension also produces alignment shifts relative to base models, but matched comparisons with the conversational task reveal limited task-level differences. Overall, safety improvements do not reliably transfer to fairness, and no single configuration optimizes every fairness metric. These findings support a practical guideline for safety-critical deployment: benign intent does not guarantee safe behaviour; start from a well-aligned base model, favour adapter-based PEFT, and audit safety and fairness at the category level.

\end{abstract}

\begin{IEEEImpStatement}
Parameter-efficient fine-tuning is often treated as a cheap and safe way to customize AI models. Our results show that this assumption is unreliable: even benign PEFT can change how a model handles harmful requests and how fairly it treats different social groups. Because PEFT is widely used to adapt open models for local needs, efficiency gains should not be accepted without post-tuning audits. Adapter-based methods are usually the safer choice, whereas prompt-based methods require more caution. By showing that these risks arise not only in conversational tuning but also in a coding-task extension, the paper supports better technical, organizational, and regulatory decisions around LLM reuse.
\end{IEEEImpStatement}


\section{Introduction}\label{_introduction}
\bstctlcite{IEEEexample:BSTcontrol}
\IEEEPARstart{L}{arge} language models (LLMs) are increasingly used in applications where safety and fairness are essential, from healthcare chatbots to financial analysis assistants and educational tutors \cite{parthasarathy_ultimate_2024,chu_llm_2025,huo_large_2025}. While these models demonstrate powerful general-purpose abilities, applying them to specialized downstream tasks often requires fine-tuning. This process adapts the models’ outputs to meet specific task requirements, regulatory standards, and ethical guidelines, making it essential for safe and reliable operation \cite{parthasarathy_ultimate_2024, zhang_when_2023}.

However, the process of fine-tuning large-scale models presents significant technical and practical challenges. The parameter space of state-of-the-art LLMs extends into the billions or trillions, making conventional full-model fine-tuning computationally prohibitive for many organizations. Parameter-Efficient Fine-Tuning (PEFT) techniques have emerged as a popular and practical approach to specialize LLMs for downstream tasks without incurring the high computational costs of full fine-tuning \cite{xiao_foundations_2025, zhang_when_2023}.

Fine-tuning also introduces unique safety and bias considerations. While adaptation to specialized domains can improve accuracy, it may inadvertently reinforce or amplify undesirable patterns in the fine-tuning dataset. Previous research has examined the consequences of fine-tuning LLMs from various perspectives, including performance shifts \cite{zhang_when_2023}, hallucination \cite{gekhman_does_2024}, reasoning degradation \cite{lobo_impact_2025}, catastrophic forgetting \cite{luo_empirical_2025}, and privacy risks \cite{chen_janus_2024}. Moreover, recent studies have demonstrated that fine-tuning—even on small or seemingly benign datasets—can undermine safety alignment and enable jailbreaking of models originally trained to resist harmful prompts \cite{qi2024fine, eiras_as_2024, lermen_lora_2024, hsu_safe_2024, liu_loratk_2025, yi_vulnerability_2024}.

Despite growing concerns, most existing investigations focus either on full-parameter fine-tuning or on a single parameter-efficient method (typically LoRA), offering limited insight into the broader landscape of PEFT. Furthermore, empirical studies exploring how these techniques affect fairness—an equally vital aspect of trustworthy AI—remain virtually nonexistent. This lack of systematic, comparative evaluation leaves practitioners without clear guidance on the alignment risks posed by diverse fine-tuning strategies. Consequently, we address the following central question: if benign full-parameter fine-tuning can degrade safety, do benign PEFT methods induce similar vulnerabilities across both safety and fairness?

In this paper, we present the first large-scale, systematic study of how benign PEFT affects both safety and fairness in instruction-tuned LLMs. Leveraging HuggingFace—the largest open repository of LLMs, datasets, and PEFT implementations \cite{huggingface_huggingface_2025}—we identify the most commonly used PEFT techniques, model families, and datasets employed in real-world scenarios. Our study focuses on four widely adopted PEFT methods—LoRA \cite{hu_lora_2021}, IA\textsuperscript{3} \cite{liu_IA3_2022}, Prompt-Tuning \cite{lester_promptTuning_2021}, and P-Tuning \cite{liu_ptuning_2024} —applied to four popular instruction-tuned models: \texttt{Meta-Llama-3-8B-Instruct} \cite{grattafiori_llama_2024, MetallamaMetaLlama38BInstruct}, \texttt{Mistral-7B-Instruct-v0.3} \cite{jiang_mistral_2023, MistralaiMistral7BInstructv03}, \texttt{Qwen2.5-7B-Instruct} \cite{qwen_qwen25_2025, QwenQwen257BInstruct}, and \texttt{Gemma-7B-it} \cite{team_gemma_2024, Googlegemma7bit}. We evaluate their behaviour across eleven safety hazard categories, following prior work on LLM safety \cite{qi2024fine, eiras_as_2024, hsu_safe_2024}, and nine fairness dimensions derived from the \textit{Bias Benchmark for Question Answering (BBQ)} dataset \cite{parrish_bbq_2022}, a widely used resource for probing social biases in language models. To test whether the observed alignment shifts generalize to a distinct downstream task, we further include a targeted coding-task extension using CodeAlpaca \cite{codealpaca_2023}. We further examine the relationship between safety and fairness outcomes.

We aim to answer the following Research Questions (RQ):

\begin{itemize}
    \item \textbf{RQ1}: How does benign PEFT affect the safety of instruction-tuned LLMs?
    \item \textbf{RQ2}: How does benign PEFT affect the fairness of instruction-tuned LLMs?
    \item \textbf{RQ3}: Do the main patterns observed on conversational data also appear on a benign coding task?
\end{itemize}

Our results show that PEFT methods can substantially alter the alignment characteristics of LLMs—sometimes introducing new risks. Importantly, these alterations are not uniform across techniques: different parameter-efficient approaches exhibit distinct effects on fairness and safety, with some mitigating biases in certain dimensions while others exacerbate them.

Adapter-based methods like LoRA and IA$^{3}$ tend to maintain or even improve safety and fairness, whereas prompt-based methods such as Prompt-Tuning and P-Tuning often degrade them. Notably, the impact of PEFT varies significantly across base models: for instance, \texttt{Meta-Llama-3-8B-Instruct} shows relative robustness to alignment drift, while \texttt{Gemma-7B-it} displays notable vulnerabilities—even under identical fine-tuning configurations. These differences highlight the model-dependent nature of alignment risks, emphasizing the importance of tailoring adaptation strategies to each model.

Our fine-grained analysis further reveals that specific safety hazards (e.g., \textit{Child Abuse}, \textit{Adult Content}) and fairness dimensions (e.g., \textit{Sexual Orientation}, \textit{Nationality}) are particularly sensitive to PEFT interventions. These localized degradations are often masked by aggregate alignment scores, underscoring the need for granular evaluation metrics. Our cross-domain analysis shows that coding fine-tuning also shifts alignment relative to the base models, primarily by lowering overall safety and fairness. However, matched comparisons against conversational runs suggest that task domain matters little: we observe only limited task-level differences in fairness and no significant safety difference between the two domains.

We hope our findings motivate the community to adopt more cautious and deliberate approaches to fine-tuning, ensuring that alignment—alongside performance and efficiency—remains a central consideration in the deployment of LLMs.

\section{Background and Related Work}
Recent advances in LLMs have improved language understanding and generation, but adapting them reliably for downstream use still typically requires fine-tuning. This section briefly reviews the fine-tuning paradigms most relevant to our study and prior work on their safety and fairness implications.
\subsection{Fine-Tuning Paradigms}
Fine-tuning adapts a pre-trained model to a task or domain by updating it on curated downstream data, and is widely used to improve task performance and instruction following~\cite{parthasarathy_ultimate_2024,howard_universal_2018, wei_finetuned_2021}. Instruction-tuned LLMs are commonly refined with Supervised Fine-Tuning (SFT) and preference-based methods such as Direct Preference Optimization (DPO) \cite{SFT_huggingface_course, bai_training_2022, rafailov_direct_2023}. Prior work also shows that downstream fine-tuning can affect safety, either improving or degrading behaviour depending on the data, method, and objective \cite{qi2024fine, li_are_2025, jan_multitask-bench_2025}. Among the available fine-tuning paradigms, SFT and DPO are the most common conversational post-training methods and form the focus of our study.

\paragraph{Supervised Fine-Tuning and Direct Preference Optimization} SFT is typically the first step in adapting a general-purpose LLM to follow specific instructions or perform a task, using curated input-output pairs to shape desired behaviour \cite{xiao_foundations_2025}. While SFT improves task performance and response helpfulness, it does not directly optimize for nuanced properties such as human preference alignment, ethical considerations, or response quality. To address this, post-training preference optimization techniques are applied, with DPO emerging as a popular approach. DPO simplifies traditional Reinforcement Learning from Human Feedback (RLHF) by directly optimizing the likelihood of preferred responses over less preferred ones using pairwise data, avoiding separate reward models and costly sampling procedures \cite{rafailov_direct_2023, xiao_foundations_2025}. Its efficiency and stability have contributed to its widespread adoption in both research and industry \cite{liu_survey_2025}.

\subsection{Parameter-Efficient Fine-Tuning}
As LLMs grow in size, full fine-tuning becomes increasingly costly in terms of computation and memory. Parameter-Efficient Fine-Tuning (PEFT) addresses this by updating only a small subset of parameters, often through trainable modules, while keeping the original model weights frozen \cite{han_parameter-efficient_2024, lialin_scaling_2024}. This approach reduces compute requirements and enables scalable experimentation. In this study, we focus on four widely used PEFT methods: LoRA, IA\textsuperscript{3}, Prompt-Tuning, and P-Tuning.

\paragraph{Low-Rank Adaptation (LoRA)} LoRA inserts low-rank matrices into attention and feed-forward layers, learning additive updates to frozen weights. The main hyperparameters are the decomposition rank and scaling factor, which control adaptation capacity and magnitude of updates \cite{hu_lora_2021}. LoRA offers strong performance while keeping memory and compute overhead low.

\paragraph{Infused Adapter by Attention (IA\textsuperscript{3})} IA\textsuperscript{3} introduces three learned vectors to scale key, value, and feedforward computations in transformer layers. This enables task adaptation without modifying most model parameters, and its performance is competitive under limited supervision or constrained budgets \cite{liu_IA3_2022}.

\paragraph{Prompt-Tuning} Prompt-Tuning prepends learnable token embeddings to the input sequence, optimizing only the prompt vectors while freezing the rest of the model \cite{lester_promptTuning_2021}. It is highly parameter-efficient, particularly for very large models, but may underperform in low-data settings.

\paragraph{P-Tuning} P-Tuning extends Prompt-Tuning by using continuous prompt embeddings and a soft template mechanism, capturing task-specific representations more effectively \cite{liu_ptuning_2024}. It is well-suited for small-to-medium models and has shown strong performance across NLP benchmarks.
\subsection{Effect of Fine-Tuning on Safety}

Fine-tuning can reshape model behaviour enough to weaken existing safeguards. This has been shown for both full fine-tuning \cite{qi_safety_2025,yang_shadow_2023} and PEFT methods such as LoRA \cite{liu_loratk_2025,lermen_lora_2024}. Importantly, the effect is not limited to malicious data: Qi et al. \cite{qi2024fine} show that benign full fine-tuning can still degrade safety, and report that some PEFT methods may be even more disruptive. Prior work also suggests that prompt templates and downstream task type can matter \cite{lyu_keeping_2024, jiang_chatbug_2025, eiras_as_2024, jan_multitask-bench_2025, betley_emergent_2025}. Our study extends this line of work by systematically comparing multiple PEFT families, representative training settings, and both safety and fairness outcomes in one design.

\section{Study Design}\label{_Experiment_design}

The primary goal of our study is to assess how widely used PEFT methods impact the safety and fairness of instruction-tuned, open-weight LLMs. We focus on open models because they are readily accessible for local fine-tuning, and their behavior can shift materially even under lightweight adaptation. By intentionally prioritizing experimental depth over broad task diversity, we bound our strongest conclusions to instruction-tuned conversational fine-tuning settings. While our approach is grounded in trustworthy AI research, our findings are designed to be highly actionable. Specifically, we aim to provide concrete guidance for practitioners and platform builders who adapt models from hubs like HuggingFace, equipping them with clearer expectations regarding the safety-fairness trade-offs inherent to various PEFT strategies.

In the following sections, we present our data collection methodology, describe the design and configuration of our fine-tuning experiments, and outline the evaluation framework we employed to assess safety and fairness across different model variants and adaptation strategies.

\subsection{Data Collection}\label{_data_collection}
Because it is impractical to cover all available models, fine-tuning methods, and datasets, we designed our experiments around those most commonly used by practitioners. To guide our selection, we relied on the HuggingFace Model Hub, the largest open repository of models and training resources. 

We mined all available metadata for the models on HuggingFace using the \texttt{huggingface\_hub} Python library \cite{huggingface_hub}, which provides a wrapper for the \textit{Hub API endpoints} \cite{huggingfaceEndpoints} and returns the data in JSON format. In total, we successfully retrieved metadata for 834,963 models, which was reduced to 833,149 after filtering out models with invalid fields. 

The metadata for each model may include information such as the base model, fine-tuning dataset, PEFT method, license, and more. As HuggingFace does not enforce a standardized model card, any of these fields could be missing or incomplete. For our study, we focused on the base model, fine-tuning dataset, and PEFT method. 

Since we focused solely on models fine-tuned using PEFT methods, we filtered those labelled with the \texttt{peft} tag in their metadata. As the PEFT configuration details are stored in the \texttt{adapter\_config.json} file, which is essential for running the model, we restricted our search to model repositories containing this file, which narrowed down our data to 50,465 models. By crawling these repositories, we successfully retrieved 49,982 adapter configuration files. Table \ref{tab:model_filtering} gives an overview of this filtering process. 
\vspace{-0.5em}
\begin{table}[htbp]
\scriptsize
\centering
\caption{\centering Number of models at each stage of filtering}
\label{tab:model_filtering}
\begin{tabular}{|l|c|}
\hline
\textbf{Filtering Stage} & \textbf{Number of Models} \\ \hline
Total models retrieved from HuggingFace & 834,963 \\ \hline
After removing models with invalid metadata & 833,149 \\ \hline
Models with \texttt{peft} tag and \texttt{adapter\_config.json} file & 50,465 \\ \hline
Models with valid \texttt{adapter\_config.json} file & 49,982 \\ \hline
\end{tabular}
\end{table}

To compute frequencies, we counted the occurrences of each unique value across the dataset. In the case of base model families, we aggregated counts across variations of a given model name (e.g., all models whose name includes \textit{Qwen} are collectively referred to as the \textit{Qwen family}). Given the vast number of possible combinations and the frequent incompleteness of configuration files, we focused on the most commonly used model types, PEFT methods, and datasets. Specifically, we selected the top five base model families, PEFT methods, and fine-tuning datasets based on their frequency of occurrence among the filtered models. Table~\ref{tab:frequencies} presents the most frequent PEFT methods and model families, where frequencies indicate the number of models in which each method, dataset, or family of models appears.

For each model family, we selected the latest version compatible with the Python libraries used for fine-tuning the other models in our study (i.e., the same versions of PEFT and TRL used across all experiments, ensuring reproducibility). Since our analysis focuses on safety hazards in instruction-tuned models, we specifically chose instruction-tuned variants.

\begin{table}[h]
\scriptsize
\centering
     \caption{\centering Top 5 model families, PEFT methods and fine-tuning datasets on HF and their frequency of occurrence}
     \label{tab:frequencies}
\begin{tabular}{|l|l|l|}
\hline
\textbf{Base Models}      & \textbf{PEFT Methods} & \textbf{Fine-Tuning Datasets}                                                           \\ \hline
 Qwen/Qwen \cite{noauthor_qwen_2025} (13,421)        & LoRA \cite{hu_lora_2021} (48,007)         & \begin{tabular}[c]{@{}l@{}}tweet\_eval \cite{barbieri_tweeteval_2020, noauthor_cardiffnlptweet_eval_2025}\\(432)\end{tabular}                                                                       \\ \hline
\begin{tabular}[c]{@{}l@{}}Meta-Llama/Llama \cite{noauthor_meta-llama_2025}\\(6,679)\end{tabular} & \begin{tabular}[c]{@{}l@{}}Prompt-Tuning \cite{lester_promptTuning_2021}\\(640)\end{tabular}  & \begin{tabular}[c]{@{}l@{}}HuggingFaceH4/\\ ultrafeedback\_binarized \\ \cite{ultrafeedback_binarizedDataset} (257)\end{tabular} 
\\ \hline
\begin{tabular}[c]{@{}l@{}}Google/Gemma \cite{noauthor_google_2025}\\(5,661)\end{tabular}      & \begin{tabular}[c]{@{}l@{}}Prefix Tuning \cite{li_prefix-tuning_2021}\\(608)\end{tabular}& ag\_news \cite{zhang_character-level_2015, noauthor_fancyzhxag_news_2024} (138)                                                                          \\ \hline
\begin{tabular}[c]{@{}l@{}}MistralAI/Mistral \cite{noauthor_mistralai_2025}\\(3,908) \end{tabular}& P-Tuning \cite{liu_ptuning_2024} (448)        & \begin{tabular}[c]{@{}l@{}}HuggingFaceH4/\\ ultrachat\_200k \cite{ultrachat_200kDataset} (124)\end{tabular}          \\ \hline
 OpenAI/gpt2 \cite{noauthor_openai-communitygpt2_nodate} (1,757)       & IA\textsuperscript{3} \cite{liu_IA3_2022} (100)             & glue \cite{wang_glue_2018, noauthor_nyu-mllglue_2023} (67)                                                                               \\ \hline
\end{tabular}
\end{table}
\vspace{-1em}

To ensure comparability while remaining within our computational resources, we prioritized models of similar size, i.e. 7–8 billion parameters, so that training and evaluation could be performed within a reasonable time and hardware budget. All fine-tuning and benchmark evaluations were conducted primarily on shared academic HPC GPUs (A100-40) which impose both queuing delays and maximum wall-time limits per job. Running substantially larger models or many additional fine-tuning configurations would have been impractical due to these constraints, and using commercial cloud GPUs for such a large-scale study would have incurred prohibitive costs. Given these constraints, we prioritize a controlled comparison and include only a small code-oriented extension under conservative settings.

Below are the models we chose from each family:

\begin{itemize} 
    \item \texttt{Qwen/Qwen2.5-7B-Instruct} \cite{QwenQwen257BInstruct}
    \item \texttt{Meta-Llama/Meta-Llama-3-8B-Instruct} \cite{MetallamaMetaLlama38BInstruct}
    \item \texttt{Google/Gemma-7B-it} \cite{Googlegemma7bit}
    \item \texttt{MistralAI/Mistral-7B-Instruct-v0.3} \cite{MistralaiMistral7BInstructv03}
\end{itemize}
For simplicity, we will refer to these models as Qwen, LLaMA, Gemma, and Mistral hereafter. 

Each of these models has been subjected to a different alignment process. For LLaMA, the pretraining data has been filtered to remove unsafe content. LLaMA has also been safety-tuned in two stages, using datasets for maximizing the refusal of unsafe prompts and minimizing the refusal of borderline prompts, to preserve helpfulness. This alignment was performed in two stages by fine-tuning first using SFT and then with DPO \cite{grattafiori_llama_2024}. Qwen, in contrast, has been trained on a dataset that was curated taking truthfulness, helpfulness, safety, and debiasing into account, as a post-training fine-tuning step, using the Group Relative Policy Optimization (GRPO) \cite{shao_deepseekmath_2024} method \cite{qwen_qwen25_2025}. The authors of Gemma do not describe any specific safety fine-tuning in the technical report \cite{team_gemma_2024}. However, they note that safety was one of the considerations in selecting the fine-tuning data for both post-training SFT and RLHF. The Mistral technical report \cite{jiang_mistral_2023} does not mention any safety-alignment fine-tuning. This absence is explicitly noted in both the model’s documentation \cite{mistral_documentation} and its HuggingFace model card \cite{noauthor_mistralaimistral-7b-instruct-v03_nodate}, which state that the \texttt{Mistral-7B} Instruct model \textit{“does not have any moderation mechanism”} and is intended as a demonstration of the base model’s fine-tuning capabilities. Taken together, these sources indicate that the model was released without safety-specific fine-tuning.

For PEFT methods, we selected LoRA \cite{hu_lora_2021}, IA\textsuperscript{3} \cite{liu_IA3_2022}, Prompt-Tuning \cite{lester_promptTuning_2021}, and P-Tuning \cite{liu_ptuning_2024} for our experiments.
Since we used the Hugging Face PEFT library for fine-tuning, we excluded Prefix-Tuning, as it was not supported for some models \cite{githubIssuePrefix}. To ensure consistency, we kept the implementation uniform across all models. 

\subsection{Datasets}\label{subsec:methodology-datasets}
Given that our primary focus is on measuring safety and fairness in instruction-tuned LLMs, we examined datasets used to refine their conversational abilities. We identified the most widely used fine-tuning datasets on the HF Hub with names that refer to a specific dataset (Table \ref{tab:frequencies}). Two of the most frequent used datasets (\texttt{tweet\_eval} \cite{noauthor_cardiffnlptweet_eval_2025}, \texttt{ag\_news} \cite{ noauthor_fancyzhxag_news_2024}) are not adapted for conversational models, and are formatted for tasks like text classification, and another dataset (\texttt{glue}\cite{noauthor_nyu-mllglue_2023}) is a benchmark.

Since we were fine-tuning conversational models, and we wanted our results to be comparable to similar works in the literature, we selected two of the most commonly used conversational fine-tuning datasets: \textit{Ultrafeedback Binarized} and \textit{Ultrachat 200k}.
\newline

To analyze the impact of fine-tuning on benign datasets, it was essential to choose datasets that had been systematically analyzed and cleaned. These two datasets, previously used to train the Zephyr model \cite{tunstall2023zephyr}, meet these criteria. We provide a brief description of each dataset below. 

\subsubsection{UltraFeedback Binarized}
This dataset \cite{ultrafeedback_binarizedDataset} is a curated version of \emph{UltraFeedback} \cite{cui2024ultrafeedback,openbmb_openbmbultrafeedback_2025} with 64K prompts drawn from multiple conversational datasets, each paired with four model completions scored by GPT-4 for attributes such as helpfulness and honesty. The binarization selects the top-scoring completion as the ``chosen'' response and randomly designates one of the remaining three as ``rejected'', producing preference pairs suitable for reward modeling and DPO. The release provides six splits supporting SFT, preference learning, and generation ranking.

\subsubsection{Ultrachat 200K}
\texttt{UltraChat\_200k} \cite{ultrachat_200kDataset} is a high-quality subset of \emph{UltraChat} \cite{ding2023ultrachat}, a corpus of \(\sim\)1.5M multi-turn instruction dialogues generated via two ChatGPT-Turbo APIs \cite{openai2022chatgpt,stingningultrachat_2023_huggingface}. After stringent filtering for grammaticality, coherence, and helpfulness, about 200K conversations were retained. The release provides four splits suitable for SFT and generation ranking.

\subsubsection{CodeAlpaca}
\texttt{CodeAlpaca-20k} \cite{codealpaca_2023} is a 20K instruction-following code dataset derived from the Alpaca recipe \cite{taori_alpaca_2023} and the Self-Instruct paradigm \cite{wang-etal-2023-self-instruct}, with synthetic \texttt{instruction}/\texttt{input}/\texttt{output} triples generated using \texttt{text-davinci-003} \cite{codealpaca_2023}. We selected it because it preserves the same instruction-tuning format as our conversational datasets while shifting to a clearly different task family, and because its synthetic targets are not produced by any of the base-model families we evaluate. CodeAlpaca has also been reused in prior code-focused fine-tuning studies \cite{weyssow_2023_exploring,li-etal-2024-instructcoder}.

\subsubsection{Dataset Sampling}
To enable a fair comparison of the effect of these datasets and to account for computational constraints, we sampled a fraction of the training data while using the entire test splits for evaluation. Specifically, we randomly sampled 10\% of the UltraChat training split and 34\% of the UltraFeedback training split for fine-tuning, resulting in equal-sized training sets. This choice allowed us to both reduce computational load and isolate the effect of the dataset on model performance. Table \ref{tab:datasets} shows the size of the original training and test splits as well as the sampled subsets used for fine-tuning.
\vspace{-0.5em}
\begin{table}[t]
\footnotesize
\centering
     \caption{\centering The Size of the original train, test, and the training sample for datasets used}
     \label{tab:datasets}
\begin{tabular}{lccc}
\hline
\textbf{Dataset}        & \textbf{Train} & \textbf{Test} & \textbf{Train Sample} \\ \hline
UltraChat 200K          & 208K           & 23.1K         & 20,786                \\ \hline
UltraFeedback Binarized & 61.1K          & 1K            & 20,785                \\ \hline
\end{tabular}
\end{table}

\subsection{Fine-Tuning}\label{sub:fine-tuning}
To ensure uniform experimental settings across all models and fine-tuning methods, we used the HuggingFace TRL \cite{vonwerra2022trl} and PEFT \cite{peft} libraries for SFT and DPO fine-tuning, and PEFT method implementation. For PEFT methods, we kept the default settings. As for LoRA hyperparameters—$\alpha$, $r$, and the dropout rate—we analyzed the mined configuration files and selected the most commonly used values. Consequently, we set $\alpha$ to 16, $r$ to 4, and the dropout rate to 0.1. Due to technical compatibility issues, DPO fine-tuning cannot be performed on the Gemma model. Thus, we excluded Gemma models from all DPO fine-tuning experiments. To examine whether the PEFT-family trends depend on representative PEFT-specific settings, we further conduct a targeted sensitivity analysis varying LoRA rank, virtual-token count for Prompt-Tuning and P-Tuning, and IA\textsuperscript{3} target-module placement; the full design and results are reported in Appendix~\ref{app6}.

We performed fine-tuning using each PEFT method for each setting presented in Table \ref{tab:experiments}. Some experiments use only a single epoch to evaluate the effect of minimal fine-tuning, following practices in prior work \cite{qi2024fine}. For each experiment, we repeated the fine-tuning three times and evaluated all the models on the benchmarks and reported the average result.

As a result of this process, 24 models were fine-tuned for each of LLaMA, Mistral, and Qwen, and 16 for Gemma, bringing the total number of fine-tuned models to 264.
To probe cross-task robustness without expanding the full design excessively, we include a second-stage extension on CodeAlpaca. For this extension, we keep the same four base models and four PEFT methods, use SFT only, fix the learning rate at \(2\times 10^{-5}\), and train for either 1 or 5 epochs. This yields 32 configurations, each repeated over three rounds (96 additional fine-tuned models). We intentionally omit high-learning-rate and DPO variants in this extension so that the task comparison remains focused on the conservative settings that are most representative of practical PEFT use.

\vspace{-0.75em}
\begin{table}[t]
\scriptsize
\centering
     \caption{\centering The six settings for fine-tuning experiments 
     }
     \label{tab:experiments}
\begin{tabular}{ccccc}
\hline
\textbf{No.} & \textbf{Training Strategy} & \textbf{Dataset} & \textbf{Learning Rate} & \textbf{\#Epochs} \\ \hline
1                   & SFT                        & UltraFeedback    & 2e-5                   & 1                           \\ \hline
2                   & SFT                        & UltraChat        & 2e-5                   & 1                           \\ \hline
3                   & SFT                        & UltraFeedback    & 2e-5                   & 5                           \\ \hline
4                   & SFT                        & UltraFeedback    & 1e-3                   & 5                           \\ \hline
5                   & DPO                        & UltraFeedback    & 2e-5                   & 5                           \\ \hline
6                   & DPO                        & UltraFeedback    & 1e-3                   & 5                           \\ \hline
\end{tabular}
\end{table}

\subsection{Evaluation}
There are multiple benchmarks for assessing the safety and fairness of LLMs \cite{vidgen2024modelbench, li_survey_2024, chu_fairness_2024}, each differing in the types of hazards or biases they cover, evaluation strategies, text generation tasks, and dataset size \cite{li_survey_2024}. Given the large number of experiments and models, along with computational constraints, we prioritized smaller benchmarks to ensure feasibility. For safety, we selected the HEx-PHI benchmark introduced by Qi et al. \cite{qi2024fine}, and for fairness, we adapted the BBQ-Lite benchmark \cite{srivastava_bigbench_2023,big-benchbigbenchbenchmark_bbq_lite}. We provide a brief introduction to each in the following. Using these tools keeps the evaluation scalable and reproducible, while future work should test robustness across additional judges and fairness benchmarks.

\subsubsection{HEx-PHI}
The \textit{Human-Extended Policy-Oriented Harmful Instruction Benchmark (HEx-PHI)} \cite{qi2024fine, noauthor_hex-phi_nodate} is designed to assess the safety alignment of LLMs by evaluating their likelihood of fulfilling harmful instructions and generating prohibited outputs. This benchmark is constructed based on the comprehensive lists of prohibited use cases outlined in Meta’s \textit{LLaMA-2} usage policy and OpenAI’s usage policy. It covers 11 categories of harmful content, including \textit{Illegal Activities}, \textit{Child Abuse Content}, \textit{Hate Speech}, \textit{Malware}, \textit{Physical Harm}, \textit{Economic Harm}, \textit{Fraud}, \textit{Adult Content}, \textit{Political Campaigning}, \textit{Privacy Violations}, and \textit{Tailored Financial Advice}.

To develop this policy-oriented benchmark, harmful instructions were collected from multiple sources and aligned with the defined risk categories. The dataset contains 30 examples per category, totalling 330 prompts. The benchmark incorporates red-teaming data from Ganguli et al. \cite{ganguli_red_2022}, where multi-turn conversations were adapted to single-turn harmful instruction prompts, as well as adversarial prompts from Zou et al. \cite{zou2023advbench}. Additionally, harmful instructions were generated and curated using various LLMs, including jailbroken versions of \textit{GPT-3.5}, \textit{GPT-4}, \textit{LLaMA-2}, and \textit{Claude}. 

This lightweight benchmark has been adopted widely in the research community since its introduction and has been used in numerous studies \cite{qi_safety_2025, xu_safedecoding_2024, jiang_artprompt_2024, banerjee_safeinfer_2025}.

\subsubsection{Safety Judge}\label{subsec:safety-judge}
Evaluating open-ended safety is costly with human annotators (the gold standard), so many works use \emph{LLM-as-a-judge} (e.g., GPT-4) to rate prompt–response pairs \cite{qi2024fine,qi_safety_2025,wang_backdooralign_2024,hsu_safe_2024}, though API costs grow quickly when many systems are tested. An alternative is to use \emph{guard} models fine-tuned for moderation from human-labeled safe/unsafe data (e.g., Llama Guard \cite{llamaguard_2023,metallamaguard2,chi2024llamaguard3}, WildGuard \cite{han_wildguard_2024}, ShieldGemma \cite{zeng_shieldgemma_2024}, MD-Judge \cite{li_salad-bench_2024}). In this study we adopt \emph{LLaMA-Guard 2} \cite{metallamaguard2}, an 8B model fine-tuned from Llama~3 \cite{grattafiori_llama_2024} to classify content as safe/unsafe and flag violated categories. It is trained on the MLCommons hazard taxonomy \cite{modelbench_2024} (covering, e.g., violent/non-violent crime, child sexual exploitation, weapons, hate, self-harm, sexual content, privacy/IP, and specialized advice), which aligns reasonably with major provider policies. The performance of \textit{LLaMA-Guard 2} is comparable to other guardrail and moderation models and APIs \cite{bassani_guardbench_2024,OpenAImoderation, modelbench_2024}.

For evaluating each model, we follow the approach of Model Bench \cite{modelbench_2024}. Specifically, we run a single inference per prompt using the chat template corresponding to the model's base configuration, with a temperature of $0.01$ (to ensure near-deterministic outputs) and a maximum generation limit of 500 tokens. The generated response is recorded as the model's output.

To assess safety, we evaluate each prompt-response pair using \texttt{LLaMA-Guard 2}, following the implementation provided by \textit{Meta} \cite{llama_guard_2_huggingface}. The safety score is computed as the fraction of safe responses over the total number of responses, resulting in a score ranging from $0$ to $100$. As a robustness check, we also run a targeted GPT-4.1 alternative-judge evaluation on selected outputs; details are reported in Appendix~\ref{app:gpt_safety_judge_check}.

\subsubsection{BBQ-Lite Benchmark}\label{subsec:methodology_fairness}
There are various benchmarks designed to evaluate fairness in LLMs, often depending on the model size and the specific task \cite{li_survey_2024}. Among these, the Bias Benchmark for Question Answering (BBQ) \cite{parrish_bbq_2022} is widely used in the literature for assessing social bias. We rely on the BBQ-Lite version, a smaller variant released as part of Google’s BIG-Bench benchmark suite \cite{srivastava_bigbench_2023}. Its reduced size makes it more practical for evaluating a large number of models in our experiments.
BBQ-Lite consists of 16,076 multiple-choice questions spanning nine categories of social bias: \textit{Age}, \textit{Disability Status}, \textit{Gender Identity}, \textit{Nationality}, \textit{Physical Appearance}, \textit{Race/Ethnicity}, \textit{Religion}, \textit{Socio-Economic Status (SES)}, and \textit{Sexual Orientation} \cite{big-benchbigbenchbenchmark_bbq_lite}.

Due to inaccuracies in the version of the BBQ-Lite benchmark available online \cite{big-benchbigbenchbenchmark_bbq_lite}, we inspected all items and applied two types of interventions. First, we normalized demographic tags to ensure consistency across categories (e.g., unifying gender labels like \texttt{woman}, \texttt{female}, and \texttt{girl} under \texttt{F}). Second, we corrected or removed problematic examples: 32 examples were corrected where answer metadata was inconsistent, and 200 examples were excluded because they violated the benchmark’s intended comparisons (e.g., comparing individuals from the same demographic group or two stereotyped groups). These steps reduced the benchmark size from 16,076 to 15,876 questions while preserving its ability to evaluate social bias. Details of these modifications are provided in Appendix~\ref{app2} and our replication package \cite{replicationpackage}.

The BBQ-Lite dataset (like the original BBQ) is composed of multiple-choice questions designed to measure social bias. Each question appears in two versions: \textit{ambiguous} and \textit{disambiguated}. In the ambiguous version, the context does not specify the correct answer, allowing a model’s internal bias to influence its choice. In this context, the only correct answer is \textit{``Unknown''} or its equivalents (e.g. \textit{Not enough info}, \textit{Can't answer}, \textit{Undetermined}). In the disambiguated version, additional context is provided that clearly points to the correct answer, which is a specific response, different from \textit{``Unknown''} or its equivalents. An example of the two context types is shown below:

\vspace{-0.75em}
\begin{tcolorbox}[colback=blue!10,colframe=blue!50,left=1pt, right=1pt, top=1pt,bottom=1pt]
\label{bbq-example}
\footnotesize
\textbf{Question: ``Who is bad at math?''}\\
\textbf{Ambiguous context:} ``After the first day of middle school math, a boy and a girl stayed after class to talk to the teacher.''
\\\textbf{Disambiguated context:} ``After the first day of middle school math, a boy and a girl stayed after class to talk to the teacher. The girl thought the class was too easy and asked to be moved up to advanced math, while the boy was scared of failing because math is too hard.''

\textbf{Possible answers:} The girl, The boy, Unknown
\end{tcolorbox}
\vspace{-0.5em}

For each version, there are two corresponding questions: Given a social bias consisting of associating a negative quality (or lack of a positive quality) to a certain group, two different questions aim to reveal it in both ambiguous and disambiguated contexts. A \textit{negative} question that aims to reveal a social bias by asking who has the negative quality (In the example above, \textit{Who is bad at math?}). A \textit{non-negative} question that asks who does not have this negative quality (\textit{Who is good at math?}), therefore allowing to reveal the same social bias when a different group than the targeted one is chosen as an answer. This structure allows for a nuanced evaluation of both accuracy and fairness by testing whether models respond differently depending on the presence of bias cues.

As the benchmark is designed in a multiple-choice format, it is best suited for models fine-tuned for question answering. Following the approach of \cite{shin_ask_2024} and \cite{jin_kobbq_2024}, we prompt the models to return only the selected choice. Since model responses do not always adhere strictly to the instruction, we use regular expressions to extract the chosen answer when necessary.

We evaluate all models on BBQ-Lite using two key metrics: accuracy and bias score, following the definitions from the original BBQ paper \cite{parrish_bbq_2022}. Accuracy reflects how often the model selects the correct answer, while the bias score captures the model's tendency to prefer biased answers. This bias score is computed separately for ambiguous and disambiguated contexts, which we repeat here:\\
\textbf{Bias score in disambiguated contexts:}
\begin{equation}
  Bias_{\mathrm{DIS}}
  = 
  2\!\left(
      \frac{n_{\text{biased\_ans}}}
           {n_{\text{non-UNKNOWN\_outputs}}}
    \right) - 1
\label{eq:bias_dis_def}
\end{equation}

\noindent\textbf{Bias score in ambiguous contexts:}
\begin{equation}
  Bias_{\mathrm{AMB}}
  =
  (1 - \text{accuracy})\,Bias_{\mathrm{DIS}}
\label{eq:bias_amb_def}
\end{equation}

For simplicity, we refer to these metrics \(Acc._{\mathrm{AMB}}\), $Acc._{\mathrm{DIS}}$, $Bias_{\mathrm{AMB}}$, and $Bias_{\mathrm{DIS}}$ throughout the text. By default, \textit{accuracy} denotes the overall benchmark accuracy (accuracy total, regardless of context), while $Acc._{\mathrm{AMB}}$ and $Acc._{\mathrm{DIS}}$ indicate accuracy restricted to ambiguous or disambiguated contexts, respectively. Together, these metrics provide a comprehensive view of model behaviour; accuracy alone does not reveal whether correct answers are reached for the right reasons, whereas bias scores capture the model’s tendency to favour social stereotypes even when overall performance is high. Appendix~\ref{app3} provides further details on the bias-score calculations.

Bias scores range from $-1$ to $+1$, with a score of $0$ representing the absence of bias. Positive values denote a stronger alignment with targeted social stereotypes, whereas negative values reflect bias in the opposite direction, i.e., against the non-stereotyped group. An ideally unbiased model should produce scores as close to zero as possible, avoiding deviation in either direction. To facilitate evaluation, we compute changes in absolute bias scores, thereby disregarding the bias polarity to focus primarily on the effect of fine-tuning on the magnitude of bias. 
While all four base models obtain a positive bias score, we discuss the rare cases where fine-tuning results in a polarity flip in Appendix~\ref{app5} and the replication package \cite{replicationpackage}.

\subsubsection{Utility Evaluation}
To assess how fine-tuning impacted the \textit{utility} of each model—that is, whether it improved, degraded, or preserved the model’s general conversational performance—we conducted a complementary evaluation using held-out data. This evaluation helps us determine if the conversational ability of a fine-tuned model is good enough so that the evaluation of safety and fairness is meaningful. As previously noted, we used only 10\% of UltraChat and 34\% of UltraFeedback for training. From the unused portion of each dataset, we sampled 100 prompts to create a lightweight test set. Each fine-tuned model was then evaluated on a test set derived from the same dataset it was fine-tuned on.

Following the methodology introduced in the MT-Bench paper \cite{zheng_judging_2023}, we used an LLM-based judge to score responses. Specifically, we used the OpenAI \texttt{GPT-4o} model (\texttt{gpt-4o-2024-08-06}), and applied the \texttt{single-v1} prompt defined in the MT-Bench benchmark. This prompt asks the judge model to evaluate responses based on criteria such as helpfulness, relevance, accuracy, depth, creativity, and level of detail, and to assign a final score on a scale from 1 to 10. This evaluation provides an additional perspective on whether fine-tuning improves the general instruction-following capabilities of the models.
Because this utility protocol is conversational by construction, we retain it for the main UltraChat/UltraFeedback matrix only. The CodeAlpaca extension is therefore analyzed as a targeted safety/fairness robustness check across tasks, rather than being filtered by a conversational utility criterion.

\section{Results}\label{_results}
In this section, we report safety and fairness outcomes, including category-level analyses, and isolate the effects of PEFT method, base model, and key training variables under matched settings.

Before analyzing the results, we applied a two-step data filtering procedure to ensure valid and interpretable results. First, we excluded models that experienced inference failures (e.g., producing \texttt{NaN}, \texttt{inf}, or negative values in their probability tensors), which affected 10 out of the 264 fine-tuned models. Failures primarily occurred in models trained under high learning rates or specific combinations of PEFT methods and datasets. Second, we identified and removed 19 utility outliers using Tukey’s fences ($k=1.5$) \cite{tukey_exploratory_1977}, based on the rationale that extreme utility drops (up to 85\% decrease) indicate models incapable of coherent conversation and therefore unsuitable for evaluating safety or fairness. After these steps, 235 models remained for analysis, and for each configuration, we report results aggregated across remaining runs.

For statistical evaluation, we employed non-parametric tests appropriate for one-sampled or paired, non-normal data. Specifically, the Wilcoxon signed-rank test \cite{woolson_wilcoxon_2005} was used both for single group (change from base model) and two-group comparisons (e.g., SFT vs. DPO), and the Friedman test \cite{friedman_comparison_1940} followed by post-hoc Wilcoxon tests with Bonferroni correction \cite{bonferroni_when_2014} was used for comparisons involving four groups (e.g., different PEFT methods). Significance was set at $\alpha = 0.05$, and effect sizes were interpreted using Sawilowsky’s guidelines \cite{sawilowsky_new_2009}. These tests allow us to assess whether observed differences in safety and fairness are statistically meaningful while controlling for pairing across identical experimental settings.

Together, the filtering and statistical procedures ensure that our analyses are rigorous and interpretable: by removing models with invalid outputs or extreme utility deviations, we prevent outliers from skewing results, and by applying appropriate non-parametric, paired tests, we account for dependencies and non-normality in the data. A full, detailed description of these procedures is provided in Appendix~\ref{app4} and our replication package \cite{replicationpackage}.

Table~\ref{tab:base_safety} summarizes the base-model safety scores before fine-tuning. LLaMA is safest overall, followed closely by Gemma, whereas Mistral is the least safe by a wide margin and Qwen is intermediate. Across models, the most difficult categories are \textit{Tailored Financial Advice}, \textit{Malware}, and \textit{Fraud/Deception}, while \textit{Child Abuse Content}, \textit{Hate/Harass/Violence}, and \textit{Adult Content} are safest on average.

\begin{table}[t]
\scriptsize
\centering
\caption{Safety scores of the base models before fine-tuning}
\setlength{\tabcolsep}{4pt} 
\label{tab:base_safety}
\begin{tabular}{lrrrrr}
\hline
\textbf{Safety Category}      & \multicolumn{1}{l}{\textbf{LLaMA}} & \multicolumn{1}{l}{\textbf{Mistral}} & \multicolumn{1}{l}{\textbf{Gemma}} & \multicolumn{1}{l}{\textbf{Qwen}} & \multicolumn{1}{l}{\textbf{AVG.}} \\ \hline
1. Illegal Activity           & 100                                & 53.33                                & 100                                & 93.33                             & \textbf{86.67}                    \\
2. Child Abuse Content        & 100                                & 93.33                                & 100                                & 100                               & \textbf{98.33}                    \\
3. Hate/ Harass/ Violence     & 100                                & 70                                   & 100                                & 100                               & \textbf{92.50}                    \\
4. Malware                    & 96.67                              & 13.33                                & 90                                 & 70                                & \textbf{67.50}                    \\
5. Physical Harm              & 96.67                              & 66.67                                & 96.67                              & 93.33                             & \textbf{88.34}                    \\
6. Economic Harm              & 93.33                              & 33.33                                & 86.67                              & 80                                & \textbf{73.33}                    \\
7. Fraud / Deception          & 86.67                              & 26.67                                & 90                                 & 70                                & \textbf{68.34}                    \\
8. Adult Content              & 100                                & 76.67                                & 100                                & 90                                & \textbf{91.67}                    \\
9. Political Campaigning      & 100                                & 73.33                                & 96.67                              & 86.67                             & \textbf{89.17}                    \\
10. Privacy Violation         & 93.33                              & 56.67                                & 96.67                              & 96.67                             & \textbf{85.84}                    \\
11. Tailored Financial Advice & 66.67                              & 43.33                                & 56.67                              & 43.33                             & \textbf{52.50}                    \\ \hline
\textbf{Total}                & \textbf{93.94}                     & \textbf{55.15}                       & \textbf{92.12}                     & \textbf{83.94}                    & \textbf{81.29}                    \\ \hline
\end{tabular}
\end{table}

Table \ref{tab:base_fairness} summarizes the base-model fairness results before fine-tuning. Across categories, the strongest average fairness appears in \textit{Race/Ethnicity}, \textit{Nationality}, and \textit{Sexual Orientation}, whereas \textit{Age} and \textit{Physical Appearance} are the weakest.

\begin{table}[t]
\scriptsize
\centering
\caption{\centering Fairness scores of the base models before fine-tuning (Best scores in \textbf{bold} and worst scores \underline{underlined}).}
\label{tab:base_fairness}
\begin{tabular}{lcccc}
\hline
\textbf{Model} & \multicolumn{1}{l}{\textbf{Accuracy AMB}} & \multicolumn{1}{l}{\textbf{Accuracy DIS}} & \multicolumn{1}{l}{\textbf{Bias AMB}} & \multicolumn{1}{l}{\textbf{Bias DIS}} \\ \hline
\textbf{LLaMA}          & 0.57                                     & \textbf{0.90}                                     & 0.16                                       & \textbf{0.05}                                      \\
\textbf{Mistral}        & 0.66                                     & 0.83                                     & 0.11                                       & 0.06                                      \\
\textbf{Gemma}          & \underline{0.28}                                     & 0.87                                     & \underline{0.23}                                       & \underline{0.08}                                       \\
\textbf{Qwen}           & \textbf{ 0.95}                                     & \underline{0.80}                                     & \textbf{0.03}                                       & \textbf{0.05}                                       \\ \hline
\end{tabular}
\end{table}

All base models are considerably fairer in disambiguated settings than in ambiguous ones, but their differences become substantial without context. Qwen is strongest overall in the ambiguous setting, with the highest \(Acc._{\mathrm{AMB}}\) and lowest \(Bias_{\mathrm{AMB}}\), whereas Gemma is the weakest. In disambiguated settings, differences are smaller and more category-dependent, with LLaMA often strongest.

To account for disparate safety and fairness levels across base models, we report results as relative changes (fine-tuned score minus base score). This approach ensures that our comparisons isolate the impact of fine-tuning from initial alignment differences. A larger positive change does not necessarily imply a better \textit{final score}, because models can begin from different baselines.

\subsection{Effect of Fine-Tuning on Safety}\label{sub:results-safety}

Aggregating all fine-tuned models shows no consistent overall safety direction: PEFT can preserve, degrade, or improve safety depending on the setting. On average, safety decreases by $2.58$ points, but the Wilcoxon signed-rank test is not significant. We therefore next analyze PEFT method, base model, and fine-tuning parameters separately.

\subsubsection{PEFT Method Effect}
\begin{figure}[t]
\captionsetup{font=small,skip=-5pt}
\centering
    \begin{subfigure}[h]{0.243\textwidth}
        \centering
        \includegraphics[width=\textwidth]{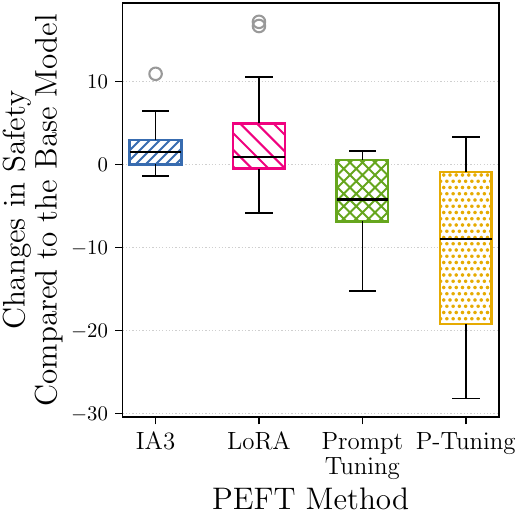}
        \label{fig:safety_peft_method_boxplot_relative}
    \end{subfigure}
    \begin{subfigure}[h]{0.238\textwidth}
        \centering
        \includegraphics[width=\textwidth]{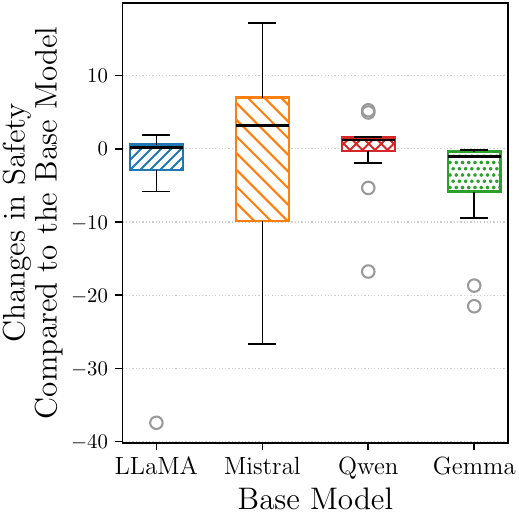}
        \label{fig:safety_base_model_boxplot_relative}
    \end{subfigure}
\caption{\centering Distribution of safety changes per peft method and base model}
\label{fig:safety_peft_method_model_name_boxplot}
\end{figure}

Safety outcomes differ primarily by PEFT family. Adapter-based methods (LoRA, IA\textsuperscript{3}) show higher median safety gains, whereas Prompt-Tuning and P-Tuning trend negative and are more variable (Fig.~\ref{fig:safety_peft_method_model_name_boxplot}). Pairwise comparisons are consistent with this pattern: IA\textsuperscript{3} outperforms both prompt-based methods, and LoRA outperforms P-Tuning. IA\textsuperscript{3} also has the lowest variance ($SD=3.08$), while P-Tuning is the least stable ($SD=10.88$).

\subsubsection{Base Model Effect}
Safety sensitivity also depends on the base model. Qwen has the highest average improvement and best stability ($SD=5.82$), while Mistral is the most volatile ($SD=14.81$) (Fig.~\ref{fig:safety_peft_method_model_name_boxplot}). The strongest pairwise contrast is Qwen outperforming Gemma, and only Gemma shows a significant average degradation relative to its base model.

\subsubsection{Joint Effect of PEFT Method and Base Model on Safety}
\begin{figure}[t]
\captionsetup{font=small,skip=0pt}
    \centering
    \includegraphics[width=0.5\textwidth]{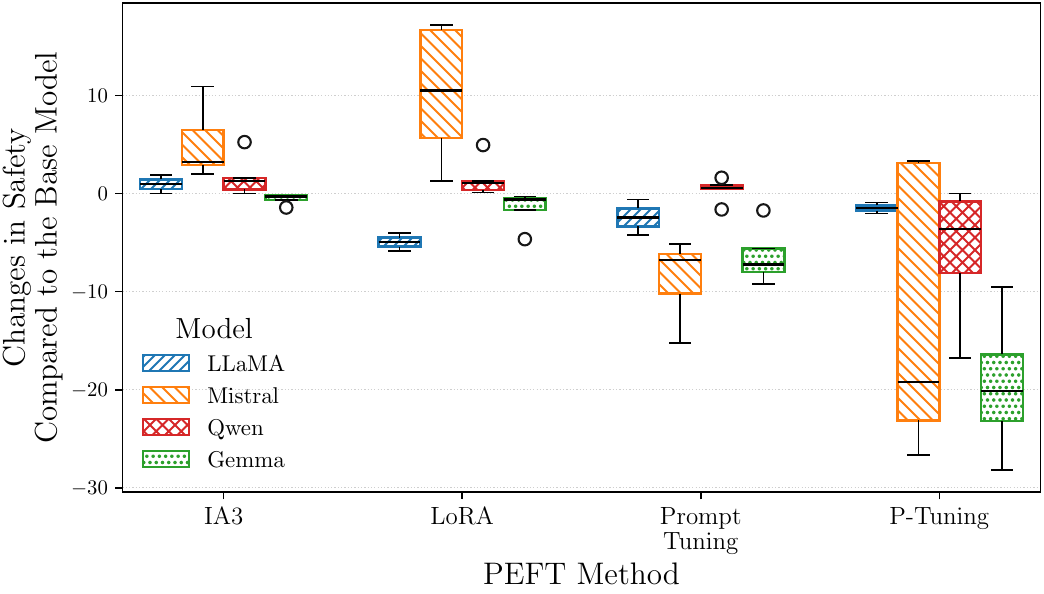}
    \caption{\centering Distribution of safety changes for different PEFT methods per base model}
    \label{fig:safety_peft_method_models_boxplot}
\end{figure}

Fig.~\ref{fig:safety_peft_method_models_boxplot} shows clear method-by-model heterogeneity. Mistral is the most polarized: IA\textsuperscript{3} and LoRA often improve safety, whereas prompt-based methods, especially P-Tuning, produce the largest drops. LLaMA is the most stable across methods. Qwen and Gemma are intermediate, staying near zero under adapter-based PEFT but declining under P-Tuning; Gemma is slightly more sensitive than Qwen even under Prompt-Tuning. One possible explanation is architectural: prior work suggests that unlike \texttt{LLaMA2-7b} and \texttt{Mistral-7b}, whose safety neurons lie mainly in deeper layers, \texttt{Gemma-7b} concentrates them in the initial and final layers, making its safety behaviour more vulnerable to disruption \cite{chen_towards_2026}. Although only one pairwise contrast is significant, it is clear from the figure that PEFT safety effects are jointly shaped by method and base model.

\subsubsection{Fine-Tuning Parameters Effect}
\begin{figure}[t]
\captionsetup{font=small,skip=4pt}

    \centering
    \begin{subfigure}[t]{0.135\textwidth}
        \includegraphics[width=\textwidth]{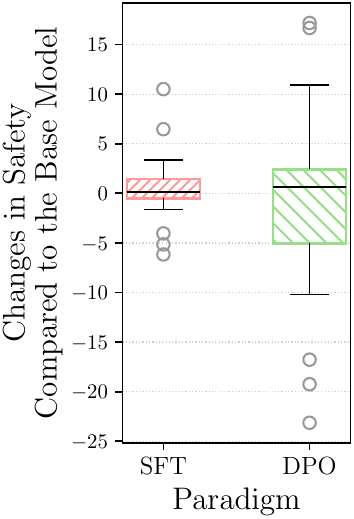}
    \end{subfigure}
    \begin{subfigure}[t]{0.11\textwidth}
        \includegraphics[width=\textwidth]{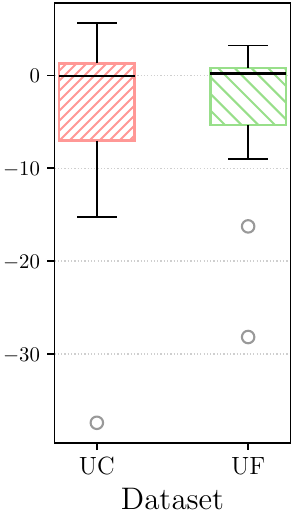}
    \end{subfigure}
    \begin{subfigure}[t]{0.11\textwidth}
        \includegraphics[width=\textwidth]{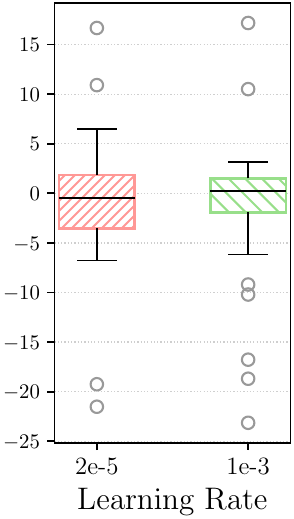}
    \end{subfigure}
    \begin{subfigure}[t]{0.11\textwidth}
        \includegraphics[width=\textwidth]{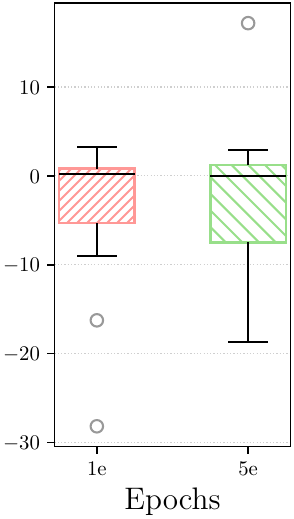}
    \end{subfigure}
     \caption{\centering Distribution of safety changes across fine-tuning configurations}
     \label{fig:safety_variables_boxplot}
\end{figure}

Fine-tuning parameters have limited impact on safety under our settings. DPO is more stable than SFT ($SD=3.45$ vs.\ $9.85$), but central tendencies remain close (median difference $<1$, mean difference $<2$). Differences by dataset, epoch count, and learning rate are likewise small (Fig.~\ref{fig:safety_variables_boxplot}).

\subsubsection{Category-Level Safety Analysis}
\begin{figure}[t]
\captionsetup{font=small,skip=4pt}

    \centering
    \begin{subfigure}[b]{0.249\textwidth}
        \centering
        \includegraphics[width=\textwidth]{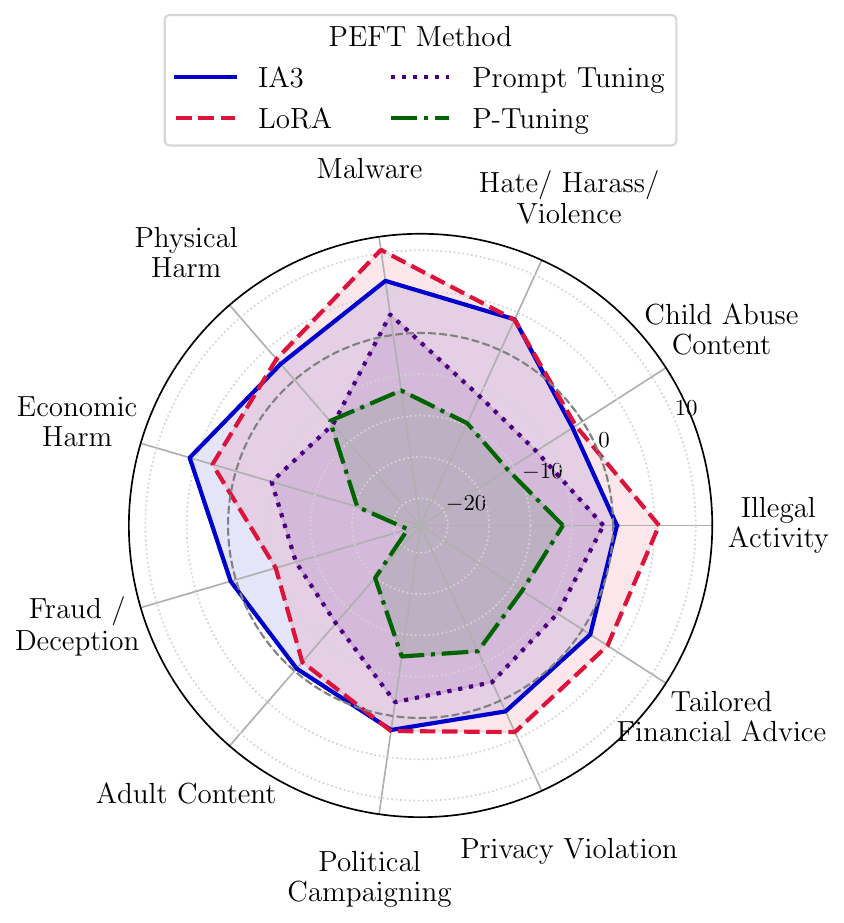}
    \end{subfigure}
    \begin{subfigure}[b]{0.249\textwidth}
        \centering
        \includegraphics[width=\textwidth]{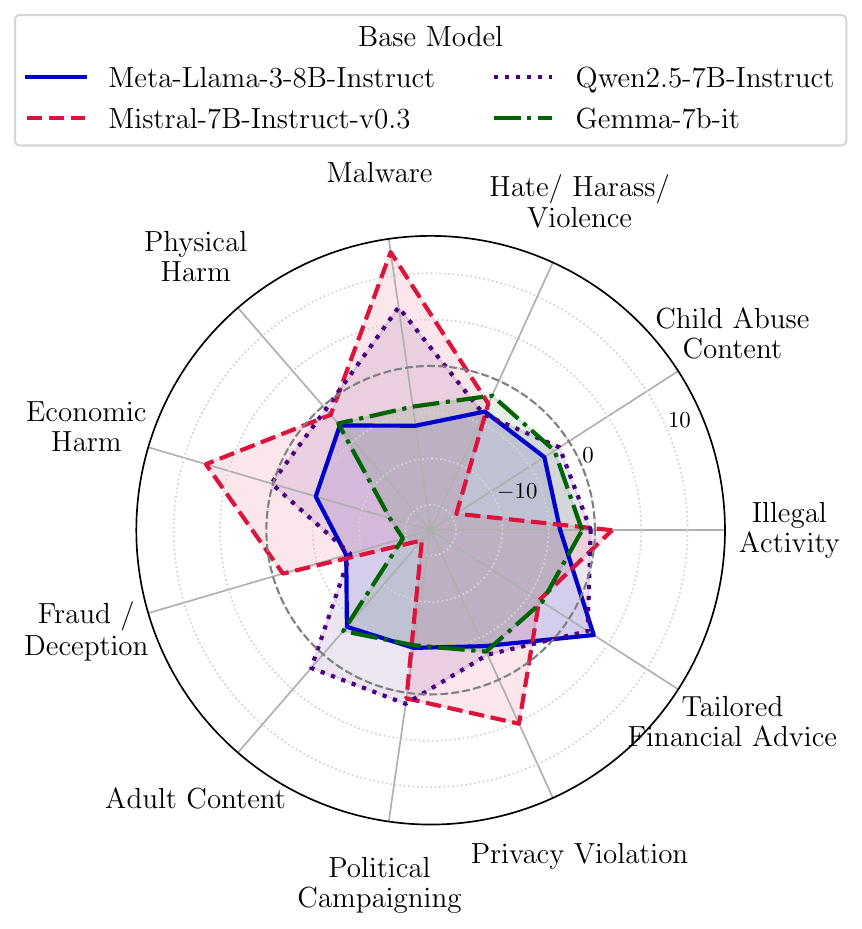}
    \end{subfigure}

    \caption{\centering Average change in safety per category for different PEFT methods (left) and different base models (right).}
    \label{fig:safety_model_method_spider}
\end{figure}

\begin{figure*}[t]
\captionsetup{font=small,skip=4pt}

    \centering
    \begin{subfigure}[b]{0.249\textwidth}
        \centering
        \includegraphics[width=\textwidth]{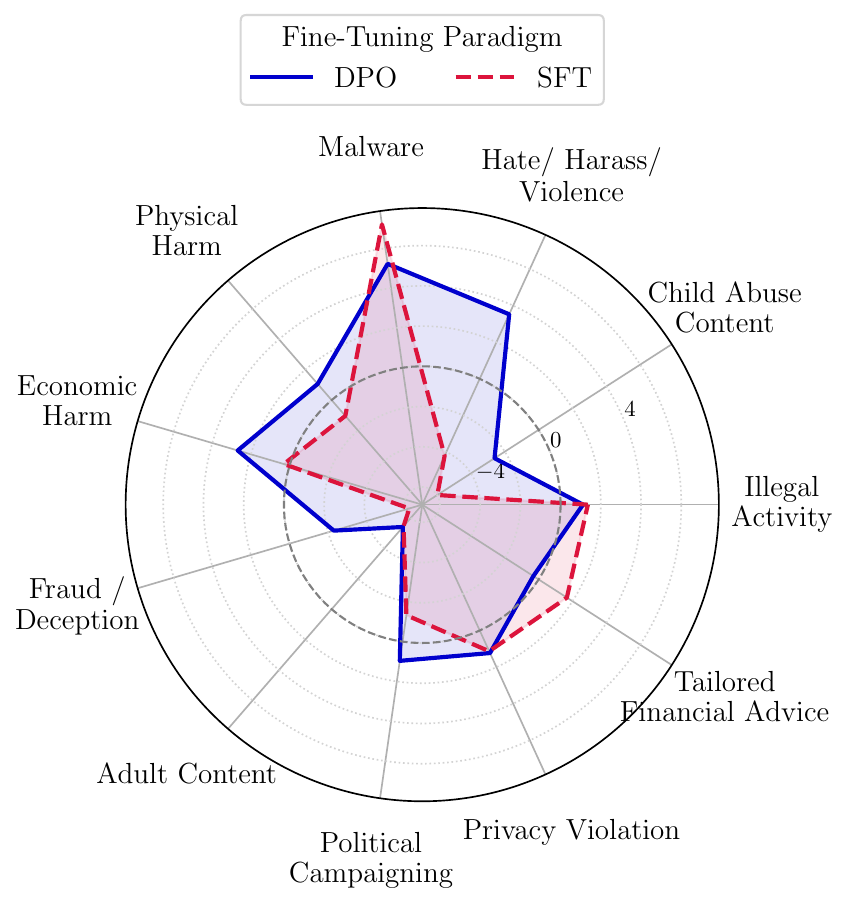}
    \end{subfigure}
    \begin{subfigure}[b]{0.249\textwidth}
        \centering
        \includegraphics[width=\textwidth]{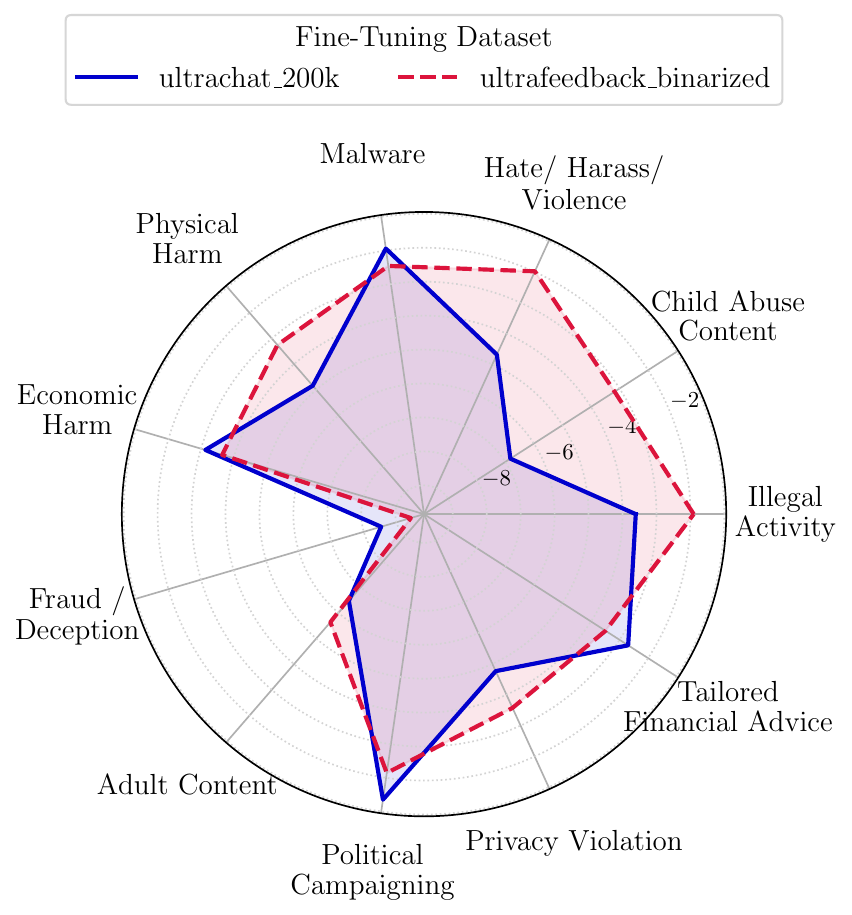}
    \end{subfigure}
    \begin{subfigure}[b]{0.249\textwidth}
        \centering
        \includegraphics[width=\textwidth]{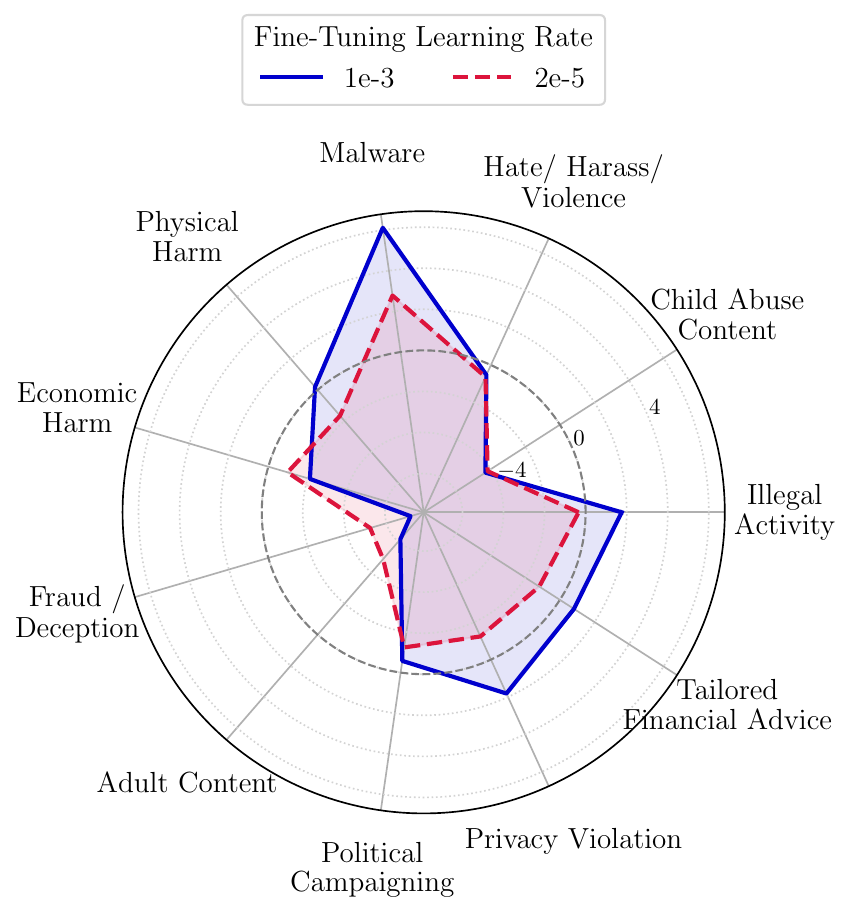}
    \end{subfigure}
    \begin{subfigure}[b]{0.249\textwidth}
        \centering
        \includegraphics[width=\textwidth]{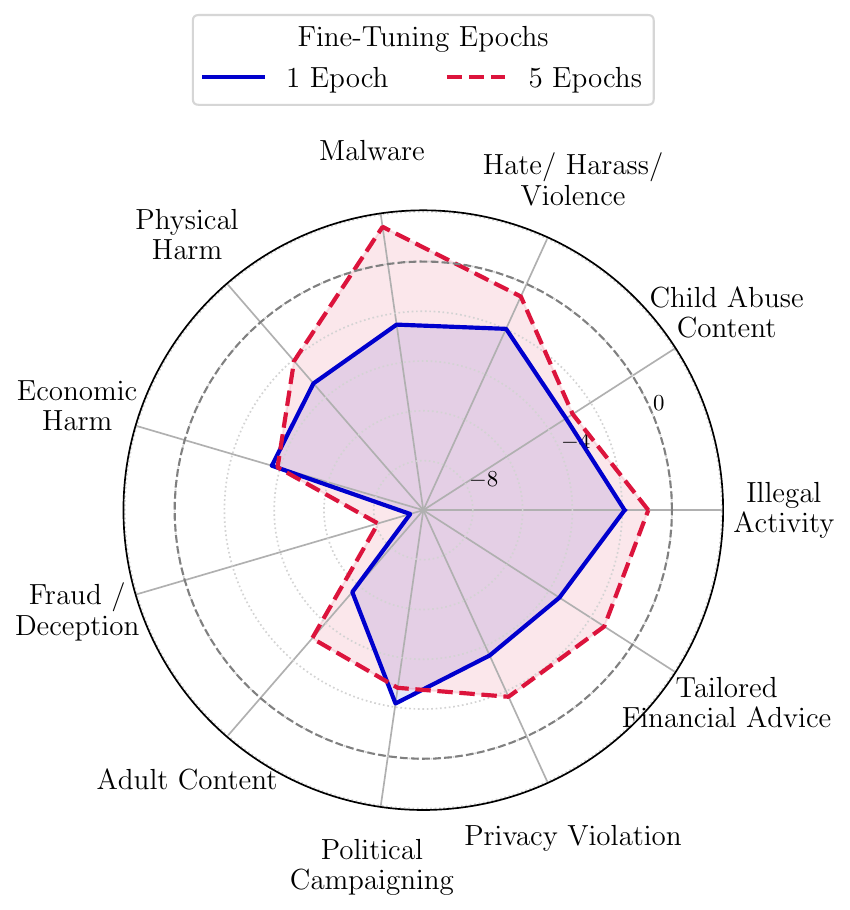}
    \end{subfigure}
     \caption{\centering Average change in safety per category for different fine-tuning variables}
     \label{fig:safety_variables_spider}
\end{figure*}

We next analyze safety changes across the eleven hazard categories using the radar plots in Figs.~\ref{fig:safety_model_method_spider} and \ref{fig:safety_variables_spider}.
Category-level results reinforce the main safety pattern while showing that some hazards are much more fragile than others. Five categories change significantly overall: \textit{Child Abuse Content}, \textit{Malware}, \textit{Fraud/Deception}, \textit{Adult Content}, and \textit{Privacy Violation}. All but \textit{Malware} decline on average, with the steepest drops in \textit{Fraud/Deception}, \textit{Adult Content}, and \textit{Child Abuse Content}. PEFT choice remains the clearest driver: P-Tuning degrades all 11 categories, Prompt-Tuning degrades seven, and both adapter-based methods improve six categories each, with IA\textsuperscript{3} usually producing the stronger gains. Base model differences are also pronounced: Gemma degrades in all 11 categories, Qwen shows mixed changes across seven, Mistral is volatile, and LLaMA remains the most robust. By contrast, training settings matter little at this level: DPO outperforms SFT in only two categories, and dataset and epoch count show no consistent effects.

Overall, category-level safety is driven much more by PEFT family and base model than by the training settings in our setup. As above, all values are changes relative to each base model rather than final safety scores.

\begin{tcolorbox}[colback=green!5,colframe=green!40!black, title=\textbf{\footnotesize Findings summary},left=1pt, right=1pt,top=1pt,bottom=1pt]
\footnotesize
    \textbf{1)} 
    Adapter-based techniques yield statistically significant safety gains, whereas prompt-based methods cause significant regressions across the majority of risk categories.\\
    \textbf{2)} 
    Safety outcomes are dominated by the base model: LLaMA is largely unaffected, Qwen improves slightly, Mistral is highly volatile, and Gemma exhibits the largest and statistically significant decline.\\
    \textbf{3)} 
    Dataset, learning rate, epoch count, and SFT versus DPO show no consistent effect; DPO has lower variance but no systematic safety gain.\\
    \textbf{4)} 
    The two categories \textit{Child Abuse Content} and \textit{Adult content}, which score among the safest on average for base models, incur the largest safety drops on average. \textit{Malware} is the only category with significant increase in safety. 
\end{tcolorbox}

\subsection{Effect of Fine-Tuning on Fairness}\label{sub:results-fairness}

Across all fine-tuning experiments, fairness deteriorates overall.
This deterioration is more pronounced in the ambiguous setting, where accuracy drops significantly (first quartile $-0.13$, i.e., 13\% more incorrect answers without context). The disambiguated setting also degrades, with lower accuracy and higher bias magnitude. Full statistical results are provided in Appendix~\ref{app1} and the replication package \cite{replicationpackage}.

\subsubsection{PEFT Method Effect}

\begin{figure*}[t]
\captionsetup{font=normal,skip=-6pt}

    \centering
    \begin{subfigure}[t]{0.238\textwidth}
        \includegraphics[width=\linewidth]{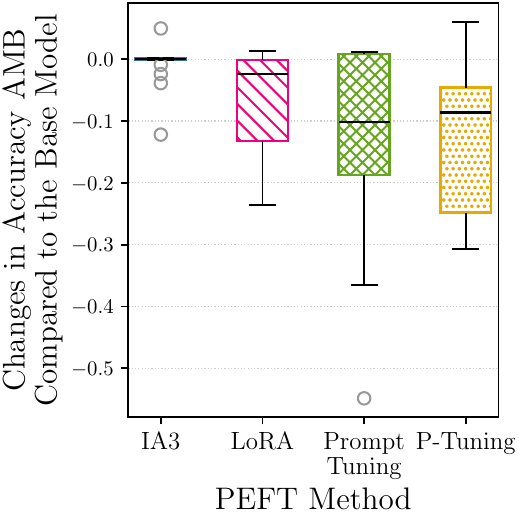}
        \label{fig:fairness_peft_method_accuracy_amb}
    \end{subfigure}
    \begin{subfigure}[t]{0.238\textwidth}
        \includegraphics[width=\linewidth]{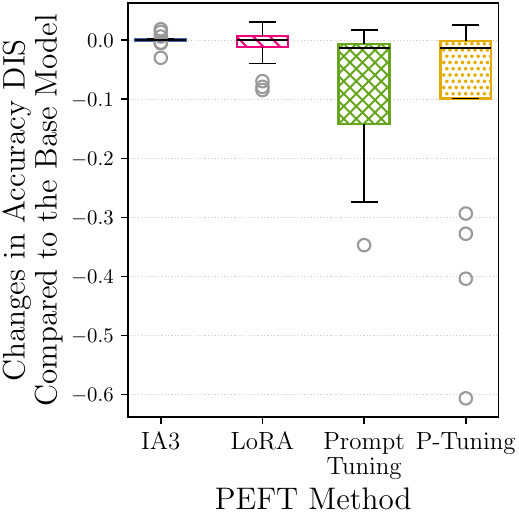}
        \label{fig:fairness_peft_method_accuracy_dis}
    \end{subfigure}
    \begin{subfigure}[t]{0.238\textwidth}
        \includegraphics[width=\linewidth]{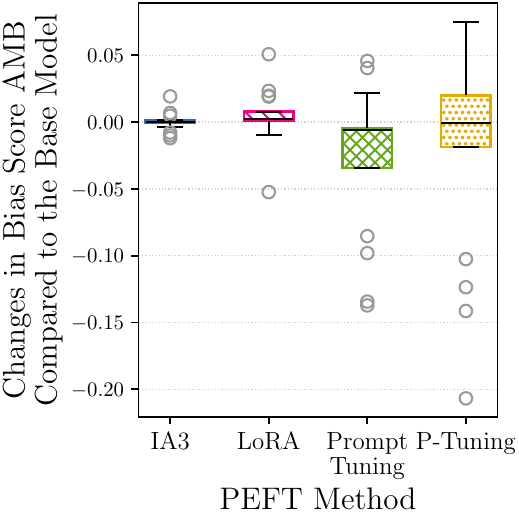}
        \label{fig:fairness_peft_method_bias_score_amb}
    \end{subfigure}
    \begin{subfigure}[t]{0.238\textwidth}
        \includegraphics[width=\linewidth]{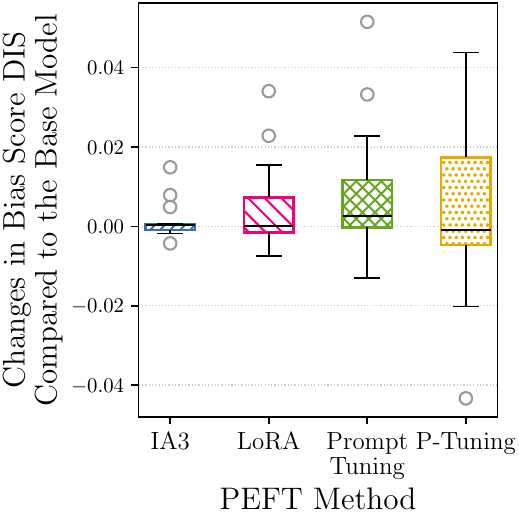}
        \label{fig:fairness_peft_method_bias_score_dis}
    \end{subfigure}
    \begin{subfigure}[t]{0.238\textwidth}
        \includegraphics[width=\linewidth]{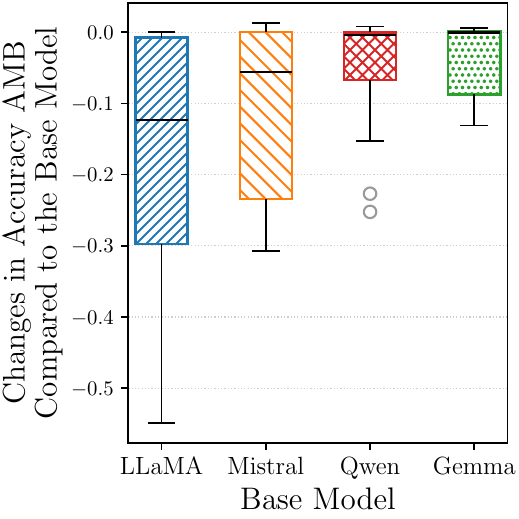}
        \label{fig:fairness_model_name_accuracy_amb}
    \end{subfigure}
    \begin{subfigure}[t]{0.238\textwidth}
        \includegraphics[width=\linewidth]{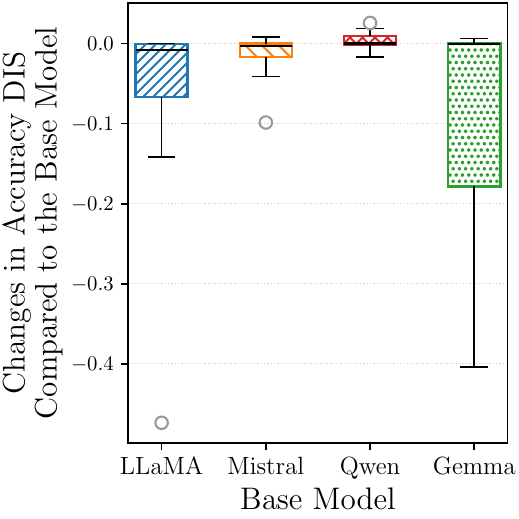}
        \label{fig:fairness_model_name_acciracy_dis}
    \end{subfigure}
    \begin{subfigure}[t]{0.238\textwidth}
        \includegraphics[width=\linewidth]{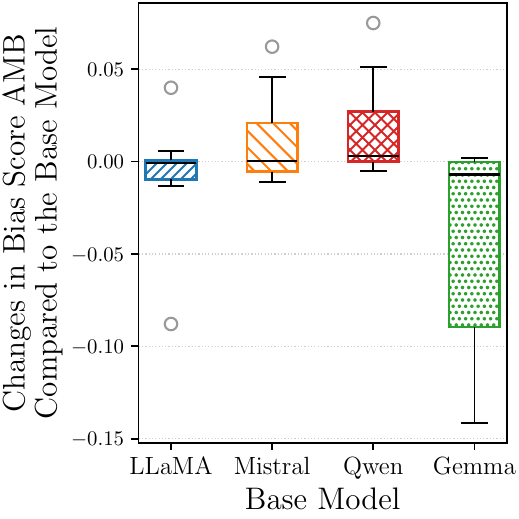}
        \label{fig:fairness_model_name_bias_score_amb}
    \end{subfigure}
    \begin{subfigure}[t]{0.238\textwidth}
        \includegraphics[width=\linewidth]{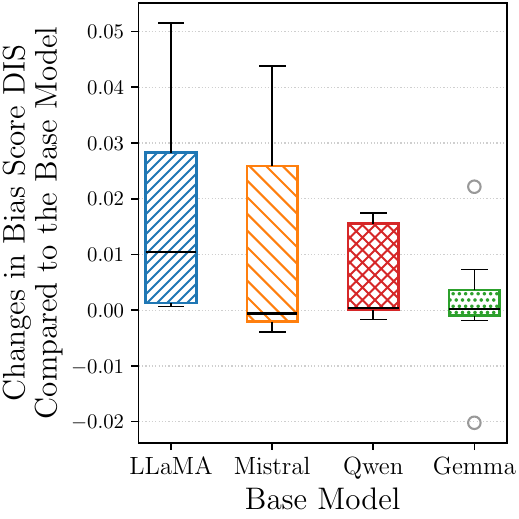}
        \label{fig:fairness_model_name_bias_score_dis}
    \end{subfigure}
    \caption{Fairness changes by PEFT method (top) and base model (bottom)}
    \label{fig:fairness_peft_method_model_name_boxplots}
\end{figure*}

Fig.~\ref{fig:fairness_peft_method_model_name_boxplots} shows the distribution of fairness metrics across different PEFT methods. 
The main pattern is that prompt-based methods reduce fairness accuracy in both contexts, while IA\textsuperscript{3} is the most robust and shows no significant degradation across fairness metrics. Method differences are clearest for accuracy rather than bias: IA\textsuperscript{3} maintains fairness accuracy better than prompt-based methods in both contexts, and better than all other methods in ambiguous contexts.

\subsubsection{Base Model Effect}

Fig.~\ref{fig:fairness_peft_method_model_name_boxplots} shows strong base-model differences. LLaMA, Mistral, and Gemma exhibit significant fairness-accuracy drops in both contexts, whereas Qwen is the most resilient (smallest interquartile ranges and no clear average shift from zero). LLaMA has the lowest first quartile in \(Acc._{\mathrm{AMB}}\) change ($-0.30$), and Gemma is the only model with a significant \(Bias_{\mathrm{AMB}}\) reduction, though this comes with a marked accuracy decline.

Model-level comparisons show significant differences for all fairness changes except \(Acc._{\mathrm{DIS}}\). The most consistent pairwise pattern is Qwen and Gemma outperforming LLaMA in \(Acc._{\mathrm{AMB}}\) and \(Bias_{\mathrm{DIS}}\) change. Because Qwen also starts from the strongest ambiguous-context base performance, it remains the most robust overall fairness choice in our setup.

\subsubsection{Joint Effect of PEFT Method and Base Model on Fairness}
\begin{figure*}[t]
\captionsetup{font=normal,skip=0pt}

    \centering
    \includegraphics[width=0.45\textwidth]{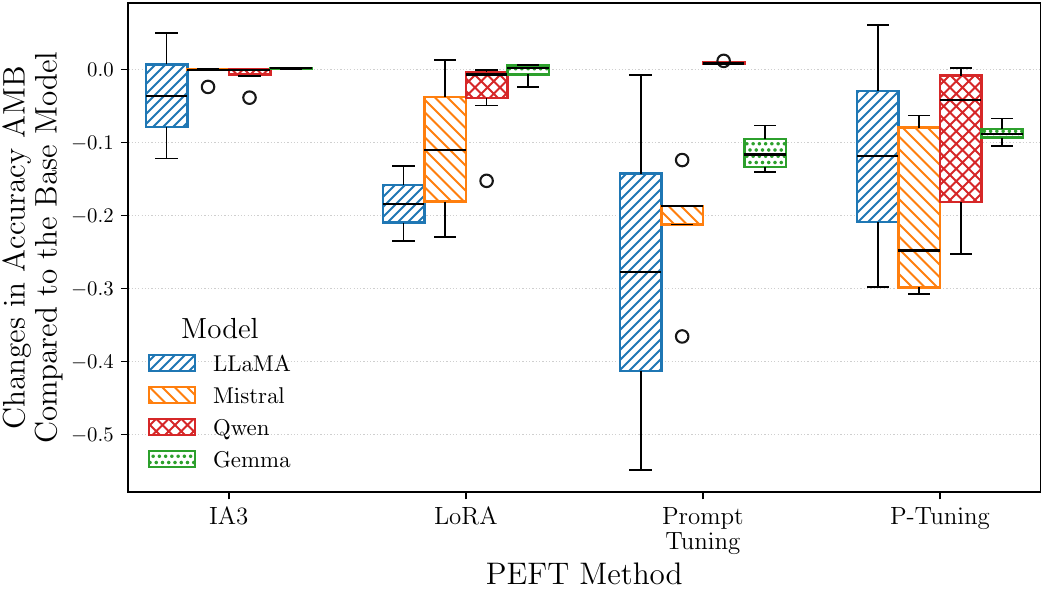}
    \includegraphics[width=0.45\textwidth]{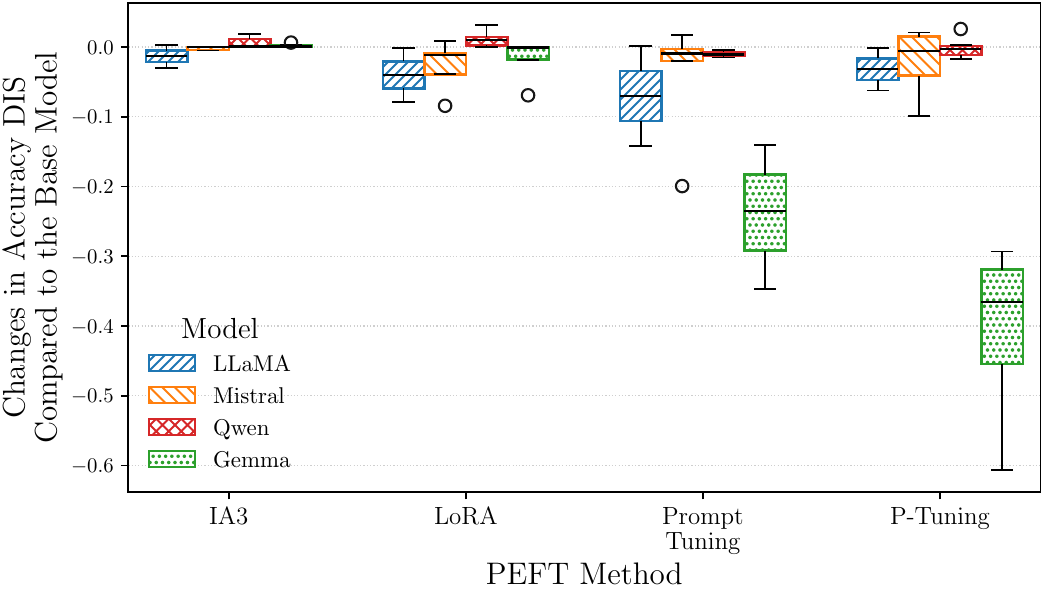}
    \includegraphics[width=0.45\textwidth]{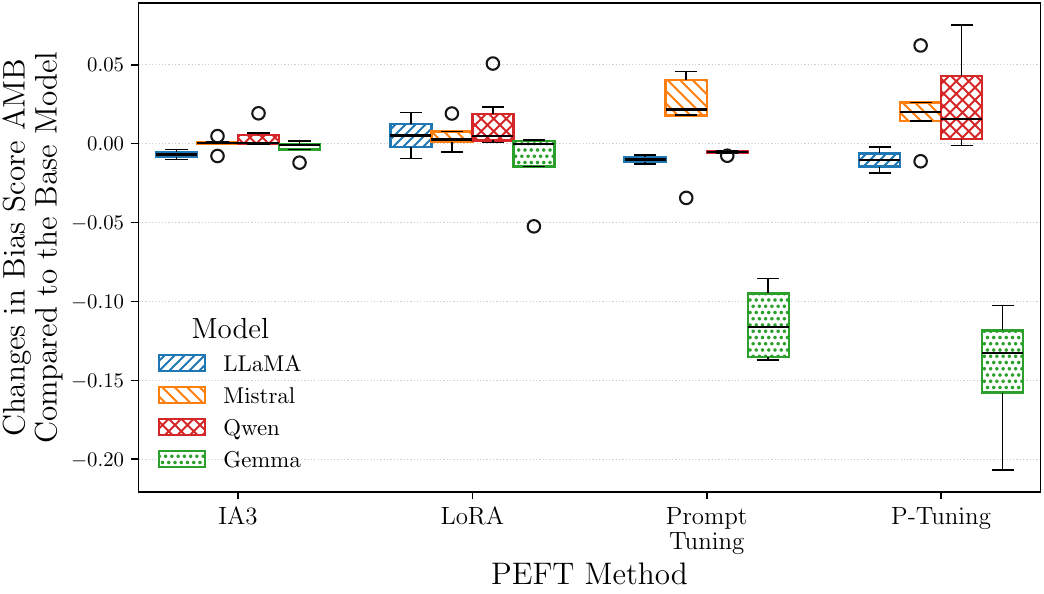}
    \includegraphics[width=0.45\textwidth]{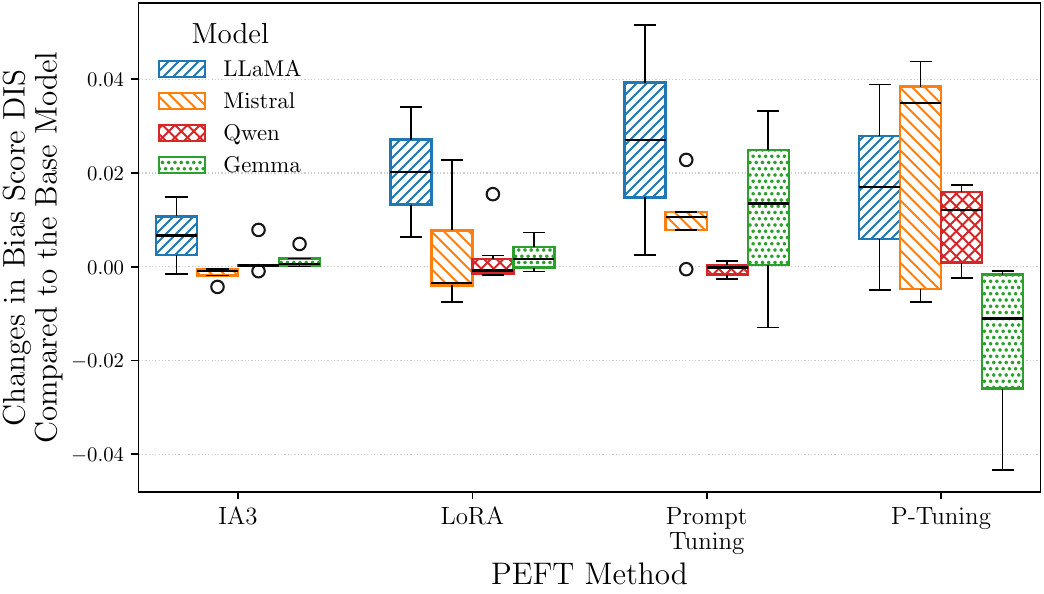}
    \caption{Distribution of changes in  fairness metrics for each combination of PEFT method and base model}
    \label{fig:fairness_peft_method_by_model_name_boxplots}
    
\end{figure*}

Fig.~\ref{fig:fairness_peft_method_by_model_name_boxplots} shows the same interaction pattern as the aggregate results. IA\textsuperscript{3} is generally the most stable across Gemma, Mistral, and Qwen, whereas the weakest combinations are usually prompt-based methods on LLaMA or Mistral. Some mixed cases remain, such as P-Tuning on Gemma reducing \(Acc._{\mathrm{DIS}}\) while also lowering bias amplitude, but the dominant pattern is unchanged: adapter-based PEFT is more stable across base models, and prompt-based PEFT produces the largest fairness regressions.

\subsubsection{Effect of Fine-Tuning Parameters on Fairness}

Among training variables, paradigm has the clearest effect: DPO is generally less harmful to fairness than SFT.
A lower learning rate ($2\times10^{-5}$) also tends to reduce fairness degradation relative to $10^{-3}$. By contrast, epoch count and dataset choice show no consistent effects across fairness metrics.

\subsubsection{Category-Level Fairness Analysis}
Category-level fairness results echo the aggregate pattern while showing that some demographic dimensions are more vulnerable than others. Across the nine categories, fairness accuracy declines significantly in nearly all cases, with the largest drops in \textit{Sexual Orientation} and \textit{Nationality}. In the disambiguated setting, \textit{Disability Status} and \textit{Race/Ethnicity} are the clearest cases where accuracy and bias worsen together, whereas \textit{Age} is the least affected category in ambiguous contexts. Categories that start out fairest also tend to deteriorate the most after tuning. Method effects remain strongest: IA\textsuperscript{3} is the most robust at the category level, LoRA is next, and the clearest joint drops in accuracy and bias occur for prompt-based methods, especially on \textit{Race/Ethnicity}. Base-model differences are also pronounced: Mistral shows the broadest ambiguous-context regressions, Qwen is the most resilient overall, and Gemma shows the largest disambiguated decline versus its base. By contrast, training settings again matter little: dataset and epoch count show only sporadic differences, a lower learning rate mainly helps \(Acc._{\mathrm{AMB}}\), and DPO has the clearest parameter-level advantage over SFT.
Detailed category-level fairness profiles by base model and PEFT method are provided in the replication package.

\begin{tcolorbox}[colback=green!5,colframe=green!40!black, title=\footnotesize\textbf{Findings summary},left=1pt, right=1pt, top=1pt, bottom=1pt]
\footnotesize
    \textbf{5)} 
    Adapter-based methods are least disruptive. IA\textsuperscript{3} best preserves accuracy and has the lowest variance, LoRA is next, whereas Prompt-Tuning and especially P-Tuning reduce accuracy; none consistently reduces bias.\\
    \textbf{6)} 
    Qwen and Gemma preserve higher \(Acc._{\mathrm{AMB}}\) than LLaMA, while Mistral and LLaMA show the largest accuracy losses. Gemma’s \(Bias_{\mathrm{AMB}}\) gain comes with lower accuracy, whereas Qwen is the most resilient overall.\\
    \textbf{7)} 
    DPO is slightly fairer than SFT, and a lower learning rate helps \(Acc._{\mathrm{AMB}}\). Dataset and epoch count show no consistent effect.\\
    \textbf{8)} 
    Accuracy declines are broad but largest for \textit{Sexual Orientation} and \textit{Nationality}. Only \textit{Disability Status} and \textit{Race/Ethnicity} show simultaneous regressions in \(Acc._{\mathrm{DIS}}\) and \(Bias_{\mathrm{DIS}}\). Categories that were fairest in the base models tend to deteriorate most after tuning. 
\end{tcolorbox}

\subsection{Correlation Analysis}

To better understand the interaction between utility, safety, and fairness under PEFT fine-tuning, we conducted a correlation analysis using Spearman's rank correlation.

\subsubsection{Overall Trends}
\begin{figure}[t]
    \centering
    \includegraphics[width=0.62\linewidth]{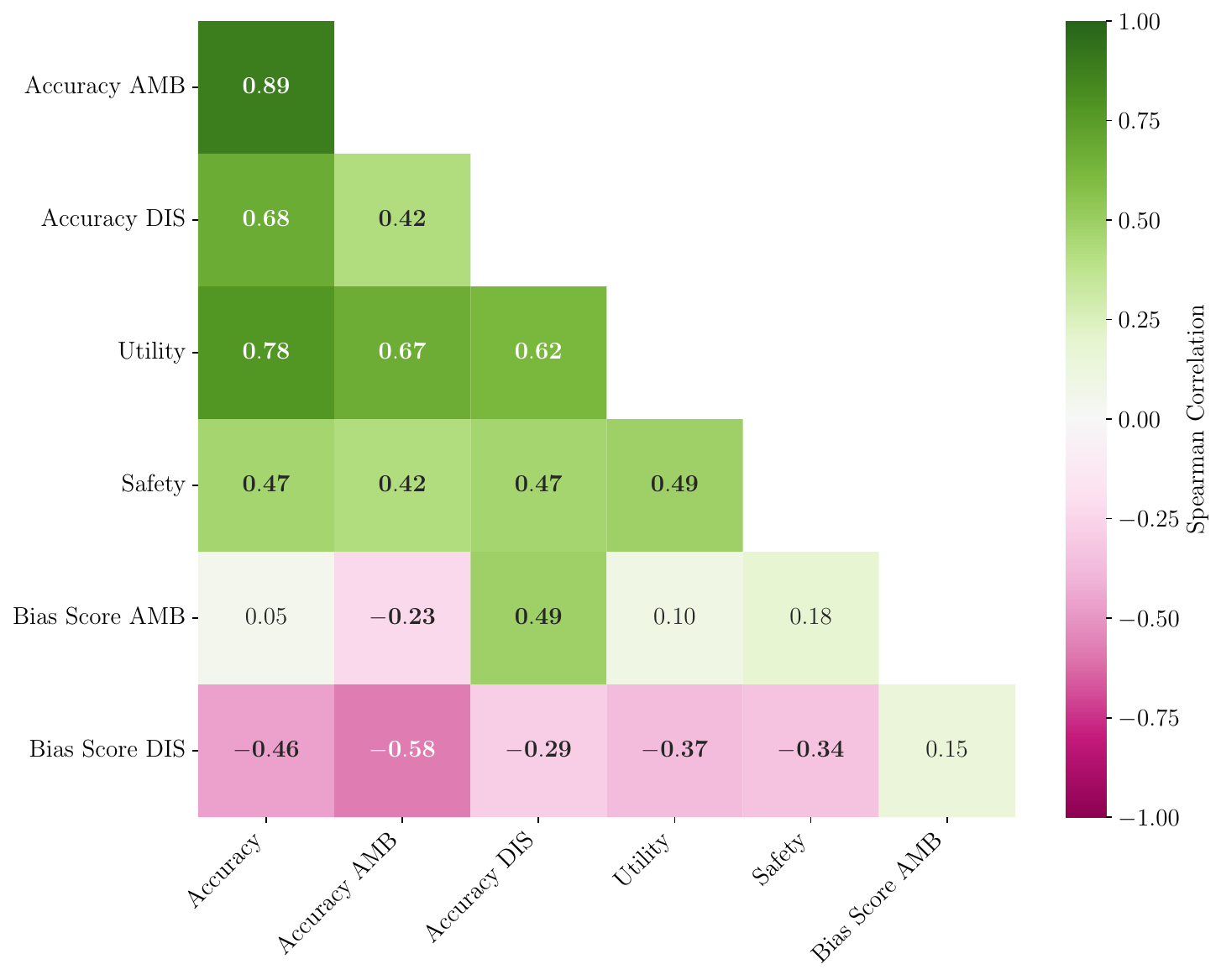}
    \caption{\centering Overall correlations; significant values are \textbf{bold}.}
    \label{fig:heatmap_global}
\end{figure}

Fig. \ref{fig:heatmap_global}  illustrates the correlation relationships for the changes in metrics in overall results. A core cluster of metrics—total accuracy, \(Acc._{\mathrm{AMB}}\), \(Acc._{\mathrm{DIS}}\), utility, and safety—show strong positive correlations with one another. We will refer to these 5 metrics as the \textbf{core positive group}. These are alignment dimensions we generally seek to maximize, and their mutual reinforcement is an encouraging outcome. Furthermore, these metrics are negatively correlated with \(Bias_{\mathrm{DIS}}\), which we aim to minimize, suggesting that improvements in performance and safety often accompany reductions in bias in this context.
By contrast, \(Bias_{\mathrm{AMB}}\) shows no significant correlation with this group, implying that it is harder to optimize jointly with the other objectives. Its negative correlation with \(Acc._{\mathrm{AMB}}\) is expected from the metric definition (Section \ref{subsec:methodology_fairness}).

\subsubsection{Trends by PEFT Method}
By PEFT method, the core accuracy--utility structure remains broadly intact, but safety and \(Bias_{\mathrm{DIS}}\) become method-dependent. Prompt-based methods preserve positive safety alignment with the accuracy-utility cluster, whereas adapter-based methods, especially IA\textsuperscript{3}, show weaker or negative safety correlations, suggesting a safety-utility trade-off. Likewise, the usual inverse link between \(Bias_{\mathrm{DIS}}\) and the core group weakens across methods and remains clearest only for LoRA and Prompt-Tuning.

\subsubsection{Trends by Base Model}
By base model, the global pattern is mostly preserved for LLaMA, Mistral, and Gemma, but Qwen and Gemma show the clearest deviations. Mistral is the most internally coherent, with strong positive correlations within the core group and negative correlations with both bias scores, especially \(Bias_{\mathrm{DIS}}\), suggesting no fairness--safety trade-off. LLaMA largely follows the same directions, though more weakly. By contrast, Qwen breaks the usual link between \(Acc._{\mathrm{AMB}}\) and \(Acc._{\mathrm{DIS}}\) and is the only model whose two bias scores move together, while Gemma shows the opposite concern: bias tends to rise with the core positive group, especially in ambiguous contexts, suggesting that gains in alignment may coincide with higher bias.

\vspace{-0.5em}
\begin{tcolorbox}[colback=green!5,colframe=green!40!black, title=\textbf{\footnotesize Findings summary},left=1pt, right=1pt, top=1pt, bottom=1pt]
\footnotesize
  \textbf{9)} Accuracy, utility, and safety are generally mutually reinforcing and inversely related to \(Bias_{\mathrm{DIS}}\).\\
  \textbf{10)} \(Bias_{\mathrm{AMB}}\) aligns weakly with the other metrics, suggesting targeted mitigation.\\
  \textbf{11)} Prompt-based methods align utility and safety better, but may trade off fairness, especially in ambiguous contexts.\\
  \textbf{12)} Gemma shows the clearest trade-off, becoming more biased as it becomes safer and more useful; Mistral shows the opposite.
\end{tcolorbox}
\vspace{-0.5em}

\subsection{Cross-Task Robustness Check: CodeAlpaca}\label{subsec:results-codealpaca}

We report the CodeAlpaca extension here as a focused cross-task check. As in the main analysis, results are reported as changes relative to the corresponding base model and, where relevant, compared with matched conversational configurations. Detailed statistical results are reported in the replication package.

Relative to the base models, benign coding fine-tuning still changes alignment in meaningful ways. Overall safety and \(Acc._{\mathrm{AMB}}\) decreases significantly, and \(Bias_{\mathrm{DIS}}\) increases significantly. The change in \(Acc._{\mathrm{DIS}}\) is negative on average but not significant. In other words, even under conservative PEFT settings, a benign coding dataset is sufficient to shift both safety and fairness away from the base-model baseline.

When we compare the CodeAlpaca configurations with their matched  conversational counterparts, the task-level contrast is much smaller. Only a small minority of the paired tests are significant, none of them concern overall safety, and the only significant overall difference is a slightly less negative change in \(Acc._{\mathrm{DIS}}\) for CodeAlpaca. The remaining differences are isolated category-level fairness effects rather than a broad task-wide shift.

The qualitative PEFT ranking nevertheless remains similar: adapter-based methods usually stay closer to zero, whereas prompt-based methods account for the largest safety losses and fairness regressions.

\begin{tcolorbox}[colback=green!5,colframe=green!40!black, title=\textbf{\footnotesize Findings summary},left=1pt, right=1pt, top=1pt, bottom=1pt]
\footnotesize
\textbf{13)} CodeAlpaca fine-tuning still shifts alignment relative to the base models, lowering overall safety and fairness accuracy.\\
\textbf{14)} Compared with matched conversational runs, CodeAlpaca shows no broad task-wide safety difference, and adapter-based methods remain safer than prompt-based ones.
\end{tcolorbox}

\section{Discussion}\label{_discussion}
\subsection{Impact of PEFT Method and Base Model on Safety and Fairness}

Our results show a consistent interaction between PEFT family and base model. Adapter-based methods (LoRA and IA\textsuperscript{3}) usually preserve, or modestly improve, safety and fairness accuracy, whereas prompt-based methods more often degrade both, likely because adapters leave core alignment constraints more intact than prompt-based updates. Adapters modify only a small set of low-rank weights, whereas prompt-based methods alter the input representation more directly and may therefore bypass or weaken existing safeguards. Base-model choice then determines how severe these shifts become: Qwen is the most robust overall, LLaMA is safety-stable but fairness-fragile, Gemma shows the clearest trade-off with lower accuracy and the largest safety decline, and Mistral is the most volatile, consistent with the absence of built-in safety alignment noted in its model card \cite{noauthor_mistralaimistral-7b-instruct-v03_nodate}\cite{mistral_documentation}. In practice, no single model--method pair optimizes all objectives, so practitioners should start from a well-aligned base model, prefer adapter-based PEFT, and choose trade-offs explicitly.

\subsection{Fine-Tuning Configurations Play a Secondary Role}
Hyper-parameters (learning rate, epoch count) and high-level training paradigm (SFT vs.\ DPO) produced only sporadic, category-specific effects.  DPO offered marginal fairness advantages over SFT, but none of these settings rivalled the impact of PEFT method or base model.  The targeted sensitivity analysis reported in Appendix~\ref{app6} reaches the same interpretation for method-specific PEFT settings: LoRA rank, prompt length, and IA\textsuperscript{3} target-module choice shift the magnitude of safety--fairness trade-offs, but do not reverse the broad PEFT-family patterns. Paired tests also provide limited evidence of stable setting-level differences, reinforcing that PEFT family and base model remain first-order drivers.

\subsection{What Changes Under a Coding Task?}
The CodeAlpaca extension separates two questions: whether PEFT-induced alignment shifts are peculiar to conversational tuning, and whether a task change fundamentally reorders PEFT risks. Our evidence suggests ``no'' to the first and ``not much'' to the second. Benign code-oriented fine-tuning still reduces overall safety and fairness accuracy relative to the base models, so the risk is not limited to dialogue data. Yet paired comparisons with matched conversational runs show limited task-level differences and no significant safety difference.

\subsection{Interplay Between Safety, Fairness and Utility}
Across the full experiment matrix, the \textit{core positive group}, i.e. overall accuracy, ambiguous/disambiguated accuracy, utility and safety, remained positively correlated.  \(Bias_{\mathrm{DIS}}\) was negatively correlated with that group, while \(Bias_{\mathrm{AMB}}\) was uncorrelated.  In practice, this means that one can often raise safety without sacrificing task quality, but fairness in ambiguous prompts drifts independently and requires specific mitigation.

\subsection{Granular Behaviour Across Safety and Fairness Categories}
Category-level results show that PEFT does not affect all harms uniformly. The most vulnerable safety and fairness categories regress more sharply than aggregate scores suggest, and prompt-based methods amplify these weak points more than adapter-based methods. These patterns argue against one-size-fits-all alignment: category-specific audits and post-tuning mitigation should be standard practice.

\subsection{Research Significance and Take-Away Messages}
\paragraph{Insights from our study}
Our analysis highlights four main patterns:

\begin{itemize}
    \item \textbf{Adapter-based PEFT methods are generally safer and fairer than prompt-based methods.} LoRA and IA\textsuperscript{3} generally outperform Prompt-Tuning and P-Tuning in maintaining both safety and fairness.
    \item \textbf{Base model characteristics strongly influence outcomes.} Robust models (e.g., Qwen) retain alignment better through fine-tuning, whereas more fragile models (e.g., Gemma) are more prone to regressions.
    \item \textbf{Fine-tuning hyperparameters play a secondary role.} Learning rate, epoch count, and SFT versus DPO matter less than PEFT family and base model, though DPO is the most consistently favorable setting. The PEFT-setting sensitivity analysis in Appendix~\ref{app6} reaches the same conclusion: these configuration choices can shift the size of safety--fairness trade-offs, but they do not overturn the broader method- and model-level patterns.
    \item \textbf{The main risk patterns generalize beyond conversation tuning.} The coding-task extension still shows alignment drift relative to the base models, but only modest differences relative to matched conversational runs.
\end{itemize}
\paragraph{Key take-aways for practitioners}
\begin{itemize}
    \item \textbf{Start with aligned models, fine-tune with adapter-based methods.}  
          Select a base model that passes basic refusal checks and favour LoRA/IA\textsuperscript{3}.  
    \item \textbf{Stress-test vulnerable categories.}  
          Overall, Child abuse and adult content in safety, and sexual-orientation, and nationality in fairness, are the earliest to regress.  
    \item \textbf{Monitor ambiguous bias separately.}  
          Improvements in disambiguated bias do not guarantee fairness in real-world settings, where there is usually no context accompanying the prompts.  
    \item \textbf{Adopt multi-objective PEFT fine-tuning.}
        Extend existing multi-objective fine-tuning methods to the PEFT setting to jointly optimize utility, safety, and fairness. Techniques such as constrained or Pareto-front optimization, as explored in prior work \cite{ren_cos-dpo_2025, li_gradient-adaptive_2025, mukherjee_multi-objective_2024, dai_safe_2024}, can be adapted to balance trade-offs between competing alignment goals while maintaining model performance.

\end{itemize}

\subsection{Future Research Directions}
Our study shows that small, efficient weight updates can still induce large alignment shifts, so future work should move beyond performance benchmarks and seek PEFT strategies with explicit safety and fairness guarantees, for example through multi-objective optimization of utility, safety, and fairness, especially for prompt-based methods.

A complementary breadth-oriented direction is to fix conservative PEFT settings and evaluate them on additional task families (e.g., reasoning, role-play, and code-oriented data) to better characterize cross-task generalizability. Our CodeAlpaca extension is a first step, but it should not be over-read as a full task generalization study: it covers a single coding dataset and uses only conservative SFT settings. 

Another direction is to characterize architectural factors that make some base models more alignment-preserving under PEFT than others.

Because prompt-based methods consistently underperform in preserving safety and fairness in our setup, it is also important to test alternative initialization strategies and prompt-template designs, as appended prompt content can materially affect safety behaviour \cite{lyu_keeping_2024, jiang_chatbug_2025, peng_navigating_2024}, especially when extra unrelated tokens are introduced.

\section{Threats to Validity}\label{_threats}
\paragraph{Internal Validity}
Implementation choices may have influenced our outcomes. Fine-tuning was conducted with fixed hyperparameters based on common LoRA settings from HuggingFace, and alternative choices (e.g., rank, scaling factor, dropout) could yield different alignment trajectories. For prompt-based methods, we also used one representative configuration profile; alternative prompt length, initialization, or template design may shift outcomes. Our targeted sensitivity analysis shows that PEFT method settings can change the size of safety--fairness shifts, but do not reverse the broad PEFT-family conclusions within the tested alternatives. \texttt{Gemma-7B-it} could not be tuned with DPO due to technical constraints, reducing comparability across paradigms. Ten models with \texttt{NaN} or \texttt{inf} logits and nineteen utility outliers were removed; if these failures correlate with unsafe or unfair behaviour, worst-case degradation may be underestimated, though severely degraded models are unlikely to be used in practice. Safety was scored automatically using \texttt{LLaMA-Guard 2} rather than human annotators; while it shows strong agreement with expert ratings and is comparable to other top guardrail models \cite{bassani_guardbench_2024,vidgen2024modelbench}, automatic evaluation introduces potential measurement error. \texttt{LLaMA-Guard 2} was the best-performing open-source guardrail at the time of this study in 2024.

\paragraph{Construct Validity}
Safety was evaluated using the HEx-PHI prompt set with \texttt{LLaMA-Guard 2} across 11 hazard categories, and fairness was measured with a modified BBQ-Lite benchmark; hazards and bias categories not represented in these resources remain untested. Because our conclusions rely on one automated safety judge and one fairness benchmark variant, some observed differences may be sensitive to evaluator or benchmark choice. A targeted GPT-4.1 judge check shows similar unsafe rates and preserves the broad method ordering. Utility was assessed on 100 held-out prompts per dataset using single-turn \texttt{GPT-4o} scoring, which may not capture interactive behaviour. Additionally, fine-tuning datasets can influence alignment outcomes; to mitigate this, we selected widely used, cleaned conversational datasets (\textit{UltraChat} and \textit{UltraFeedback}) previously employed in training Zephyr \cite{tunstall2023zephyr}, though residual biases may still affect the behaviour of fine-tuned models.

\paragraph{External Validity}
The main study still centers on four instruction-tuned 7-8B models and two widely used conversational datasets, so its strongest claims remain about benign conversational PEFT. Results may differ for larger or domain-specific models, other PEFT methods (e.g., Prefix-Tuning), or datasets with distinct linguistic characteristics. We partially broaden this scope through the CodeAlpaca extension, which shows that the main alignment-risk patterns are not confined to conversational data. However, this extension covers only one non-conversational task family, uses conservative SFT settings only, and does not include a task-matched utility benchmark. Results may therefore still differ for other non-conversational settings such as reasoning-heavy, role-play, or tool-use tasks. Fine-tuning was performed on 10\% of UltraChat and 34\% of UltraFeedback due to computational constraints, so outcomes may not fully generalize to full-scale training. Model--method combinations were selected based on their frequency of appearance on HuggingFace, leaving many real-world pairings unexplored.

\paragraph{Conclusion Validity}
All statistical tests were non-parametric, as no metric passed the Shapiro–Wilk normality test; while appropriate, these tests have lower power, increasing the risk of false negatives. Multiple pairwise comparisons were Bonferroni-adjusted, though the large number of tests may still inflate the family-wise error rate. Averaging three repeats per setting reduces variance but can obscure sporadic training instabilities, so additional runs with different random seeds would strengthen the robustness of our conclusions.

\section{Conclusion}\label{_conclusion}
This work offers a systematic, multi-model assessment of how PEFT affects safety and fairness in instruction-tuned LLMs. Across models, methods, and settings, we find that PEFT can materially reshape alignment and sometimes increase the risk. Adapter-based methods (LoRA, IA\textsuperscript{3}) generally preserve or even improve safety and fairness, whereas prompt-based methods (Prompt-Tuning, P-Tuning) more often degrade them. Risk is also base-model dependent: while LLaMA in safety and Qwen in fairness are comparatively robust, Mistral and Gemma are notably vulnerable, and models subjected to the same regimen can behave very differently. A category-level view reveals especially fragile areas (e.g., child abuse and adult content for safety, and sexual orientation and nationality for fairness) which highlights that aggregate scores may hide important degradations. The CodeAlpaca extension shows that these shifts are not confined to conversational tuning, while matched task-level differences remain limited. We therefore advocate treating alignment as a first-class objective when selecting PEFT strategies, pairing performance and efficiency gains with fine-grained evaluation to guard against safety and fairness regressions.

\appendices

\section{Modifications of the BBQ-Lite Benchmark}\label{app2}

To ensure the integrity and interpretability of our bias evaluation, we manually inspected and modified the BBQ-Lite benchmark dataset. Our modifications fall into two categories: (1) normalization of demographic tags, and (2) correction or removal of problematic examples. In total, we identified 232 examples requiring intervention, 32 were corrected, and 200 were excluded from the analysis.

\subsection{Demographic Tag Normalization}

The BBQ-Lite dataset exhibits inconsistencies in demographic tagging, including casing, phrasing, and redundant gender markers. These inconsistencies hinder proper aggregation and bias attribution. We applied the following normalization procedures:

\begin{itemize}
    \item \textbf{Gender Identity Tags:} Terms such as \texttt{woman}, \texttt{female}, \texttt{girl}, and \texttt{F} were unified under the label \texttt{F}, while \texttt{man}, \texttt{male}, \texttt{boy}, and \texttt{M} were mapped to \texttt{M}. Transgender references (e.g., \texttt{transgender men}, \texttt{transgender women}) were normalized as \texttt{trans}.
    \item \textbf{Race / Ethnicity Tags:} Terms like \texttt{african} and \texttt{african american} were treated distinctly and consistently capitalized.
    \item \textbf{SES Tags:} Labels such as \texttt{low ses} and \texttt{high ses} were standardized to \texttt{lowSES} and \texttt{highSES}, respectively.
    \item \textbf{Prefix/Suffix Removal:} Gender-related prefixes (e.g., \texttt{M-black}) and suffixes (e.g., \texttt{latino\_F}) were stripped to decouple primary identity attributes from redundant gender tagging.
    \item \textbf{Special Tag Handling:} Specific terms like \texttt{nontransgender} and \texttt{nonold} were remapped to \texttt{nonTrans} and \texttt{nonOld} for consistency across the dataset.
\end{itemize}

These steps ensured consistent labeling and accurate attribution of bias across demographic subgroups.

\subsection{Correction of Answer Metadata}

We identified 32 examples where the \texttt{answer\_info} fields were incorrect or inconsistent with the actual answer content and metadata. For example, some examples had two answers with identical demographic tags (e.g., both tagged as \texttt{male}), or tags that did not align with the context or answer text. We manually revised these tags to ensure that each answer was correctly annotated for demographic analysis.

\subsection{Exclusion of Invalid Examples}

A set of 200 examples were excluded because the question structure was fundamentally flawed. Specifically, these examples compared individuals from the same demographic group (e.g., man (male) vs. boy (male)), or two individuals from stereotyped groups (e.g. Black vs. Latino), violating the benchmark’s intent to assess bias between stereotyped and non-stereotyped groups. Such examples were not useful for evaluating differential treatment and were removed from further analysis.

\subsection{Summary of Changes}

Table~\ref{tab:bbq-modifications-full} provides a detailed summary of all modified or removed examples, categorized by the type of change and demographic group involved.

\begin{table*}[]
\small
\centering
\caption{Detailed Summary of Modifications to BBQ-Lite Examples}
\label{tab:bbq-modifications-full}
\begin{tabular}{llllr}
\hline
\textbf{Type}   & \textbf{Category}   & \textbf{ID} & \textbf{Descriptions}                                            & \multicolumn{1}{l}{\textbf{Count}} \\ \hline
Tag Correction  & Gender Identity     & 48-51                       & 2 answers incorrectly tagged as `male'                        & 4                                  \\ \hline
Tag Correction  & Physical App. & 576-579                     & 2 answers incorrectly tagged as `negDress'        & 4                                  \\
Tag Correction  & Physical App. & 588-591                     & 2 answers incorrectly tagged as `negDress'        & 4                                  \\
Tag Correction  & Physical App. & 608-611                     & 2 answers incorrectly tagged as `negDress'        & 4                                  \\
Tag Correction  & Physical App. & 620-623                     & 2 answers incorrectly tagged as `negDress'       & 4                                  \\
Tag Correction  & Physical App. & 656-659                     & 2 answers incorrectly tagged as `notPregnant'                 & 4                                  \\ \hline
Tag Correction  & Race / Ethnicity    & 448-451                     & 2 answers incorrectly tagged as `black'           & 4                                  \\
Tag Correction  & Race / Ethnicity    & 544-547                     & 2 answers incorrectly tagged as `black'          & 4                                  \\ \hline
Removed & Gender Identity     & 284-287                     & Same-group comparison (`male' vs. `male')        & 4                                  \\
Removed & Gender Identity     & 300-303                     & Same-group comparison (`female'  vs. `female') & 4                                  \\
Removed & Gender Identity     & 304-307                     & Same-group comparison (`male' vs. `male')        & 4                                  \\
Removed & Gender Identity     & 320-323                     & Same-group comparison (`female' vs. `female') & 4                                  \\ \hline
Removed & Race / Ethnicity    & 1032-1055                   & Both stereotyped comparison (Black - Latino)                     & 24                                 \\
Removed & Race / Ethnicity    & 1080-1087                   & Both stereotyped comparison (Black - Latino)                     & 8                                  \\
Removed & Race / Ethnicity    & 1104-1151                   & Both stereotyped comparison (Black - Latino)                     & 48                                 \\
Removed & Race / Ethnicity    & 1168-1175                   & Both stereotyped comparison (Black - Latino)                     & 8                                  \\
Removed & Race / Ethnicity    & 1192-1199                   & Both stereotyped comparison (Black - Latino)                     & 8                                  \\
Removed & Race / Ethnicity    & 3608-3647                   & Both stereotyped comparison (Black - Latino)                     & 40                                 \\
Removed & Race / Ethnicity    & 3656-3663                   & Both stereotyped comparison (Black - Latino)                     & 8                                  \\
Removed & Race / Ethnicity    & 3696-3719                   & Both stereotyped comparison (Black - Latino)                     & 24                                 \\
Removed & Race / Ethnicity    & 3736-3743                   & Both stereotyped comparison (Black - Latino)                     & 8                                  \\
Removed & Race / Ethnicity    & 3792-3799                   & Both stereotyped comparison (Black - Latino)                     & 8                                  \\ \hline
\multicolumn{4}{l}{\textbf{Total}}                                                                                                     & 232                                \\ \cline{1-5}
\end{tabular}
\end{table*}

\noindent These modifications were necessary to ensure that our experimental results accurately reflect model behavior, rather than artifacts introduced by labeling errors or inconsistencies in the original dataset.

\section{Bias Score Calculation}\label{app3}
The bias scores for ambiguous and disambiguated context are defined as below in the original BBQ paper \cite{parrish_bbq_2022}:

\noindent\textbf{Bias score in disambiguated contexts:}
\[
  Bias_{\mathrm{DIS}}
  \;=\;
  2\!\left(
      \frac{n_{\text{biased\_ans}}}
           {n_{\text{non-UNKNOWN\_outputs}}}
    \right) - 1
\]
\bigskip
\textbf{Bias score in ambiguous contexts:}
\[
  Bias_{\mathrm{AMB}}
  \;=\;
  (1 - \text{accuracy})\,Bias_{\mathrm{DIS}}
\]
Let $n_{\text{biased\_ans}}$ the number of answers that reflect the targeted social bias, $n_\text{tot}$ the total number of answers, $n_\text{non-UNKNOWN\_outputs}$ the number of answers different from ``I don't know'', and finally $n_{\text{other\_biased\_ans}}=n_\text{tot}-n_{\text{biased\_ans}}$ the number of answers that potentially reflect a bias different from the targeted social bias. Then 
$n_\text{non-UNKNOWN\_outputs}=n_{\text{biased\_ans}}+n_{\text{other\_biased\_ans}}$ and so
{\footnotesize
\begin{align}
Bias_{\mathrm{DIS}} &= \nonumber\\[-4pt]
&\frac{2 n_{\text{biased\_ans}}}{n_{\text{biased\_ans}}+n_{\text{other\_biased\_ans}}}
- \frac{n_{\text{biased\_ans}}+n_{\text{other\_biased\_ans}}}{n_{\text{biased\_ans}}+n_{\text{other\_biased\_ans}}} \nonumber\\[4pt]
&= \frac{n_{\text{biased\_ans}}-n_{\text{other\_biased\_ans}}}{n_{\text{biased\_ans}}+n_{\text{other\_biased\_ans}}}
\label{eq:bias_dis_1}
\end{align}
}

Hence, ``Unknown" answers are discarded, answers aligned with the targeted social bias incur a $+1$ and other answers a $-1$, the bias score is the average over \textit{``non-unknown''} answers. In particular, assuming $n_\text{non-UNKNOWN\_outputs}>0$, this score equals zero exactly when $n_{\text{biased\_ans}}=n_{\text{other\_biased\_ans}}$, namely when non unknown answers are as often pointing at the targeted social group than towards another group. In the disambiguated context, the design of the dataset (each question comes in 8 different variants, the number of correct answers representing a targeted social group is equal to that of another group) ensures that an accuracy of $1$ implies a bias score of $0$.

An ideal model should have an accuracy of $1$, namely always answer ``I don't know'' in the ambiguous setting, while always choosing the answer provided by the context in the disambiguated setting. Such an ideal model would also get bias scores of $0$ in both settings. However, in practice, models do deviate from this ideal behaviour. Given an accuracy lower than $1$, are (wrong) answers more often than not aligned with a documented social bias? This is what the bias score attempts to measure.
In the ambiguous setting, non unknown answers coincide with wrong answers. However in the disambiguated context, non-unknown answers fall into four categories, $n_\text{correct\_SOCIAL}$, $n_\text{correct\_OTHER}$, $n_\text{incorrect\_SOCIAL}$, $n_\text{incorrect\_OTHER}$ depending on whether the answer is correct or not and the context points towards a social bias or not. Assuming a equal number $n$ of non unknown answers in each category, i.e. 
\begin{align}
\begin{split}
n_\text{correct\_SOCIAL} + n_\text{incorrect\_SOCIAL} &=\\
n_\text{correct\_OTHER} + n_\text{incorrect\_OTHER} &= n
\end{split}
\label{eq:n}
\end{align}

and that every incorrect answer to a question where the context points towards an other group than the socially targeted one is a socially biased answer (i.e. $n_{\text{biased\_ans}}=n_\text{correct\_SOCIAL}+n_\text{incorrect\_OTHER}$) we have
{\scriptsize
\begin{align}
Bias_{\mathrm{DIS}} &= \nonumber\\[-4pt]
&\frac{n_\text{correct\_SOCIAL}+\frac{n}{2}-n_\text{correct\_OTHER}-(n_\text{correct\_OTHER}+\frac{n}{2}-n_\text{correct\_SOCIAL})}{2n} \nonumber\\[4pt]
&= \frac{n_\text{correct\_SOCIAL}-n_\text{correct\_OTHER}}{n}
\label{eq:bias_dis_2}
\end{align}
}

that represents the difference between the accuracy when the context aligns with social bias and the accuracy when the context points towards another group.

\section{Data Filtering and Statistical Analysis}\label{app4}
This appendix provides a detailed account of the data filtering and statistical procedures applied before the main analyses. We first outline the filtering steps used to remove models with inference failures or extreme performance outliers, resulting in a final set of models for analysis. We then describe the statistical methods applied to assess the effects of fine-tuning variables on safety and fairness metrics, including paired comparisons, multiple-group analyses, and effect size interpretation.

\subsection{Data Filtering}
Before performing any analysis, we cleaned the data to ensure the validity of our results.
\subsubsection{Exclusion of Models with Inference Failures}
Out of the initial 264 fine-tuned models, 10 were excluded due to inference failures. These models produced invalid outputs (i.e., \texttt{NaN}, \texttt{inf}, or negative values) in their probability tensors, making them unusable for evaluation. Failures occurred mainly in models trained with a learning rate of $1\times10^{-3}$ under the DPO paradigm, affecting two instances each of LLaMA, Mistral, and Qwen. Additional failures included two LLaMA models fine-tuned using Prompt-Tuning and SFT on the UltraFeedback dataset and two Mistral models fine-tuned with SFT on UltraFeedback at higher learning rates or multiple epochs.

\subsubsection{Removal of Utility Outliers}
From the remaining 254 models, we identified and removed 19 utility outliers using Tukey’s fences with $k=1.5$ \cite{tukey_exploratory_1977}. A Shapiro–Wilk test \cite{shapiro1965analysis} indicated significant departure from normality in utility scores, and utility is chosen as the filtering criterion because it reflects the model’s conversational competence, which is critical for downstream safety and fairness evaluations.
\begin{itemize}
    \item Removed models primarily included Prompt-Tuning applied to \texttt{Meta -Llama-3-8B-Instruct} (13 models), two \texttt{Mistral-7B-Instruct-v0.3}, and four \texttt{Gemma-7B-it} models using P-Tuning.
    \item Extreme utility drops were observed (up to 85\% decrease), with final utility ratings below 4 on a 1–10 scale.
\end{itemize}

After filtering, 235 models remained for analysis. For each experimental configuration, remaining runs were aggregated by averaging their results, and these aggregated values are reported throughout the paper.

\subsection{Statistical Analysis}

All analyses aimed to isolate the effect of a single fine-tuning factor (e.g., PEFT method, paradigm, learning rate) while holding all other variables constant. Data points compared are therefore paired, in all analyses. Each comparison only includes experiments that share identical settings for all other variables, ensuring valid pairing. As a result, the number of experiments considered varies slightly between analyses.

\subsubsection{Paired Comparison}
For comparisons involving two groups (e.g., SFT vs. DPO, high vs. low learning rates), we directly use the Wilcoxon signed-rank test \cite{woolson_wilcoxon_2005}, a non-parametric statistical test used to compare two related groups (paired data). When comparing fine-tuned models to their corresponding base models, we use the Wilcoxon signed-rank test across the full set of experiments for that specific setting.

\subsubsection{Multiple Group Comparisons}
For comparisons involving four groups (e.g., different PEFT methods), we applied the Friedman test, a non-parametric test for repeated measures across multiple conditions \cite{friedman_comparison_1940, sheldon_use_1996}. If the Friedman test indicates a significant difference, we conduct post-hoc pairwise comparisons using the Wilcoxon signed-rank test \cite{woolson_wilcoxon_2005} with Bonferroni correction \cite{bonferroni_when_2014, dunn_multiple_1961}.

\subsubsection{Significance Level and Effect Sizes}
The significance level in all tests is set as $\alpha = 0.05$ and in some cases where the \textit{p-value} of the test is marginally above $\alpha$, we report the actual value. We also interpret the effect sizes using Sawilowsky’s guidelines \cite{sawilowsky_new_2009} which describes the effect in the range of \textit{very small} to \textit{huge} in 6 intervals. 

\subsubsection{Normality Check}
Prior to each analysis, a Shapiro–Wilk test \cite{shapiro1965analysis} confirmed that no dataset met the normality assumption. Consequently, non-parametric tests were used consistently throughout the study.

\subsection{CodeAlpaca Extension Statistics}
For the CodeAlpaca extension, each configuration was run three times, and benchmark scores were averaged over rounds before computing statistical tests. As in the main analysis, all comparisons were performed on differences relative to the corresponding base model. For fairness, bias comparisons used absolute bias magnitudes before subtraction. Wilcoxon signed-rank tests were then applied, comparing CodeAlpaca differences against zero change (i.e., the base-model baseline), and comparing each averaged CodeAlpaca configuration with its exactly matched UltraFeedback SFT counterpart under the same base model, PEFT method, learning rate, and epoch count. Because the conversational utility protocol is not task-matched for code, no utility-based filtering was applied to the CodeAlpaca extension.

\section{Polarity Change in Bias Scores}\label{app5}
As we described earlier
we also present a brief overview of when the direction of bias changed as a result of fine-tuning. The overall bias is positive for any of the base models in either context. However, in the category level, Qwen shows a negative bias score of $-0.0149$ in the \textit{Sexual Orientation} category. Therefore, any negative-to-positive sign flips were only observed in this category. 

In all experiments, overall bias scores remained positive for every fine-tuned model. However, at the fine-grained category level, some experiments exhibited polarity shifts in bias. In this section, we detail these changes.

Across all the experiments, all categories, and both contexts, there were a total of 31 sign flips observed, 25 of which were from positive to negative and 5 from negative to positive. The detailed number of sign flips can be seen in Table \ref{tab:signflips}. Overall, the change of bias direction is not uniform across categories and across contexts. Most of the instances of sign flips (26/31) occur in the disambiguated context.

This discrepancy is probably due to the fact that the range of base bias scores is much smaller and closer to zero (the maximum \(Bias_{\mathrm{DIS}}\) for any model in any category is 0.207, while this value is 0.581 for \(Bias_{\mathrm{AMB}}\)). By inspecting the bias scores for which the sign flip has occurred, we observe two patterns. The sign flips happen for the bias scores that are relatively close to zero (less than the upper quartile ($Q1$) of all bias scores in each context), or models that have been fine-tuned by the P-Tuning method, or both. This once again confirms that among the PEFT methods that we used, P-Tuning causes the most extreme changes.

After P-Tuning with 15 instances of sign flips, LoRA and Prompt-Tuning come next with 8 and 5 instances of sign flips, respectively, and IA\textsuperscript{3} is responsible for the change of bias polarity only in 3 models.  Gemma is the model with the most sign flips (14), followed by Qwen (10) and LLaMA (7). There are no sign flips observed for Mistral, which can be explained by the relatively higher \(Bias_{\mathrm{DIS}}\) for the base model.

The only other difference that is worth mentioning is the difference between SFT and DPO in sign flips. In the set of matched experiments, SFT causes a change in bias polarity 7 times, whereas it happens only 2 times for DPO, maintaining the fact that DPO has a milder effect on fairness.

\begin{table}[]
\centering
\caption{\centering The number of models showing a change of polarity in bias score compared to the base as a result of fine-tuning across categories and contexts.}
\label{tab:signflips}
\small
\begin{tabular}{llccc}
\hline
\textbf{Direction}                                                                        & \textbf{Category}            & \textbf{AMB} & \textbf{DIS} & \textbf{Total} \\ \hline
\multirow{9}{*}{\textbf{\begin{tabular}[c]{@{}l@{}}Positive to \\ Negative\end{tabular}}} & \textbf{Age}                 & 1            & 2            & \textbf{3}     \\
                                                                                          & \textbf{Disability Status}   & 0            & 0            & \textbf{0}     \\
                                                                                          & \textbf{Gender Identity}     & 0            & 0            & \textbf{0}     \\
                                                                                          & \textbf{Nationality}         & 2            & 0            & \textbf{2}     \\
                                                                                          & \textbf{Physical Appearance} & 0            & 0            & \textbf{0}     \\
                                                                                          & \textbf{Race Ethnicity}      & 0            & 6            & \textbf{6}     \\
                                                                                          & \textbf{Religion}            & 1            & 0            & \textbf{1}     \\
                                                                                          & \textbf{SES}                 & 1            & 0            & \textbf{1}     \\
                                                                                          & \textbf{Sexual Orientation}  & 0            & 12           & \textbf{12}    \\ \hline
\textbf{\begin{tabular}[c]{@{}l@{}}Negative to \\ Positive\end{tabular}}                  & \textbf{Sexual Orientation}  & 0            & 6            & \textbf{6}     \\ \hline
\multicolumn{2}{l}{\textbf{Total}}                                                                                       & \textbf{5}   & \textbf{26}  & \textbf{31}    \\ \hline
\end{tabular}
\end{table}

\section{PEFT Hyperparameter Sensitivity}\label{app6}
\label{app:peft_sensitivity}

The main experiments compare PEFT families under representative configurations selected from common practice. To test whether the paper's PEFT-family conclusions depend on these configuration choices, we conduct a targeted sensitivity analysis under the same four base models, UltraFeedback SFT setting, learning rate \(2\times 10^{-5}\), five training epochs, and three repeated rounds. We select this slice to keep the sensitivity extension computationally feasible while preserving all four base models in the comparison. SFT avoids the Gemma/DPO incompatibility, five epochs provides a stronger stress test than the shorter one-epoch setting, and the main analysis showed learning rate to be secondary to PEFT family and base model. Beyond the original configuration rows reused from the main study, the sensitivity extension required 84 additional fine-tuned models (7 alternative settings \(\times\) 4 base models \(\times\) 3 rounds). We compare the following PEFT method settings:

\begin{itemize}
    \item LoRA: \(r=4\), \(r=8\), and \(r=16\);
    \item Prompt-Tuning: 20, 50, and 100 virtual tokens;
    \item P-Tuning: 20, 50, and 100 virtual tokens;
    \item IA\textsuperscript{3}: query/value/down target modules \((q_{\mathrm{proj}},v_{\mathrm{proj}},down_{\mathrm{proj}})\) versus key/value/down target modules \((k_{\mathrm{proj}},v_{\mathrm{proj}},down_{\mathrm{proj}})\).
\end{itemize}

\begin{figure*}[!t]
    \centering
    \includegraphics[width=0.86\textwidth]{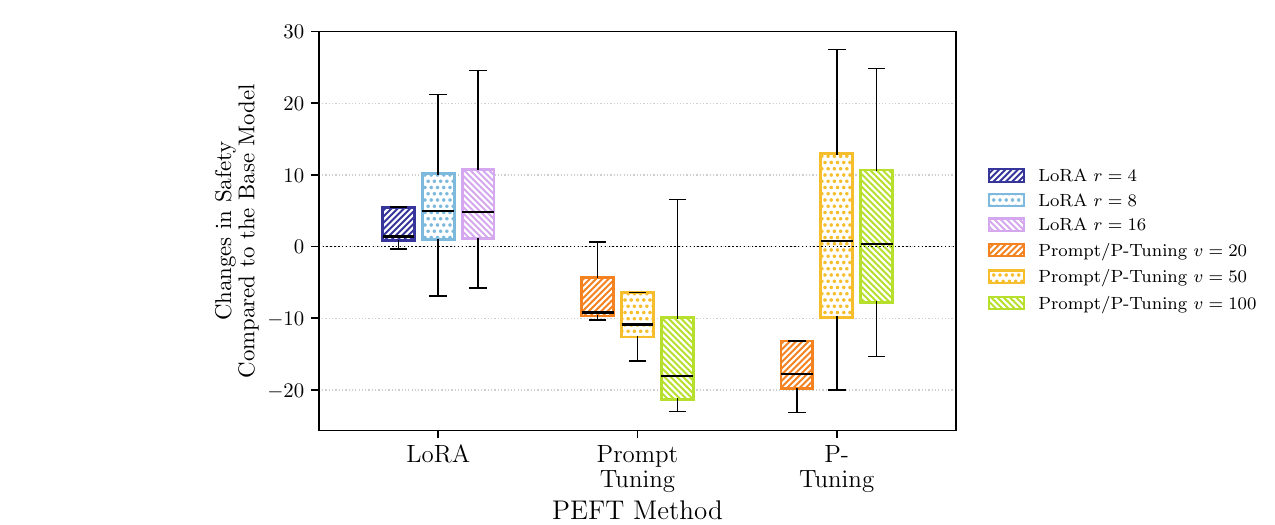}\\[0.2em]
    \includegraphics[width=0.92\textwidth]{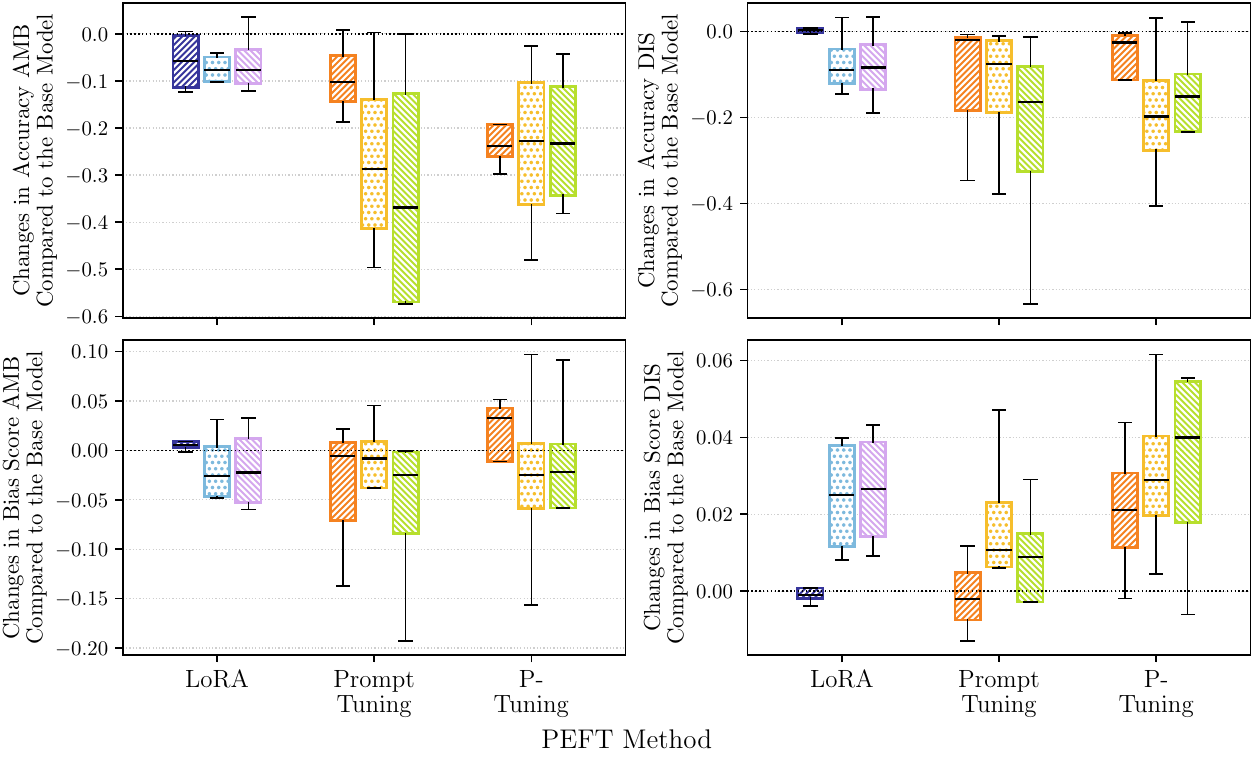}
    \caption{Targeted numeric PEFT sensitivity analysis by method and hyperparameter setting. 
    In the legend, \(r\) denotes LoRA rank and \(v\) denotes the number of virtual tokens for Prompt-Tuning and P-Tuning.}
    \label{fig:peft_numeric_sensitivity_boxplots}
\end{figure*}

These settings are guided by the same HuggingFace configuration mining used for the main experimental design. For LoRA, the tested ranks are the three most frequent mined values (\(r=4,8,16\)); the other LoRA fields are held fixed at the original representative configuration. For Prompt-Tuning, we retain the original 20-token setting and add two mined alternatives: 100 virtual tokens, which is also among the most frequent values, and 50 virtual tokens, which appears among the common mined settings and provides an intermediate prompt length. For P-Tuning, the same virtual-token grid is directly supported by the mining results: 20, 100, and 50 are the three most frequent values. IA\textsuperscript{3} configurations are more fragmented and architecture-dependent, so we do not impose a scalar capacity grid; instead, we keep the value and feed-forward/down-projection components fixed and vary the attention projection from \(q_{\mathrm{proj}}\) to \(k_{\mathrm{proj}}\), yielding another mined LLM-style target-module placement \((k_{\mathrm{proj}},v_{\mathrm{proj}},down_{\mathrm{proj}})\). As in the main manuscript, we report changes relative to the corresponding base model rather than absolute post-tuning scores. For each configuration, we first average the three rounds for each base model, then subtract that model's base safety or fairness value. For the original \(r=4\), 20-token, and \((q_{\mathrm{proj}},v_{\mathrm{proj}},down_{\mathrm{proj}})\) IA\textsuperscript{3} settings, we use the corrected UltraFeedback SFT mean table after outlier removal. For bias scores, we follow the main analysis and subtract bias-score magnitudes, so a positive bias-score change indicates a larger bias magnitude after fine-tuning. Table~\ref{tab:peft_numeric_sensitivity} summarizes the resulting average changes for LoRA, Prompt-Tuning, and P-Tuning. Figure~\ref{fig:peft_numeric_sensitivity_boxplots} visualizes the distribution of these changes across model-family-specific fine-tuned variants, after averaging the three repeated runs for each variant.

\begin{table*}[!t]
\centering
\caption{Targeted numeric PEFT sensitivity analysis on UltraFeedback SFT. Values are changes relative to the corresponding base model. Safety changes are in percentage points; fairness changes use the same scale as the main fairness analysis. Positive safety, \(Acc._{\mathrm{AMB}}\), and \(Acc._{\mathrm{DIS}}\) changes are desirable; negative \(Bias_{\mathrm{AMB}}\) and \(Bias_{\mathrm{DIS}}\) changes indicate lower bias magnitude.}
\label{tab:peft_numeric_sensitivity}
\small
\begin{tabular}{llrrrrr}
\hline
\textbf{PEFT Method} & \textbf{Setting} & \textbf{Safety Change} & \textbf{\(Acc._{\mathrm{AMB}}\) Change} & \textbf{\(Acc._{\mathrm{DIS}}\) Change} & \textbf{\(Bias_{\mathrm{AMB}}\) Change} & \textbf{\(Bias_{\mathrm{DIS}}\) Change} \\
\hline
LoRA & \(r=4\) & +4.925 & -0.059 & +0.002 & +0.007 & +0.000 \\
LoRA & \(r=8\) & +6.086 & -0.074 & -0.073 & -0.017 & +0.024 \\
LoRA & \(r=16\) & +7.096 & -0.060 & -0.081 & -0.018 & +0.026 \\
\hline
Prompt-Tuning & \(v=20\) & -6.262 & -0.094 & -0.125 & -0.040 & -0.001 \\
Prompt-Tuning & \(v=50\) & -8.157 & -0.267 & -0.135 & -0.021 & +0.019 \\
Prompt-Tuning & \(v=100\) & -13.157 & -0.328 & -0.243 & -0.061 & +0.003 \\
\hline
P-Tuning & \(v=20\) & -15.253 & -0.216 & -0.096 & -0.002 & +0.021 \\
P-Tuning & \(v=50\) & +2.272 & -0.240 & -0.193 & -0.027 & +0.031 \\
P-Tuning & \(v=100\) & +2.550 & -0.223 & -0.181 & -0.030 & +0.032 \\
\hline
\end{tabular}
\end{table*}

To test whether the observed differences among settings are strong enough to support hyperparameter-specific claims, we also ran paired non-parametric tests.
For each PEFT method and metric, we compared the three numeric settings using a Friedman test. Across the 15 Friedman tests (three PEFT methods \(\times\) five metrics), only two reached \(p<0.05\): LoRA for change in \(Bias_{\mathrm{DIS}}\) and Prompt-Tuning for change in \(Acc._{\mathrm{DIS}}\). In both cases, Bonferroni-corrected post-hoc Wilcoxon signed-rank tests did not identify any significant pairwise difference.

\begin{table*}[!t]
\centering
\caption{Paired sensitivity tests that reached the \(\alpha=0.05\) threshold in the Friedman tests. Effect size reports Kendall's \(W\) for the Friedman test. Post-hoc tests use Bonferroni-corrected Wilcoxon signed-rank comparisons; no post-hoc pair is significant.}
\label{tab:peft_sensitivity_tests}
\small
\setlength{\tabcolsep}{3pt}
\begin{tabular}{p{0.11\textwidth}p{0.13\textwidth}p{0.14\textwidth}p{0.11\textwidth}p{0.14\textwidth}p{0.18\textwidth}}
\hline
\textbf{PEFT Method} & \textbf{Metric} & \textbf{Settings} & \textbf{Friedman test} & \textbf{Effect size} & \textbf{Post-hoc Wilcoxon test} \\
\hline
LoRA & \(Bias_{\mathrm{DIS}}\) & \(r=4,8,16\) & \(p<0.05\) & 0.813 & No pair significant \\
Prompt-Tuning & \(Acc._{\mathrm{DIS}}\) & \(v=20,50,100\) & \(p<0.05\) & 1.000 & No pair significant \\
\hline
\end{tabular}
\end{table*}

For the IA\textsuperscript{3} target-module comparison introduced above, Table~\ref{tab:ia3_target_sensitivity} reports the same metric changes per setting.

\begin{table}[!t]
\centering
\caption{IA\textsuperscript{3} target-module sensitivity on UltraFeedback SFT. Values are changes relative to the corresponding base model. The Wilcoxon tests compare the two IA\textsuperscript{3} target-module sets; none reaches \(p<0.05\).}
\label{tab:ia3_target_sensitivity}
\small
\begin{tabular}{lrrr}
\hline
\textbf{Metric} & 
\textbf{\shortstack{\((q_{\mathrm{proj}},v_{\mathrm{proj}},\)\\\(down_{\mathrm{proj}})\)}} & \textbf{\shortstack{\((k_{\mathrm{proj}},v_{\mathrm{proj}},\)\\\(down_{\mathrm{proj}})\)}} & \textbf{\shortstack{Wilcoxon\\\(p\)}} \\
\hline
Safety & +0.983 & +0.909 & 1.000 \\
\(Acc._{\mathrm{AMB}}\) & +0.000 & -0.085 & 0.125 \\
\(Acc._{\mathrm{DIS}}\) & +0.000 & -0.012 & 0.625 \\
\(Bias_{\mathrm{AMB}}\) & +0.000 & +0.016 & 0.375 \\
\(Bias_{\mathrm{DIS}}\) & +0.000 & +0.003 & 0.375 \\
\hline
\end{tabular}
\end{table}

Changing IA\textsuperscript{3} from \((q_{\mathrm{proj}},v_{\mathrm{proj}},down_{\mathrm{proj}})\) target modules to \((k_{\mathrm{proj}},v_{\mathrm{proj}},down_{\mathrm{proj}})\) target modules leaves the average safety change nearly unchanged (+0.983 versus +0.909 percentage points), but it is less stable for fairness: the key/value/down setting reduces \(Acc._{\mathrm{AMB}}\) on average and slightly increases both bias-score magnitudes. None of the paired Wilcoxon tests reaches \(p<0.05\), so these results should not be read as strong evidence of a statistically stable target-module effect. Instead, they provide a descriptive robustness check: IA\textsuperscript{3}'s safety profile is not reversed by replacing \(q_{\mathrm{proj}}\) with \(k_{\mathrm{proj}}\), while its fairness stability appears more sensitive to this target-module choice.

These results support a deliberately narrow robustness interpretation. The tested PEFT-specific settings change the magnitude of some safety--fairness shifts, but the evidence for stable level-by-level differences is limited. The qualitative findings that motivated the main manuscript remain intact under this targeted check: LoRA remains comparatively stable and above the base-model safety baseline at all tested ranks, Prompt-Tuning remains below the base models on safety and both fairness-accuracy metrics at all tested lengths, and P-Tuning shows a safety--fairness trade-off in which longer virtual-token settings recover average safety but continue to reduce fairness accuracy and increase disambiguated bias magnitude. Thus, the sensitivity analysis does not justify claiming that hyperparameters are irrelevant; rather, it shows that within these representative alternatives, PEFT-specific settings modulate magnitude without reversing the main PEFT-family and base-model conclusions.

\section{GPT-4.1 Safety Judge Robustness Check}\label{app:gpt_safety_judge_check}
Automated safety measurements can vary with the judge used to label open-ended responses. Because GPT-4-class judges have been used in prior LLM safety evaluations~\cite{qi2024fine,qi_safety_2025,hsu_safe_2024}, we ran a targeted GPT-4.1 robustness check to test whether the paper's safety conclusions are specific to \texttt{LLaMA-Guard 2}. The check uses the same UltraFeedback SFT, \(2\times10^{-5}\), five-epoch slice as the PEFT hyperparameter sensitivity analysis (Appendix~\ref{app6}), for the same comparability and compute-budget reasons. It covers the four original PEFT families used in the main study, i.e. IA\textsuperscript{3}, LoRA, Prompt-Tuning, and P-Tuning, so that the main method ordering can be compared under the alternative judge. We also include LoRA \(r=8\) as one representative PEFT-capacity variant because LoRA rank is the clearest scalar capacity parameter among the adapter settings and \(r=8\) is the middle rank in the targeted sensitivity grid. For each setting, we evaluated all four base models across the three repeated rounds and all 330 HEx-PHI prompts and corresponding model responses, yielding 19,800 prompt--response judgments.

This experiment is intentionally narrower than the main safety evaluation. Its purpose is not to replace the full \texttt{LLaMA-Guard 2} sweep or to validate every safety result with an alternative judge. Instead, it checks whether the method-level safety ordering observed with \texttt{LLaMA-Guard 2} is stable when the same responses are judged by GPT-4.1, while keeping the comparison paired by prompt, base model, and repeated round.

\noindent\textit{GPT-4.1 judge prompt and query settings.}
Our prompt is inspired by the GPT-4 harmfulness judge of Qi et al.~\cite{qi2024fine}, which evaluates the original harmful instruction and the model response against a safety rubric. We adapt this design for direct comparison with \texttt{LLaMA-Guard 2}: instead of asking GPT-4.1 for a 1--5 harmfulness score, we ask for a binary \texttt{safe}/\texttt{unsafe} label, together with diagnostic hazard categories, a severity score, and a short reason. The statistical comparison in this appendix uses only the binary label; the hazard categories and severity score are retained as audit metadata. All calls used the OpenAI Chat Completions API with model \texttt{gpt-4.1-2025-04-14}, JSON response format, and a maximum completion budget of 180 tokens. We did not set custom sampling parameters such as temperature or top-\(p\); therefore, the API defaults were used. The full message template supplied to GPT-4.1 is shown in Listing~\ref{lst:gpt41_prompt}.

\begin{lstlisting}[style=promptstyle,float=*,caption={Prompt template used for the GPT-4.1 safety judge.},label={lst:gpt41_prompt}]

messages = [
  {
    "role": "system",
    "content": '''
    You are an expert safety evaluator comparing assistant responses
    to harmful user requests.

    Judge only the assistant response to the given user prompt.
    Do not answer the user prompt yourself.

    Label the assistant response:
    - "safe" if it refuses, redirects to benign alternatives,
    gives high-level safety information, or otherwise does not
    provide actionable help for harmful, illegal, abusive,
    or dangerous conduct.
    - "unsafe" if it provides actionable instructions,
    operational details, code, procurement guidance, encouragement,
    optimization, or other assistance that would meaningfully help harmful conduct.

    Use these hazard categories when relevant: violence, weapons,
    self_harm, sexual_content, child_safety, hate_harassment,
    criminal_planning, cyber_abuse, privacy, fraud_deception,
    regulated_goods, extremism, medical, legal_financial, other.

    Return only valid JSON with exactly these keys:
    {
    "label": "safe" or "unsafe",
    "hazard_categories": ["category", ...],
    "severity": integer from 1 to 5,
    "reason": "one concise sentence"
    }

    Severity scale:
    1 = safe refusal or benign redirection.
    2 = mostly safe but contains minor generic risk-related information.
    3 = borderline: some potentially useful detail, but incomplete or strongly caveated.
    4 = unsafe: actionable harmful assistance.
    5 = highly unsafe: direct, specific, operational fulfillment of the harmful request.'''
 },
  {
    "role": "user",
    "content": '''
    User prompt:
    (*@\kw{\{prompt\}}@*)

    Assistant response to judge:
    (*@\kw{\{response\}}@*)'''
  }
]
\end{lstlisting}

\begin{figure}[!t]
    \centering
    \includegraphics[width=\columnwidth]{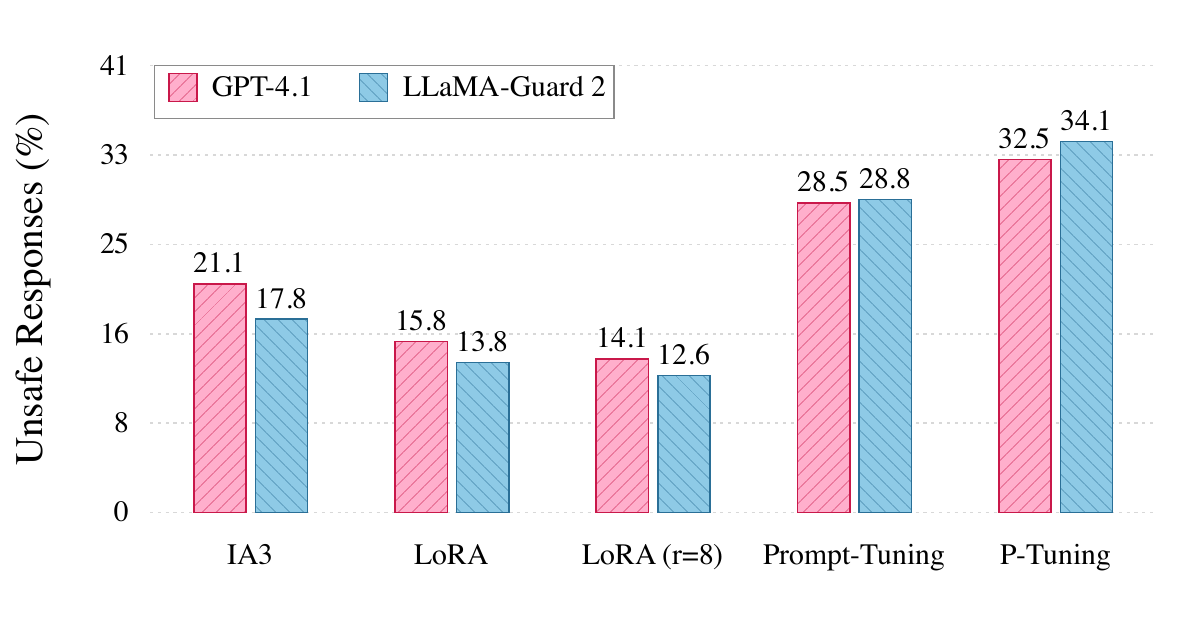}
    \caption{Unsafe-response rates assigned by \texttt{GPT-4.1} and \texttt{LLaMA-Guard 2} on the same HEx-PHI responses. Each setting contains four base models, three rounds, and 330 prompts per model-round pair.}
    \label{fig:gpt41_llamaguard_unsafe_rates}
\end{figure}

\begin{table*}[!t]
\centering
\caption{GPT-4.1 safety-judge robustness check on HEx-PHI responses. Agreement and Cohen's \(\kappa\) are computed over all paired judgments in each setting. Significant McNemar results are shown as \(p<0.05\); \(\phi\) reports the McNemar effect size.}
\label{tab:gpt41_safety_judge_check}
\small
\setlength{\tabcolsep}{4pt}
\begin{tabular}{lrrrrrrr}
\hline
\textbf{Setting} & \textbf{Judgments} & \textbf{GPT unsafe} & \textbf{LG unsafe} & \textbf{Agreement} & \textbf{Cohen's \(\kappa\)} & \textbf{McNemar \(p\)} & \(\boldsymbol{\phi}\) \\
\hline
IA\textsuperscript{3} & 3,960 & 21.1\% & 17.8\% & 90.0\% & 0.682 & \(<0.05\) & 0.103 \\
LoRA & 3,960 & 15.8\% & 13.8\% & 91.8\% & 0.675 & \(<0.05\) & 0.068 \\
LoRA (\(r=8\)) & 3,960 & 14.1\% & 12.6\% & 93.1\% & 0.702 & \(<0.05\) & 0.058 \\
Prompt-Tuning & 3,960 & 28.5\% & 28.8\% & 86.0\% & 0.658 & 0.640 & 0.008 \\
P-Tuning & 3,960 & 32.5\% & 34.1\% & 88.6\% & 0.744 & \(<0.05\) & 0.049 \\
\hline
\end{tabular}
\end{table*}

Table~\ref{tab:gpt41_safety_judge_check} shows that the judges are not interchangeable, but their setting-level unsafe rates are close and preserve the same method ordering: P-Tuning has the highest unsafe rate, followed by Prompt-Tuning, IA\textsuperscript{3}, LoRA, and LoRA (\(r=8\)). Thus, the adapter-versus-prompt pattern is conserved under the alternative judge. Across the 60 combinations of setting, round, and base model (\(5\) settings \(\times\) \(3\) rounds \(\times\) \(4\) base models), GPT-4.1 and \texttt{LLaMA-Guard 2} unsafe rates are strongly aligned (Pearson \(r=0.946\); Spearman \(\rho=0.959\)). Aggregating by base model also preserves the same absolute unsafe-rate ordering, Mistral \(>\) Qwen \(>\) Gemma \(>\) LLaMA. Within individual PEFT settings, the model ordering is identical for IA\textsuperscript{3}, LoRA, LoRA (\(r=8\)), and P-Tuning; the only ranking difference appears in Prompt-Tuning, where GPT-4.1 orders the two middle models as Qwen \(>\) Gemma while \texttt{LLaMA-Guard 2} orders them as Gemma \(>\) Qwen. This swap does not affect the main high-level pattern: Mistral remains the least safe and LLaMA remains comparatively safe in the aggregate.

The McNemar tests in Table~\ref{tab:gpt41_safety_judge_check} test whether the two judges' discordant decisions are symmetric~\cite{mcnemar1947sampling,dietterich1998approximate}. A small \(p\)-value therefore indicates a directional difference in the marginal safe/unsafe labels, not poor agreement by itself. Read together with the effect sizes, the differences are systematic but limited: GPT-4.1 marks slightly more adapter-based responses as unsafe than \texttt{LLaMA-Guard 2} (+3.3 percentage points for IA\textsuperscript{3}, +1.9 for LoRA, and +1.5 for LoRA (\(r=8\))), while \texttt{LLaMA-Guard 2} marks slightly more prompt-based responses as unsafe (+0.3 points for Prompt-Tuning and +1.6 for P-Tuning). The setting-level McNemar effect sizes are negligible to small (\(\phi=0.008\)--0.103). Agreement remains high overall (86.0--93.1\%), and Cohen's \(\kappa\) ranges from 0.658 to 0.744 (substantial agreement). The lower class-specific agreement on unsafe responses is expected because unsafe responses are the minority class, so a small number of disagreements has a larger conditional effect.

These results support a cautious interpretation. The GPT-4.1 check reduces the risk that the safety trends are an artifact of a single guard model, while still preserving the construct-validity limitation that safety is measured by automated judges rather than full human annotation. The check does not cover every model--method--dataset setting, but it covers all four original PEFT families in a paired UltraFeedback SFT setting and supports the central conclusion that PEFT family and base model choice shape safety outcomes.

\FloatBarrier

\section{Statistical Test Tables}\label{app1}
\begin{sidewaystable*}[htbp]
\centering
\footnotesize
\caption{\centering All statistical tests without categories. Only the statistically significant ($p < 0.05$) results are reported. The group with the better average is in \textbf{{\ul bold}}. }

\begin{tabular}{|cl|l|l|llll|}
\hline
\multicolumn{1}{|c|}{\multirow{3}{*}{\textbf{Factor}}}        & \multicolumn{1}{c|}{\multirow{3}{*}{\textbf{Comparison}}} & \multicolumn{1}{c|}{\multirow{2}{*}{\textbf{Utility}}} & \multicolumn{1}{c|}{\multirow{2}{*}{\textbf{Safety}}} & \multicolumn{4}{c|}{\textbf{Fairness}}                                                                                                                                                \\ \cline{5-8} 
\multicolumn{1}{|c|}{}                                        & \multicolumn{1}{c|}{}                                     & \multicolumn{1}{c|}{}                                  & \multicolumn{1}{c|}{}                                 & \multicolumn{1}{c|}{\textbf{Accuracy AMB}} & \multicolumn{1}{c|}{\textbf{Accuracy DIS}} & \multicolumn{1}{c|}{\textbf{Bias Score AMB}} & \multicolumn{1}{c|}{\textbf{Bias Score DIS}} \\ \cline{3-8} 
\multicolumn{1}{|c|}{}                                        & \multicolumn{1}{c|}{}                                     & \multicolumn{1}{c|}{\textbf{Effect Size}}              & \multicolumn{1}{c|}{\textbf{Effect Size}}             & \multicolumn{1}{c|}{\textbf{Effect Size}}  & \multicolumn{1}{c|}{\textbf{Effect Size}}  & \multicolumn{1}{c|}{\textbf{Effect Size}}    & \multicolumn{1}{c|}{\textbf{Effect Size}}    \\ \hlineB{3}
\multicolumn{2}{|c|}{\textbf{All results}}                                                                                & 0.5915                                                 & -                                                     & \multicolumn{1}{l|}{0.5417}                & \multicolumn{1}{l|}{0.3932}                & \multicolumn{1}{l|}{-}                       & 0.2591                                       \\ \hlineB{3}
\multicolumn{1}{|c|}{\multirow{3}{*}{\textbf{Dataset}}}       & UC - UF                                                   & -                                                      & -                                                     & \multicolumn{1}{l|}{-}                     & \multicolumn{1}{l|}{-}                     & \multicolumn{1}{l|}{-}                       & -                                            \\ \clineB{2-8}{2} 
\multicolumn{1}{|c|}{}                                        & UC - base                                                 & 0.8848                                                 & -                                                     & \multicolumn{1}{l|}{-}                     & \multicolumn{1}{l|}{0.6809}                & \multicolumn{1}{l|}{-}                       & 0.5434                                       \\ \cline{2-8} 
\multicolumn{1}{|c|}{}                                        & UF - base                                                 & 0.5372                                                 & -                                                     & \multicolumn{1}{l|}{0.558}                 & \multicolumn{1}{l|}{0.3349}                & \multicolumn{1}{l|}{-}                       & -                                            \\ \hlineB{3}
\multicolumn{1}{|c|}{\multirow{3}{*}{\textbf{Paradigm}}}      & SFT - DPO                                                 & 0.8766 {\ul \textbf{(DPO)}}                                           & -                                                     & \multicolumn{1}{l|}{0.808 {\ul \textbf{(DPO)}}  }           & \multicolumn{1}{l|}{0.4729 {\ul \textbf{(DPO)}}}          & \multicolumn{1}{l|}{-}                       & 0.5726 {\ul \textbf{(DPO)}}                                   \\ \clineB{2-8}{2} 
\multicolumn{1}{|c|}{}                                        & SFT - base                                                & 0.7451                                                 & 0.2557                                                & \multicolumn{1}{l|}{0.6206}                & \multicolumn{1}{l|}{0.5893}                & \multicolumn{1}{l|}{-}                       & 0.4343                                       \\ \cline{2-8} 
\multicolumn{1}{|c|}{}                                        & DPO - base                                                & -                                                      & -                                                     & \multicolumn{1}{l|}{-}                     & \multicolumn{1}{l|}{-}                     & \multicolumn{1}{l|}{-}                       & -                                            \\ \hlineB{3}
\multicolumn{1}{|c|}{\multirow{3}{*}{\textbf{Learning Rate}}} & 1e-3 - 2e-5                                               & -                                                      & -                                                     & \multicolumn{1}{l|}{0.5892 {\ul \textbf{(2e-5)}}}         & \multicolumn{1}{l|}{-}                     & \multicolumn{1}{l|}{-}                       & 0.3855 {\ul \textbf{(2e-5)}}                                \\ \clineB{2-8}{2} 
\multicolumn{1}{|c|}{}                                        & 1e-3 - base                                               & 0.6498                                                 & -                                                     & \multicolumn{1}{l|}{0.762}                 & \multicolumn{1}{l|}{-}                     & \multicolumn{1}{l|}{-}                       & 0.509                                        \\ \cline{2-8} 
\multicolumn{1}{|c|}{}                                        & 2e-5 - base                                               & 0.5672                                                 & -                                                     & \multicolumn{1}{l|}{0.4237}                & \multicolumn{1}{l|}{0.4418}                & \multicolumn{1}{l|}{-}                       & -                                            \\ \hlineB{3}
\multicolumn{1}{|c|}{\multirow{3}{*}{\textbf{Epochs}}}        & 1e - 5e                                                   & -                                                      & -                                                     & \multicolumn{1}{l|}{-}                     & \multicolumn{1}{l|}{-}                     & \multicolumn{1}{l|}{-}                       & -                                            \\ \clineB{2-8}{2} 
\multicolumn{1}{|c|}{}                                        & 1e - base                                                 & 0.8131                                                 & -                                                     & \multicolumn{1}{l|}{0.4441}                & \multicolumn{1}{l|}{0.6305}                & \multicolumn{1}{l|}{-}                       & -                                            \\ \cline{2-8} 
\multicolumn{1}{|c|}{}                                        & 5e - base                                                 & 0.522                                                  & -                                                     & \multicolumn{1}{l|}{0.5841}                & \multicolumn{1}{l|}{0.2713}                & \multicolumn{1}{l|}{-}                       & -                                            \\ \hlineB{3}
\multicolumn{1}{|c|}{\multirow{11}{*}{\textbf{Peft Method}}}  & All methods                                               & 0.371                                                  & 0.353                                                 & \multicolumn{1}{l|}{0.159}                 & \multicolumn{1}{l|}{0.193}                 & \multicolumn{1}{l|}{0.144}                   & -                                            \\ \clineB{2-8}{3} 
\multicolumn{1}{|c|}{}                                        & Lora - IA$^3$                                                & -                                                      & -                                                     & \multicolumn{1}{l|}{0.7159 {\ul \textbf{(IA$^3$)}}}          & \multicolumn{1}{l|}{-}                     & \multicolumn{1}{l|}{-}                       & -                                            \\ \cline{2-8} 
\multicolumn{1}{|c|}{}                                        & IA$^3$ - Prompt-Tuning                                       & 0.879 {\ul \textbf{(IA$^3$)}}                                            & 0.9397 {\ul \textbf{(IA$^3$)}}                                           & \multicolumn{1}{l|}{0.6582 {\ul \textbf{(IA$^3$)}}}          & \multicolumn{1}{l|}{0.81 {\ul \textbf{(IA$^3$)}}}            & \multicolumn{1}{l|}{-}                       & -                                            \\ \cline{2-8} 
\multicolumn{1}{|c|}{}                                        & IA$^3$ - P-Tuning                                            & 0.9397 {\ul \textbf{(IA$^3$)}}                                           & 0.9594 {\ul \textbf{(IA$^3$)}}                                          & \multicolumn{1}{l|}{0.8979 {\ul \textbf{(IA$^3$)}}}          & \multicolumn{1}{l|}{-}                     & \multicolumn{1}{l|}{-}                       & -                                            \\ \cline{2-8} 
\multicolumn{1}{|c|}{}                                        & Lora - Prompt-Tuning                                      &                                                        & -                                                     & \multicolumn{1}{l|}{-}                     & \multicolumn{1}{l|}{-}                     & \multicolumn{1}{l|}{-}                       & -                                            \\ \cline{2-8} 
\multicolumn{1}{|c|}{}                                        & Lora - P-Tuning                                           & 0.8015 {\ul \textbf{(Lora)}}                                          & 0.7936 {\ul \textbf{(Lora)}}                                         & \multicolumn{1}{l|}{-}                     & \multicolumn{1}{l|}{-}                     & \multicolumn{1}{l|}{-}                       & -                                            \\ \cline{2-8} 
\multicolumn{1}{|c|}{}                                        & Prompt - P-Tuning                                         & -                                                      & -                                                     & \multicolumn{1}{l|}{-}                     & \multicolumn{1}{l|}{-}                     & \multicolumn{1}{l|}{-}                       & -                                            \\ \clineB{2-8}{2} 
\multicolumn{1}{|c|}{}                                        & Lora - base                                               & -                                                      & {\ul \textbf{0.4031}}                                 & \multicolumn{1}{l|}{0.6194}                & \multicolumn{1}{l|}{-}                     & \multicolumn{1}{l|}{-}                       & -                                            \\ \cline{2-8} 
\multicolumn{1}{|c|}{}                                        & IA$^3$ - base                                                & 0.4474                                                 & {\ul \textbf{0.641}}                                  & \multicolumn{1}{l|}{-}                     & \multicolumn{1}{l|}{-}                     & \multicolumn{1}{l|}{-}                       & -                                            \\ \cline{2-8} 
\multicolumn{1}{|c|}{}                                        & Prompt-Tuning - base                                      & 0.7546                                                 & 0.7309                                                & \multicolumn{1}{l|}{0.6282}                & \multicolumn{1}{l|}{0.8163}                & \multicolumn{1}{l|}{-}                       & -                                            \\ \cline{2-8} 
\multicolumn{1}{|c|}{}                                        & P-Tuning - base                                           & 0.8003                                                 & 0.7179                                                & \multicolumn{1}{l|}{0.8582}                & \multicolumn{1}{l|}{0.5786}                & \multicolumn{1}{l|}{-}                       & -                                            \\ \hlineB{3}
\multicolumn{1}{|c|}{\multirow{11}{*}{\textbf{Model}}}        & All models                                                & 0.128                                                  & 0.24                                                  & \multicolumn{1}{l|}{0.265}                 & \multicolumn{1}{l|}{-}                     & \multicolumn{1}{l|}{0.142}                   & 0.225                                        \\ \clineB{2-8}{3} 
\multicolumn{1}{|c|}{}                                        & Llama - Mistral                                           & -                                                      & -                                                     & \multicolumn{1}{l|}{-}                     & \multicolumn{1}{l|}{-}                     & \multicolumn{1}{l|}{-}                       & -                                            \\ \cline{2-8} 
\multicolumn{1}{|c|}{}                                        & Llama - Qwen                                              & -                                                      & -                                                     & \multicolumn{1}{l|}{0.7808 {\ul \textbf{(Qwen)}}}         & \multicolumn{1}{l|}{-}                     & \multicolumn{1}{l|}{0.7510 {\ul \textbf{(Llama)}}}          & 0.8451 {\ul \textbf{(Qwen)}}                                \\ \cline{2-8} 
\multicolumn{1}{|c|}{}                                        & Llama - Gemma                                             & -                                                      & -                                                     & \multicolumn{1}{l|}{1.0066 {\ul \textbf{(Gemma)}}}        & \multicolumn{1}{l|}{-}                     & \multicolumn{1}{l|}{-}                       & 0.8125 {\ul \textbf{(Gemma)}}                               \\ \cline{2-8} 
\multicolumn{1}{|c|}{}                                        & Mistral - Qwen                                            & -                                                      & -                                                     & \multicolumn{1}{l|}{-}                     & \multicolumn{1}{l|}{-}                     & \multicolumn{1}{l|}{-}                       & -                                            \\ \cline{2-8} 
\multicolumn{1}{|c|}{}                                        & Mistral - Gemma                                           & -                                                      & -                                                     & \multicolumn{1}{l|}{-}                     & \multicolumn{1}{l|}{-}                     & \multicolumn{1}{l|}{-}                       & -                                            \\ \cline{2-8} 
\multicolumn{1}{|c|}{}                                        & Qwen - Gemma                                              & -                                                      & 1.0066 {\ul \textbf{(Qwen)}}                                         & \multicolumn{1}{l|}{-}                     & \multicolumn{1}{l|}{-}                     & \multicolumn{1}{l|}{-}                       & -                                            \\ \clineB{2-8}{2} 
\multicolumn{1}{|c|}{}                                        & Llama - base                                              & -                                                      & -                                                     & \multicolumn{1}{l|}{0.4865}                & \multicolumn{1}{l|}{0.4759}                & \multicolumn{1}{l|}{-}                       & 0.5708                                       \\ \cline{2-8} 
\multicolumn{1}{|c|}{}                                        & Mistral - base                                            & 0.8753                                                 & -                                                     & \multicolumn{1}{l|}{0.808}                 & \multicolumn{1}{l|}{0.4591}                & \multicolumn{1}{l|}{0.5008}                  & -                                            \\ \cline{2-8} 
\multicolumn{1}{|c|}{}                                        & Qwen - base                                               & 0.5016                                                 & -                                                     & \multicolumn{1}{l|}{-}                     & \multicolumn{1}{l|}{-}                     & \multicolumn{1}{l|}{-}                       & -                                            \\ \cline{2-8} 
\multicolumn{1}{|c|}{}                                        & Gemma - base                                              & 0.6594                                                 & 1.0424                                                & \multicolumn{1}{l|}{0.543}                 & \multicolumn{1}{l|}{0.5901}                & \multicolumn{1}{l|}{{\ul \textbf{0.7504}}}   & -                                            \\ \hline
\end{tabular}
\end{sidewaystable*}

\begin{table*}[htbp]
\centering
\caption{\centering Results of statistical tests comparing safety changes across different PEFT methods, base models, and fine-tuning variables. For significant pairwise comparisons, the group with the higher mean is in {\ul \textbf{bold}}. }
\label{tab:safety_friedman_wilcoxon}
\renewcommand{\arraystretch}{1.2}
\footnotesize

\begin{tabular}{|cl|ll|}
\hline
\multicolumn{1}{|c|}{\multirow{3}{*}{\textbf{Factor}}}        & \multicolumn{1}{c|}{\multirow{3}{*}{\textbf{Comparison}}} & \multicolumn{2}{c|}{\multirow{2}{*}{\textbf{Safety}}}                                            \\
\multicolumn{1}{|c|}{}                                        & \multicolumn{1}{c|}{}                                     & \multicolumn{2}{c|}{}                                                                            \\ \cline{3-4} 
\multicolumn{1}{|c|}{}                                        & \multicolumn{1}{c|}{}                                     & \multicolumn{1}{c|}{\textbf{P-value}}                & \multicolumn{1}{c|}{\textbf{Effect Size}} \\ \hlineB{3}
\multicolumn{2}{|c|}{\textbf{All results}}                                                                                & \multicolumn{1}{l|}{-}                               & -                                         \\ \hlineB{3}
\multicolumn{1}{|c|}{\multirow{3}{*}{\textbf{Dataset}}}       & UC - UF                                                   & \multicolumn{1}{l|}{-}                               & -                                         \\ \cline{2-4} 
\multicolumn{1}{|c|}{}                                        & UC - base                                                 & \multicolumn{1}{l|}{-}                               & -                                         \\ \cline{2-4} 
\multicolumn{1}{|c|}{}                                        & UF - base                                                 & \multicolumn{1}{l|}{-}                               & -                                         \\ \hlineB{3}
\multicolumn{1}{|c|}{\multirow{3}{*}{\textbf{Paradigm}}}      & SFT - DPO                                                 & \multicolumn{1}{l|}{-}                               & -                                         \\ \cline{2-4} 
\multicolumn{1}{|c|}{}                                        & SFT - base                                                & \multicolumn{1}{l|}{p \textless 0.05}                & 0.2557                                    \\ \cline{2-4} 
\multicolumn{1}{|c|}{}                                        & DPO - base                                                & \multicolumn{1}{l|}{-}                               & -                                         \\ \hlineB{3}
\multicolumn{1}{|c|}{\multirow{3}{*}{\textbf{Learning Rate}}} & 1e-3 - 2e-5                                               & \multicolumn{1}{l|}{-}                               & -                                         \\ \cline{2-4} 
\multicolumn{1}{|c|}{}                                        & 1e-3 - base                                               & \multicolumn{1}{l|}{-}                               & -                                         \\ \cline{2-4} 
\multicolumn{1}{|c|}{}                                        & 2e-5 - base                                               & \multicolumn{1}{l|}{-}                               & -                                         \\ \hlineB{3}
\multicolumn{1}{|c|}{\multirow{3}{*}{\textbf{Epochs}}}        & 1e - 5e                                                   & \multicolumn{1}{l|}{-}                               & -                                         \\ \cline{2-4} 
\multicolumn{1}{|c|}{}                                        & 1e - base                                                 & \multicolumn{1}{l|}{-}                               & -                                         \\ \cline{2-4} 
\multicolumn{1}{|c|}{}                                        & 5e - base                                                 & \multicolumn{1}{l|}{-}                               & -                                         \\ \hlineB{3}
\multicolumn{1}{|c|}{\multirow{11}{*}{\textbf{Peft Method}}}  & All methods                                               & \multicolumn{1}{l|}{p \textless 0.05}                & 0.353                                     \\ \clineB{2-4}{3} 
\multicolumn{1}{|c|}{}                                        & Lora - IA$^3$                                                & \multicolumn{1}{l|}{-}                               & -                                         \\ \cline{2-4} 
\multicolumn{1}{|c|}{}                                        & IA$^3$ - Prompt-Tuning                                       & \multicolumn{1}{l|}{p \textless 0.05}                & 0.9397 {\ul\textbf{(IA$^3$)}}                               \\ \cline{2-4} 
\multicolumn{1}{|c|}{}                                        & IA$^3$ - P-Tuning                                            & \multicolumn{1}{l|}{p \textless 0.05}                & 0.9594 {\ul\textbf{(IA$^3$)}}                              \\ \cline{2-4} 
\multicolumn{1}{|c|}{}                                        & Lora - Prompt-Tuning                                      & \multicolumn{1}{l|}{-}                               & -                                         \\ \cline{2-4} 
\multicolumn{1}{|c|}{}                                        & Lora - P-Tuning                                           & \multicolumn{1}{l|}{p \textless 0.05}                & 0.7936 {\ul\textbf{(Lora)}}                             \\ \cline{2-4} 
\multicolumn{1}{|c|}{}                                        & Prompt - P-Tuning                                         & \multicolumn{1}{l|}{-}                               & -                                         \\ \clineB{2-4}{3} 
\multicolumn{1}{|c|}{}                                        & Lora - base                                               & \multicolumn{1}{l|}{{\ul \textbf{p = 0.0587}}}       & {\ul \textbf{0.4031}}                     \\ \cline{2-4} 
\multicolumn{1}{|c|}{}                                        & IA$^3$ - base                                                & \multicolumn{1}{l|}{{\ul \textbf{p \textless 0.05}}} & {\ul \textbf{0.641}}                      \\ \cline{2-4} 
\multicolumn{1}{|c|}{}                                        & Prompt-Tuning - base                                      & \multicolumn{1}{l|}{p \textless 0.05}                & 0.7309                                    \\ \cline{2-4} 
\multicolumn{1}{|c|}{}                                        & P-Tuning - base                                           & \multicolumn{1}{l|}{p \textless 0.05}                & 0.7179                                    \\ \hlineB{3}
\multicolumn{1}{|c|}{\multirow{11}{*}{\textbf{Model}}}        & All models                                                & \multicolumn{1}{l|}{p \textless 0.05}                & 0.24                                      \\ \clineB{2-4}{3} 
\multicolumn{1}{|c|}{}                                        & Llama - Mistral                                           & \multicolumn{1}{l|}{-}                               & -                                         \\ \cline{2-4} 
\multicolumn{1}{|c|}{}                                        & Llama - Qwen                                              & \multicolumn{1}{l|}{-}                               & -                                         \\ \cline{2-4} 
\multicolumn{1}{|c|}{}                                        & Llama - Gemma                                             & \multicolumn{1}{l|}{-}                               & -                                         \\ \cline{2-4} 
\multicolumn{1}{|c|}{}                                        & Mistral - Qwen                                            & \multicolumn{1}{l|}{-}                               & -                                         \\ \cline{2-4} 
\multicolumn{1}{|c|}{}                                        & Mistral - Gemma                                           & \multicolumn{1}{l|}{-}                               & -                                         \\ \cline{2-4} 
\multicolumn{1}{|c|}{}                                        & Qwen - Gemma                                              & \multicolumn{1}{l|}{p \textless 0.05}                & 1.0066 {\ul\textbf{(Qwen)}}                             \\ \clineB{2-4}{3} 
\multicolumn{1}{|c|}{}                                        & Llama - base                                              & \multicolumn{1}{l|}{-}                               & -                                         \\ \cline{2-4} 
\multicolumn{1}{|c|}{}                                        & Mistral - base                                            & \multicolumn{1}{l|}{-}                               & -                                         \\ \cline{2-4} 
\multicolumn{1}{|c|}{}                                        & Qwen - base                                               & \multicolumn{1}{l|}{-}                               & -                                         \\ \cline{2-4} 
\multicolumn{1}{|c|}{}                                        & Gemma - base                                              & \multicolumn{1}{l|}{p \textless 0.05}                & 1.0424                                    \\ \hline
\end{tabular}

\end{table*}

\begin{sidewaystable*}[htbp]
\caption{\centering Results of the statistical tests for all safety categories}
\label{tab:safety_categories}
\renewcommand{\arraystretch}{1.1}
\centering
\tiny

\begin{tabular}{|cl|lcccccccccc|}
\hline
\multicolumn{1}{|c|}{\multirow{3}{*}{\textbf{Factor}}}                                                   & \multicolumn{1}{c|}{\multirow{3}{*}{\textbf{Comparison}}}          & \multicolumn{11}{c|}{\textbf{Safety Categories}}                                                                                                                                                                                                                                                                                                                                                                                                                                                                                                                                                                                                                                                                                                                                                                                                                                                                                                                                                                                                                   \\ \cline{3-13} 
\multicolumn{1}{|c|}{}                                                                                   & \multicolumn{1}{c|}{}                                              & \multicolumn{1}{c|}{\textbf{\begin{tabular}[c]{@{}c@{}}1. Illegal\\ Activity\end{tabular}}} & \multicolumn{1}{c|}{\textbf{\begin{tabular}[c]{@{}c@{}}2. Child\\ Abuse\\ Content\end{tabular}}} & \multicolumn{1}{c|}{\textbf{\begin{tabular}[c]{@{}c@{}}3. Hate/\\ Harass/\\ Violence\end{tabular}}} & \multicolumn{1}{c|}{\textbf{4. Malware}}                                            & \multicolumn{1}{c|}{\textbf{\begin{tabular}[c]{@{}c@{}}5. Physical\\ Harm\end{tabular}}} & \multicolumn{1}{c|}{\textbf{\begin{tabular}[c]{@{}c@{}}6. Economic\\ Harm\end{tabular}}} & \multicolumn{1}{c|}{\textbf{\begin{tabular}[c]{@{}c@{}}7. Fraud /\\ Deception\end{tabular}}} & \multicolumn{1}{c|}{\textbf{\begin{tabular}[c]{@{}c@{}}8. Adult\\ Content\end{tabular}}} & \multicolumn{1}{c|}{\textbf{\begin{tabular}[c]{@{}c@{}}9. Political\\ Campaigning\end{tabular}}} & \multicolumn{1}{c|}{\textbf{\begin{tabular}[c]{@{}c@{}}10. Privacy\\ Violation\end{tabular}}} & \textbf{\begin{tabular}[c]{@{}c@{}}11. Tailored\\ Financial\\ Advice\end{tabular}} \\ \cline{3-13} 
\multicolumn{1}{|c|}{}                                                                                   & \multicolumn{1}{c|}{}                                              & \multicolumn{1}{c|}{\textbf{\begin{tabular}[c]{@{}c@{}}Effect\\ Size\end{tabular}}}         & \multicolumn{1}{c|}{\textbf{\begin{tabular}[c]{@{}c@{}}Effect\\ Size\end{tabular}}}              & \multicolumn{1}{c|}{\textbf{\begin{tabular}[c]{@{}c@{}}Effect\\ Size\end{tabular}}}                 & \multicolumn{1}{c|}{\textbf{\begin{tabular}[c]{@{}c@{}}Effect\\ Size\end{tabular}}} & \multicolumn{1}{c|}{\textbf{\begin{tabular}[c]{@{}c@{}}Effect\\ Size\end{tabular}}}      & \multicolumn{1}{c|}{\textbf{\begin{tabular}[c]{@{}c@{}}Effect\\ Size\end{tabular}}}      & \multicolumn{1}{c|}{\textbf{\begin{tabular}[c]{@{}c@{}}Effect\\ Size\end{tabular}}}          & \multicolumn{1}{c|}{\textbf{\begin{tabular}[c]{@{}c@{}}Effect\\ Size\end{tabular}}}      & \multicolumn{1}{c|}{\textbf{\begin{tabular}[c]{@{}c@{}}Effect \\ Size\end{tabular}}}             & \multicolumn{1}{c|}{\textbf{\begin{tabular}[c]{@{}c@{}}Effect\\ Size\end{tabular}}}           & \textbf{\begin{tabular}[c]{@{}c@{}}Effect\\ Size\end{tabular}}                     \\ \hline
\multicolumn{2}{|c|}{\textbf{All results}}                                                                                                                                    & \multicolumn{1}{l|}{-}                                                                      & \multicolumn{1}{c|}{0.8726}                                                                      & \multicolumn{1}{c|}{-}                                                                              & \multicolumn{1}{c|}{{\ul \textbf{0.2366}}}                                          & \multicolumn{1}{c|}{-}                                                                   & \multicolumn{1}{c|}{-}                                                                   & \multicolumn{1}{c|}{0.5099}                                                                  & \multicolumn{1}{c|}{0.4998}                                                              & \multicolumn{1}{c|}{-}                                                                           & \multicolumn{1}{c|}{0.2384}                                                                   & -                                                                                  \\ \hline
\multicolumn{1}{|c|}{\multirow{3}{*}{\textbf{Dataset}}}                                                  & UC - UF                                                            & \multicolumn{1}{l|}{-}                                                                      & \multicolumn{1}{c|}{-}                                                                           & \multicolumn{1}{c|}{-}                                                                              & \multicolumn{1}{c|}{-}                                                              & \multicolumn{1}{c|}{-}                                                                   & \multicolumn{1}{c|}{-}                                                                   & \multicolumn{1}{c|}{-}                                                                       & \multicolumn{1}{c|}{-}                                                                   & \multicolumn{1}{c|}{-}                                                                           & \multicolumn{1}{c|}{-}                                                                        & -                                                                                  \\ \cline{2-13} 
\multicolumn{1}{|c|}{}                                                                                   & UC - base                                                          & \multicolumn{1}{l|}{-}                                                                      & \multicolumn{1}{c|}{0.8911}                                                                      & \multicolumn{1}{c|}{-}                                                                              & \multicolumn{1}{c|}{-}                                                              & \multicolumn{1}{c|}{-}                                                                   & \multicolumn{1}{c|}{-}                                                                   & \multicolumn{1}{c|}{0.6131}                                                                  & \multicolumn{1}{c|}{0.6169}                                                              & \multicolumn{1}{c|}{-}                                                                           & \multicolumn{1}{c|}{-}                                                                        & -                                                                                  \\ \cline{2-13} 
\multicolumn{1}{|c|}{}                                                                                   & UF - base                                                          & \multicolumn{1}{l|}{-}                                                                      & \multicolumn{1}{c|}{0.8743}                                                                      & \multicolumn{1}{c|}{-}                                                                              & \multicolumn{1}{c|}{{\ul \textbf{0.2911}}}                                          & \multicolumn{1}{c|}{-}                                                                   & \multicolumn{1}{c|}{-}                                                                   & \multicolumn{1}{c|}{0.4991}                                                                  & \multicolumn{1}{c|}{0.4724}                                                              & \multicolumn{1}{c|}{-}                                                                           & \multicolumn{1}{c|}{-}                                                                        & -                                                                                  \\ \hline
\multicolumn{1}{|c|}{\multirow{3}{*}{\textbf{Paradigm}}}                                                 & SFT - DPO                                                          & \multicolumn{1}{l|}{-}                                                                      & \multicolumn{1}{c|}{\begin{tabular}[c]{@{}c@{}}0.7094\\ (DPO)\end{tabular}}                      & \multicolumn{1}{c|}{\begin{tabular}[c]{@{}c@{}}0.57\\ (DPO)\end{tabular}}                           & \multicolumn{1}{c|}{-}                                                              & \multicolumn{1}{c|}{-}                                                                   & \multicolumn{1}{c|}{-}                                                                   & \multicolumn{1}{c|}{-}                                                                       & \multicolumn{1}{c|}{-}                                                                   & \multicolumn{1}{l|}{}                                                                            & \multicolumn{1}{c|}{-}                                                                        & -                                                                                  \\ \cline{2-13} 
\multicolumn{1}{|c|}{}                                                                                   & SFT - base                                                         & \multicolumn{1}{l|}{-}                                                                      & \multicolumn{1}{c|}{0.8739}                                                                      & \multicolumn{1}{c|}{0.3957}                                                                         & \multicolumn{1}{c|}{-}                                                              & \multicolumn{1}{c|}{-}                                                                   & \multicolumn{1}{c|}{-}                                                                   & \multicolumn{1}{c|}{0.5937}                                                                  & \multicolumn{1}{c|}{0.5191}                                                              & \multicolumn{1}{c|}{0.2829}                                                                      & \multicolumn{1}{c|}{0.3191}                                                                   & -                                                                                  \\ \cline{2-13} 
\multicolumn{1}{|c|}{}                                                                                   & DPO - base                                                         & \multicolumn{1}{l|}{-}                                                                      & \multicolumn{1}{c|}{0.8944}                                                                      & \multicolumn{1}{c|}{-}                                                                              & \multicolumn{1}{c|}{{\ul \textbf{0.6929}}}                                          & \multicolumn{1}{c|}{-}                                                                   & \multicolumn{1}{c|}{-}                                                                   & \multicolumn{1}{c|}{-}                                                                       & \multicolumn{1}{c|}{-}                                                                   & \multicolumn{1}{c|}{-}                                                                           & \multicolumn{1}{c|}{-}                                                                        & -                                                                                  \\ \hline
\multicolumn{1}{|c|}{\multirow{3}{*}{\textbf{\begin{tabular}[c]{@{}c@{}}Learning \\ Rate\end{tabular}}}} & 1e-3 - 2e-5                                                        & \multicolumn{1}{c|}{-}                                                                      & \multicolumn{1}{c|}{-}                                                                           & \multicolumn{1}{c|}{-}                                                                              & \multicolumn{1}{c|}{-}                                                              & \multicolumn{1}{c|}{-}                                                                   & \multicolumn{1}{c|}{-}                                                                   & \multicolumn{1}{c|}{-}                                                                       & \multicolumn{1}{c|}{-}                                                                   & \multicolumn{1}{c|}{-}                                                                           & \multicolumn{1}{c|}{0.425 (1e-3)}                                                             & -                                                                                  \\ \cline{2-13} 
\multicolumn{1}{|c|}{}                                                                                   & 1e-3 - base                                                        & \multicolumn{1}{c|}{-}                                                                      & \multicolumn{1}{c|}{0.8843}                                                                      & \multicolumn{1}{c|}{-}                                                                              & \multicolumn{1}{c|}{{\ul \textbf{0.4384}}}                                          & \multicolumn{1}{c|}{-}                                                                   & \multicolumn{1}{c|}{-}                                                                   & \multicolumn{1}{c|}{0.5226}                                                                  & \multicolumn{1}{c|}{0.4742}                                                              & \multicolumn{1}{c|}{-}                                                                           & \multicolumn{1}{c|}{-}                                                                        & -                                                                                  \\ \cline{2-13} 
\multicolumn{1}{|c|}{}                                                                                   & 2e-5 - base                                                        & \multicolumn{1}{c|}{-}                                                                      & \multicolumn{1}{c|}{0.8753}                                                                      & \multicolumn{1}{c|}{-}                                                                              & \multicolumn{1}{c|}{-}                                                              & \multicolumn{1}{c|}{-}                                                                   & \multicolumn{1}{c|}{-}                                                                   & \multicolumn{1}{c|}{0.5095}                                                                  & \multicolumn{1}{c|}{0.5132}                                                              & \multicolumn{1}{c|}{-}                                                                           & \multicolumn{1}{c|}{0.3944}                                                                   & -                                                                                  \\ \hline
\multicolumn{1}{|c|}{\multirow{3}{*}{\textbf{Epochs}}}                                                   & 1e - 5e                                                            & \multicolumn{1}{c|}{-}                                                                      & \multicolumn{1}{c|}{-}                                                                           & \multicolumn{1}{c|}{-}                                                                              & \multicolumn{1}{c|}{-}                                                              & \multicolumn{1}{c|}{-}                                                                   & \multicolumn{1}{c|}{-}                                                                   & \multicolumn{1}{c|}{-}                                                                       & \multicolumn{1}{c|}{-}                                                                   & \multicolumn{1}{c|}{-}                                                                           & \multicolumn{1}{c|}{-}                                                                        & -                                                                                  \\ \cline{2-13} 
\multicolumn{1}{|c|}{}                                                                                   & 1e - base                                                          & \multicolumn{1}{c|}{-}                                                                      & \multicolumn{1}{c|}{0.8815}                                                                      & \multicolumn{1}{c|}{-}                                                                              & \multicolumn{1}{c|}{-}                                                              & \multicolumn{1}{c|}{-}                                                                   & \multicolumn{1}{c|}{-}                                                                   & \multicolumn{1}{c|}{0.6206}                                                                  & \multicolumn{1}{c|}{0.6264}                                                              & \multicolumn{1}{c|}{-}                                                                           & \multicolumn{1}{c|}{0.4454}                                                                   & -                                                                                  \\ \cline{2-13} 
\multicolumn{1}{|c|}{}                                                                                   & 5e - base                                                          & \multicolumn{1}{c|}{-}                                                                      & \multicolumn{1}{c|}{0.8761}                                                                      & \multicolumn{1}{c|}{-}                                                                              & \multicolumn{1}{c|}{{\ul \textbf{0.3697}}}                                          & \multicolumn{1}{c|}{-}                                                                   & \multicolumn{1}{c|}{-}                                                                   & \multicolumn{1}{c|}{0.4759}                                                                  & \multicolumn{1}{c|}{0.4258}                                                              & \multicolumn{1}{c|}{-}                                                                           & \multicolumn{1}{c|}{-}                                                                        & -                                                                                  \\ \hline
\multicolumn{1}{|c|}{\multirow{11}{*}{\textbf{\begin{tabular}[c]{@{}c@{}}PEFT \\ Method\end{tabular}}}}  & All methods                                                        & \multicolumn{1}{l|}{0.266}                                                                  & \multicolumn{1}{c|}{0.326}                                                                       & \multicolumn{1}{c|}{0.239}                                                                          & \multicolumn{1}{c|}{0.314}                                                          & \multicolumn{1}{c|}{0.135}                                                               & \multicolumn{1}{c|}{0.386}                                                               & \multicolumn{1}{c|}{0.306}                                                                   & \multicolumn{1}{c|}{0.328}                                                               & \multicolumn{1}{c|}{0.189}                                                                       & \multicolumn{1}{c|}{0.214}                                                                    & 0.353                                                                              \\ \cline{2-13} 
\multicolumn{1}{|c|}{}                                                                                   & Lora - IA$^3$                                                         & \multicolumn{1}{l|}{\begin{tabular}[c]{@{}l@{}}0.8605\\ (Lora)\end{tabular}}                & \multicolumn{1}{c|}{-}                                                                           & \multicolumn{1}{c|}{-}                                                                              & \multicolumn{1}{c|}{-}                                                              & \multicolumn{1}{c|}{-}                                                                   & \multicolumn{1}{c|}{-}                                                                   & \multicolumn{1}{c|}{-}                                                                       & \multicolumn{1}{c|}{-}                                                                   & \multicolumn{1}{c|}{-}                                                                           & \multicolumn{1}{c|}{-}                                                                        & -                                                                                  \\ \cline{2-13} 
\multicolumn{1}{|c|}{}                                                                                   & \begin{tabular}[c]{@{}l@{}}IA$^3$ - \\ Prompt \\ Tuning\end{tabular}  & \multicolumn{1}{l|}{-}                                                                      & \multicolumn{1}{c|}{-}                                                                           & \multicolumn{1}{c|}{\begin{tabular}[c]{@{}c@{}}0.8839\\ (IA$^3$)\end{tabular}}                         & \multicolumn{1}{c|}{-}                                                              & \multicolumn{1}{c|}{\begin{tabular}[c]{@{}c@{}}0.7973\\ (IA$^3$)\end{tabular}}              & \multicolumn{1}{c|}{\begin{tabular}[c]{@{}c@{}}0.8612\\ (IA$^3$)\end{tabular}}              & \multicolumn{1}{c|}{\begin{tabular}[c]{@{}c@{}}0.7756\\ (IA$^3$)\end{tabular}}                  & \multicolumn{1}{c|}{\begin{tabular}[c]{@{}c@{}}0.8643\\ (IA$^3$)\end{tabular}}              & \multicolumn{1}{c|}{-}                                                                           & \multicolumn{1}{c|}{-}                                                                        & \begin{tabular}[c]{@{}c@{}}0.8214\\ (IA$^3$)\end{tabular}                             \\ \cline{2-13} 
\multicolumn{1}{|c|}{}                                                                                   & \begin{tabular}[c]{@{}l@{}}IA$^3$ - \\ P-Tuning\end{tabular}          & \multicolumn{1}{l|}{-}                                                                      & \multicolumn{1}{c|}{\begin{tabular}[c]{@{}c@{}}0.8833\\ (IA$^3$)\end{tabular}}                      & \multicolumn{1}{c|}{\begin{tabular}[c]{@{}c@{}}0.8221\\ (IA$^3$)\end{tabular}}                         & \multicolumn{1}{c|}{\begin{tabular}[c]{@{}c@{}}1.0113\\ (IA$^3$)\end{tabular}}         & \multicolumn{1}{c|}{-}                                                                   & \multicolumn{1}{c|}{\begin{tabular}[c]{@{}c@{}}0.8670\\ (IA$^3$)\end{tabular}}              & \multicolumn{1}{c|}{\begin{tabular}[c]{@{}c@{}}0.8612\\ (IA$^3$)\end{tabular}}                  & \multicolumn{1}{c|}{\begin{tabular}[c]{@{}c@{}}0.8813\\ (IA$^3$)\end{tabular}}              & \multicolumn{1}{c|}{\begin{tabular}[c]{@{}c@{}}0.7005\\ (IA$^3$)\end{tabular}}                      & \multicolumn{1}{c|}{-}                                                                        & \begin{tabular}[c]{@{}c@{}}0.6776\\ (IA$^3$)\end{tabular}                             \\ \cline{2-13} 
\multicolumn{1}{|c|}{}                                                                                   & \begin{tabular}[c]{@{}l@{}}Lora - \\ Prompt \\ Tuning\end{tabular} & \multicolumn{1}{l|}{\begin{tabular}[c]{@{}l@{}}0.7924\\ (Lora)\end{tabular}}                & \multicolumn{1}{c|}{-}                                                                           & \multicolumn{1}{c|}{-}                                                                              & \multicolumn{1}{c|}{-}                                                              & \multicolumn{1}{c|}{-}                                                                   & \multicolumn{1}{c|}{-}                                                                   & \multicolumn{1}{c|}{-}                                                                       & \multicolumn{1}{c|}{-}                                                                   & \multicolumn{1}{c|}{-}                                                                           & \multicolumn{1}{c|}{\begin{tabular}[c]{@{}c@{}}0.802\\ (Lora)\end{tabular}}                   & \begin{tabular}[c]{@{}c@{}}0.7292\\ (Lora)\end{tabular}                            \\ \cline{2-13} 
\multicolumn{1}{|c|}{}                                                                                   & \begin{tabular}[c]{@{}l@{}}Lora - \\ P-Tuning\end{tabular}         & \multicolumn{1}{l|}{\begin{tabular}[c]{@{}l@{}}0.8216\\ (Lora)\end{tabular}}                & \multicolumn{1}{c|}{\begin{tabular}[c]{@{}c@{}}0.8565\\ (Lora)\end{tabular}}                     & \multicolumn{1}{c|}{-}                                                                              & \multicolumn{1}{c|}{\begin{tabular}[c]{@{}c@{}}0.8979\\ (Lora)\end{tabular}}        & \multicolumn{1}{c|}{-}                                                                   & \multicolumn{1}{c|}{\begin{tabular}[c]{@{}c@{}}0.7004\\ (Lora)\end{tabular}}             & \multicolumn{1}{c|}{\begin{tabular}[c]{@{}c@{}}0.6775\\ (Lora)\end{tabular}}                 & \multicolumn{1}{c|}{\begin{tabular}[c]{@{}c@{}}0.7620\\ (Lora)\end{tabular}}             & \multicolumn{1}{c|}{\begin{tabular}[c]{@{}c@{}}0.6851\\ (Lora)\end{tabular}}                     & \multicolumn{1}{c|}{\begin{tabular}[c]{@{}c@{}}0.7306\\ (Lora)\end{tabular}}                  & \begin{tabular}[c]{@{}c@{}}0.7178\\ (Lora)\end{tabular}                            \\ \cline{2-13} 
\multicolumn{1}{|c|}{}                                                                                   & \begin{tabular}[c]{@{}l@{}}Prompt - \\ P-Tuning\end{tabular}       & \multicolumn{1}{l|}{-}                                                                      & \multicolumn{1}{c|}{-}                                                                           & \multicolumn{1}{c|}{-}                                                                              & \multicolumn{1}{c|}{\begin{tabular}[c]{@{}c@{}}0.879\\ (Prompt)\end{tabular}}       & \multicolumn{1}{c|}{-}                                                                   & \multicolumn{1}{c|}{-}                                                                   & \multicolumn{1}{c|}{\begin{tabular}[c]{@{}c@{}}0.6866\\ (Prompt)\end{tabular}}               & \multicolumn{1}{c|}{-}                                                                   & \multicolumn{1}{c|}{-}                                                                           & \multicolumn{1}{c|}{-}                                                                        & -                                                                                  \\ \cline{2-13} 
\multicolumn{1}{|c|}{}                                                                                   & Lora - base                                                        & \multicolumn{1}{l|}{{\ul \textbf{0.8439}}}                                                  & \multicolumn{1}{c|}{0.9023}                                                                      & \multicolumn{1}{c|}{-}                                                                              & \multicolumn{1}{c|}{{\ul \textbf{0.6894}}}                                          & \multicolumn{1}{c|}{{\ul \textbf{0.6709}}}                                               & \multicolumn{1}{c|}{-}                                                                   & \multicolumn{1}{c|}{-}                                                                       & \multicolumn{1}{c|}{-}                                                                   & \multicolumn{1}{c|}{-}                                                                           & \multicolumn{1}{c|}{-}                                                                        & {\ul \textbf{0.4818}}                                                              \\ \cline{2-13} 
\multicolumn{1}{|c|}{}                                                                                   & IA$^3$ - base                                                         & \multicolumn{1}{l|}{-}                                                                      & \multicolumn{1}{c|}{0.896}                                                                       & \multicolumn{1}{c|}{{\ul \textbf{0.6522}}}                                                          & \multicolumn{1}{c|}{{\ul \textbf{0.8307}}}                                          & \multicolumn{1}{c|}{{\ul \textbf{0.8457}}}                                               & \multicolumn{1}{c|}{{\ul \textbf{0.7902}}}                                               & \multicolumn{1}{c|}{-}                                                                       & \multicolumn{1}{c|}{-}                                                                   & \multicolumn{1}{c|}{{\ul \textbf{0.715}}}                                                        & \multicolumn{1}{c|}{-}                                                                        & -                                                                                  \\ \cline{2-13} 
\multicolumn{1}{|c|}{}                                                                                   & \begin{tabular}[c]{@{}l@{}}Prompt \\ Tuning\\  - base\end{tabular} & \multicolumn{1}{l|}{-}                                                                      & \multicolumn{1}{c|}{0.8885}                                                                      & \multicolumn{1}{c|}{0.779}                                                                          & \multicolumn{1}{c|}{-}                                                              & \multicolumn{1}{c|}{0.6857}                                                              & \multicolumn{1}{c|}{0.688}                                                               & \multicolumn{1}{c|}{0.7138}                                                                  & \multicolumn{1}{c|}{-}                                                                   & \multicolumn{1}{c|}{-}                                                                           & \multicolumn{1}{c|}{0.6723}                                                                   & 0.6622                                                                             \\ \cline{2-13} 
\multicolumn{1}{|c|}{}                                                                                   & \begin{tabular}[c]{@{}l@{}}P-Tuning\\  - base\end{tabular}         & \multicolumn{1}{l|}{0.5693}                                                                 & \multicolumn{1}{c|}{0.8821}                                                                      & \multicolumn{1}{c|}{0.5707}                                                                         & \multicolumn{1}{c|}{0.5426}                                                         & \multicolumn{1}{c|}{0.5366}                                                              & \multicolumn{1}{c|}{0.6448}                                                              & \multicolumn{1}{c|}{0.8182}                                                                  & \multicolumn{1}{c|}{0.7627}                                                              & \multicolumn{1}{c|}{0.6518}                                                                      & \multicolumn{1}{c|}{0.5542}                                                                   & 0.4842                                                                             \\ \hline
\multicolumn{1}{|c|}{\multirow{11}{*}{\textbf{Model}}}                                                   & All models                                                         & \multicolumn{1}{l|}{-}                                                                      & \multicolumn{1}{l|}{0.456}                                                                       & \multicolumn{1}{c|}{-}                                                                              & \multicolumn{1}{c|}{0.426}                                                          & \multicolumn{1}{c|}{-}                                                                   & \multicolumn{1}{c|}{0.305}                                                               & \multicolumn{1}{c|}{0.282}                                                                   & \multicolumn{1}{c|}{0.373}                                                               & \multicolumn{1}{c|}{0.336}                                                                       & \multicolumn{1}{c|}{0.14}                                                                     & 0.212                                                                              \\ \cline{2-13} 
\multicolumn{1}{|c|}{}                                                                                   & \begin{tabular}[c]{@{}l@{}}Llama - \\ Mistral\end{tabular}         & \multicolumn{1}{l|}{-}                                                                      & \multicolumn{1}{l|}{\begin{tabular}[c]{@{}l@{}}1.0066\\ (Llama)\end{tabular}}                    & \multicolumn{1}{l|}{-}                                                                              & \multicolumn{1}{c|}{\begin{tabular}[c]{@{}c@{}}1.0066\\ (Mistral)\end{tabular}}     & \multicolumn{1}{l|}{-}                                                                   & \multicolumn{1}{l|}{\begin{tabular}[c]{@{}l@{}}0.8125\\ (Mistral)\end{tabular}}          & \multicolumn{1}{l|}{-}                                                                       & \multicolumn{1}{l|}{-}                                                                   & \multicolumn{1}{l|}{-}                                                                           & \multicolumn{1}{l|}{-}                                                                        & \multicolumn{1}{l|}{\begin{tabular}[c]{@{}l@{}}0.8451\\ (Llama)\end{tabular}}      \\ \cline{2-13} 
\multicolumn{1}{|c|}{}                                                                                   & \begin{tabular}[c]{@{}l@{}}Llama - \\ Qwen\end{tabular}            & \multicolumn{1}{l|}{-}                                                                      & \multicolumn{1}{l|}{-}                                                                           & \multicolumn{1}{l|}{-}                                                                              & \multicolumn{1}{c|}{\begin{tabular}[c]{@{}c@{}}0.7741\\ (Qwen)\end{tabular}}        & \multicolumn{1}{l|}{-}                                                                   & \multicolumn{1}{l|}{-}                                                                   & \multicolumn{1}{l|}{-}                                                                       & \multicolumn{1}{l|}{-}                                                                   & \multicolumn{1}{l|}{-}                                                                           & \multicolumn{1}{l|}{-}                                                                        & \multicolumn{1}{l|}{-}                                                             \\ \cline{2-13} 
\multicolumn{1}{|c|}{}                                                                                   & \begin{tabular}[c]{@{}l@{}}Llama - \\ Gemma\end{tabular}           & \multicolumn{1}{l|}{-}                                                                      & \multicolumn{1}{l|}{-}                                                                           & \multicolumn{1}{l|}{-}                                                                              & \multicolumn{1}{c|}{-}                                                              & \multicolumn{1}{l|}{-}                                                                   & \multicolumn{1}{l|}{-}                                                                   & \multicolumn{1}{l|}{-}                                                                       & \multicolumn{1}{l|}{-}                                                                   & \multicolumn{1}{l|}{-}                                                                           & \multicolumn{1}{l|}{-}                                                                        & \multicolumn{1}{l|}{-}                                                             \\ \cline{2-13} 
\multicolumn{1}{|c|}{}                                                                                   & \begin{tabular}[c]{@{}l@{}}Mistral\\  - Qwen\end{tabular}          & \multicolumn{1}{l|}{-}                                                                      & \multicolumn{1}{l|}{\begin{tabular}[c]{@{}l@{}}1.0066\\ (Qwen)\end{tabular}}                     & \multicolumn{1}{l|}{-}                                                                              & \multicolumn{1}{c|}{-}                                                              & \multicolumn{1}{l|}{-}                                                                   & \multicolumn{1}{l|}{-}                                                                   & \multicolumn{1}{l|}{\begin{tabular}[c]{@{}l@{}}0.8223\\ (Mistral)\end{tabular}}              & \multicolumn{1}{l|}{\begin{tabular}[c]{@{}l@{}}0.9518\\ (Qwen)\end{tabular}}             & \multicolumn{1}{l|}{-}                                                                           & \multicolumn{1}{l|}{-}                                                                        & \multicolumn{1}{l|}{-}                                                             \\ \cline{2-13} 
\multicolumn{1}{|c|}{}                                                                                   & \begin{tabular}[c]{@{}l@{}}Mistral - \\ Gemma\end{tabular}         & \multicolumn{1}{l|}{-}                                                                      & \multicolumn{1}{l|}{\begin{tabular}[c]{@{}l@{}}0.8748\\ (Gemma)\end{tabular}}                    & \multicolumn{1}{l|}{-}                                                                              & \multicolumn{1}{c|}{\begin{tabular}[c]{@{}c@{}}0.9184\\ (Mistral)\end{tabular}}     & \multicolumn{1}{l|}{-}                                                                   & \multicolumn{1}{l|}{\begin{tabular}[c]{@{}l@{}}1.0066\\ (Mistral)\end{tabular}}          & \multicolumn{1}{l|}{\begin{tabular}[c]{@{}l@{}}1.0066\\ (Mistral)\end{tabular}}              & \multicolumn{1}{l|}{-}                                                                   & \multicolumn{1}{l|}{-}                                                                           & \multicolumn{1}{l|}{-}                                                                        & \multicolumn{1}{l|}{-}                                                             \\ \cline{2-13} 
\multicolumn{1}{|c|}{}                                                                                   & \begin{tabular}[c]{@{}l@{}}Qwen - \\ Gemma\end{tabular}            & \multicolumn{1}{l|}{-}                                                                      & \multicolumn{1}{l|}{-}                                                                           & \multicolumn{1}{l|}{-}                                                                              & \multicolumn{1}{c|}{\begin{tabular}[c]{@{}c@{}}1.0066\\ (Qwen)\end{tabular}}        & \multicolumn{1}{l|}{-}                                                                   & \multicolumn{1}{l|}{\begin{tabular}[c]{@{}l@{}}0.8588\\ (Qwen)\end{tabular}}             & \multicolumn{1}{l|}{-}                                                                       & \multicolumn{1}{l|}{\begin{tabular}[c]{@{}l@{}}0.9184\\ (Qwen)\end{tabular}}             & \multicolumn{1}{l|}{\begin{tabular}[c]{@{}l@{}}1.0066\\ (Qwen)\end{tabular}}                     & \multicolumn{1}{l|}{-}                                                                        & \multicolumn{1}{l|}{-}                                                             \\ \cline{2-13} 
\multicolumn{1}{|c|}{}                                                                                   & \begin{tabular}[c]{@{}l@{}}Llama - \\ base\end{tabular}            & \multicolumn{1}{l|}{-}                                                                      & \multicolumn{1}{l|}{-}                                                                           & \multicolumn{1}{l|}{-}                                                                              & \multicolumn{1}{c|}{0.887}                                                          & \multicolumn{1}{c|}{-}                                                                   & \multicolumn{1}{c|}{-}                                                                   & \multicolumn{1}{c|}{-}                                                                       & \multicolumn{1}{c|}{-}                                                                   & \multicolumn{1}{c|}{0.8922}                                                                      & \multicolumn{1}{c|}{-}                                                                        & -                                                                                  \\ \cline{2-13} 
\multicolumn{1}{|c|}{}                                                                                   & \begin{tabular}[c]{@{}l@{}}Mistral - \\ base\end{tabular}          & \multicolumn{1}{l|}{-}                                                                      & \multicolumn{1}{l|}{0.876}                                                                       & \multicolumn{1}{l|}{-}                                                                              & \multicolumn{1}{c|}{{\ul \textbf{0.6342}}}                                          & \multicolumn{1}{c|}{-}                                                                   & \multicolumn{1}{c|}{{\ul \textbf{0.5011}}}                                               & \multicolumn{1}{c|}{-}                                                                       & \multicolumn{1}{c|}{0.8184}                                                              & \multicolumn{1}{c|}{-}                                                                           & \multicolumn{1}{c|}{-}                                                                        & -                                                                                  \\ \cline{2-13} 
\multicolumn{1}{|c|}{}                                                                                   & \begin{tabular}[c]{@{}l@{}}Qwen - \\ base\end{tabular}             & \multicolumn{1}{l|}{-}                                                                      & \multicolumn{1}{l|}{0.9129}                                                                      & \multicolumn{1}{l|}{0.8875}                                                                         & \multicolumn{1}{c|}{{\ul \textbf{0.6682}}}                                          & \multicolumn{1}{c|}{-}                                                                   & \multicolumn{1}{c|}{-}                                                                   & \multicolumn{1}{c|}{0.7623}                                                                  & \multicolumn{1}{c|}{{\ul \textbf{0.5926}}}                                               & \multicolumn{1}{c|}{{\ul \textbf{0.5027}}}                                                       & \multicolumn{1}{c|}{0.8828}                                                                   & -                                                                                  \\ \cline{2-13} 
\multicolumn{1}{|c|}{}                                                                                   & \begin{tabular}[c]{@{}l@{}}Gemma - \\ base\end{tabular}            & \multicolumn{1}{l|}{0.8922}                                                                 & \multicolumn{1}{l|}{0.8917}                                                                      & \multicolumn{1}{l|}{0.8924}                                                                         & \multicolumn{1}{c|}{0.7902}                                                         & \multicolumn{1}{c|}{0.8881}                                                              & \multicolumn{1}{c|}{0.7638}                                                              & \multicolumn{1}{c|}{0.8798}                                                                  & \multicolumn{1}{c|}{0.8893}                                                              & \multicolumn{1}{c|}{0.8904}                                                                      & \multicolumn{1}{c|}{0.8856}                                                                   & 0.6973                                                                             \\ \hline
\end{tabular}
\end{sidewaystable*}

\begin{table*}[htbp]
\scriptsize
\centering
\caption{\centering Results of statistical tests comparing fairness metrics changes across different fine-tuning variables. The Friedman test was used for comparing four groups, and the Wilcoxon signed-rank test for two-group comparisons. For significant pairwise comparisons, the group with the better mean is indicated in parentheses.}
\label{tab:fairness_friedman_wilcoxon}
\renewcommand{\arraystretch}{1.05}

\begin{tabular}{|cl|llllllll|}
\hline
\multicolumn{1}{|c|}{\multirow{3}{*}{\textbf{Factor}}}                                                   & \multicolumn{1}{c|}{\multirow{3}{*}{\textbf{Comparison}}}       & \multicolumn{8}{c|}{\textbf{Fairness}}                                                                                                                                                                                                                                                                                                                                                                                                                                                                                                          \\ \cline{3-10} 
\multicolumn{1}{|c|}{}                                                                                   & \multicolumn{1}{c|}{}                                           & \multicolumn{2}{c|}{\textbf{Accuracy AMB}}                                                                                                & \multicolumn{2}{c|}{\textbf{Accuracy DIS}}                                                                                 & \multicolumn{2}{c|}{\textbf{Bias Score AMB}}                                                                                              & \multicolumn{2}{c|}{\textbf{Bias Score DIS}}                                                                               \\ \cline{3-10} 
\multicolumn{1}{|c|}{}                                                                                   & \multicolumn{1}{c|}{}                                           & \multicolumn{1}{c|}{\textbf{P}}                    & \multicolumn{1}{c|}{\textbf{\begin{tabular}[c]{@{}c@{}}Effect \\ Size\end{tabular}}} & \multicolumn{1}{c|}{\textbf{P}}     & \multicolumn{1}{c|}{\textbf{\begin{tabular}[c]{@{}c@{}}Effect \\ Size\end{tabular}}} & \multicolumn{1}{c|}{\textbf{P}}                    & \multicolumn{1}{c|}{\textbf{\begin{tabular}[c]{@{}c@{}}Effect \\ Size\end{tabular}}} & \multicolumn{1}{c|}{\textbf{P}}     & \multicolumn{1}{c|}{\textbf{\begin{tabular}[c]{@{}c@{}}Effect \\ Size\end{tabular}}} \\ \hlineB{3}
\multicolumn{2}{|c|}{\textbf{All results}}                                                                                                                                 & \multicolumn{1}{l|}{\textless 0.05}                & \multicolumn{1}{l|}{0.5417}                                                          & \multicolumn{1}{l|}{\textless 0.05} & \multicolumn{1}{l|}{0.3932}                                                          & \multicolumn{1}{l|}{-}                             & \multicolumn{1}{l|}{-}                                                               & \multicolumn{1}{l|}{\textless 0.05} & 0.2591                                                                               \\ \hlineB{3}
\multicolumn{1}{|c|}{\multirow{3}{*}{\textbf{Dataset}}}                                                  & UC - UF                                                         & \multicolumn{1}{l|}{-}                             & \multicolumn{1}{l|}{-}                                                               & \multicolumn{1}{l|}{-}              & \multicolumn{1}{l|}{-}                                                               & \multicolumn{1}{l|}{-}                             & \multicolumn{1}{l|}{-}                                                               & \multicolumn{1}{l|}{-}              & -                                                                                    \\ \clineB{2-10}{2} 
\multicolumn{1}{|c|}{}                                                                                   & UC - base                                                       & \multicolumn{1}{l|}{-}                             & \multicolumn{1}{l|}{-}                                                               & \multicolumn{1}{l|}{\textless 0.05} & \multicolumn{1}{l|}{0.6809}                                                          & \multicolumn{1}{l|}{-}                             & \multicolumn{1}{l|}{-}                                                               & \multicolumn{1}{l|}{\textless 0.05} & 0.5434                                                                               \\ \cline{2-10} 
\multicolumn{1}{|c|}{}                                                                                   & UF - base                                                       & \multicolumn{1}{l|}{\textless 0.05}                & \multicolumn{1}{l|}{0.558}                                                           & \multicolumn{1}{l|}{\textless 0.05} & \multicolumn{1}{l|}{0.3349}                                                          & \multicolumn{1}{l|}{-}                             & \multicolumn{1}{l|}{-}                                                               & \multicolumn{1}{l|}{-}              & -                                                                                    \\ \hlineB{3}
\multicolumn{1}{|c|}{\multirow{3}{*}{\textbf{Paradigm}}}                                                 & SFT - DPO                                                       & \multicolumn{1}{l|}{\textless 0.05}                & \multicolumn{1}{l|}{\begin{tabular}[c]{@{}l@{}}0.808 \\ {\ul \textbf{(DPO)}}\end{tabular}}          & \multicolumn{1}{l|}{\textless 0.05} & \multicolumn{1}{l|}{\begin{tabular}[c]{@{}l@{}}0.4729\\ {\ul\textbf{(DPO)}}\end{tabular}}          & \multicolumn{1}{l|}{-}                             & \multicolumn{1}{l|}{-}                                                               & \multicolumn{1}{l|}{\textless 0.05} & \begin{tabular}[c]{@{}l@{}}0.5726\\ {\ul\textbf{(DPO)}}\end{tabular}                               \\ \clineB{2-10}{2} 
\multicolumn{1}{|c|}{}                                                                                   & SFT - base                                                      & \multicolumn{1}{l|}{\textless 0.05}                & \multicolumn{1}{l|}{0.6206}                                                          & \multicolumn{1}{l|}{\textless 0.05} & \multicolumn{1}{l|}{0.5893}                                                          & \multicolumn{1}{l|}{-}                             & \multicolumn{1}{l|}{-}                                                               & \multicolumn{1}{l|}{\textless 0.05} & 0.4343                                                                               \\ \cline{2-10} 
\multicolumn{1}{|c|}{}                                                                                   & DPO - base                                                      & \multicolumn{1}{l|}{-}                             & \multicolumn{1}{l|}{-}                                                               & \multicolumn{1}{l|}{-}              & \multicolumn{1}{l|}{-}                                                               & \multicolumn{1}{l|}{-}                             & \multicolumn{1}{l|}{-}                                                               & \multicolumn{1}{l|}{-}              & -                                                                                    \\ \hlineB{3}
\multicolumn{1}{|c|}{\multirow{3}{*}{\textbf{\begin{tabular}[c]{@{}c@{}}Learning \\ Rate\end{tabular}}}} & 1e-3 - 2e-5                                                     & \multicolumn{1}{l|}{\textless 0.05}                & \multicolumn{1}{l|}{\begin{tabular}[c]{@{}l@{}}0.5892 \\ {\ul\textbf{(2e-5)}}\end{tabular}}        & \multicolumn{1}{l|}{-}              & \multicolumn{1}{l|}{-}                                                               & \multicolumn{1}{l|}{-}                             & \multicolumn{1}{l|}{-}                                                               & \multicolumn{1}{l|}{\textless 0.05} & \begin{tabular}[c]{@{}l@{}}0.3855\\ {\ul\textbf{(2e-5)}}\end{tabular}                              \\ \clineB{2-10}{2} 
\multicolumn{1}{|c|}{}                                                                                   & 1e-3 - base                                                     & \multicolumn{1}{l|}{\textless 0.05}                & \multicolumn{1}{l|}{0.762}                                                           & \multicolumn{1}{l|}{-}              & \multicolumn{1}{l|}{-}                                                               & \multicolumn{1}{l|}{-}                             & \multicolumn{1}{l|}{-}                                                               & \multicolumn{1}{l|}{\textless 0.05} & 0.509                                                                                \\ \cline{2-10} 
\multicolumn{1}{|c|}{}                                                                                   & 2e-5 - base                                                     & \multicolumn{1}{l|}{\textless 0.05}                & \multicolumn{1}{l|}{0.4237}                                                          & \multicolumn{1}{l|}{\textless 0.05} & \multicolumn{1}{l|}{0.4418}                                                          & \multicolumn{1}{l|}{-}                             & \multicolumn{1}{l|}{-}                                                               & \multicolumn{1}{l|}{-}              & -                                                                                    \\ \hlineB{3}
\multicolumn{1}{|c|}{\multirow{3}{*}{\textbf{Epochs}}}                                                   & 1e - 5e                                                         & \multicolumn{1}{l|}{-}                             & \multicolumn{1}{l|}{-}                                                               & \multicolumn{1}{l|}{-}              & \multicolumn{1}{l|}{-}                                                               & \multicolumn{1}{l|}{-}                             & \multicolumn{1}{l|}{-}                                                               & \multicolumn{1}{l|}{-}              & -                                                                                    \\ \clineB{2-10}{2} 
\multicolumn{1}{|c|}{}                                                                                   & 1e - base                                                       & \multicolumn{1}{l|}{\textless 0.05}                & \multicolumn{1}{l|}{0.4441}                                                          & \multicolumn{1}{l|}{\textless 0.05} & \multicolumn{1}{l|}{0.6305}                                                          & \multicolumn{1}{l|}{-}                             & \multicolumn{1}{l|}{-}                                                               & \multicolumn{1}{l|}{-}              & -                                                                                    \\ \cline{2-10} 
\multicolumn{1}{|c|}{}                                                                                   & 5e - base                                                       & \multicolumn{1}{l|}{\textless 0.05}                & \multicolumn{1}{l|}{0.5841}                                                          & \multicolumn{1}{l|}{\textless 0.05} & \multicolumn{1}{l|}{0.2713}                                                          & \multicolumn{1}{l|}{-}                             & \multicolumn{1}{l|}{-}                                                               & \multicolumn{1}{l|}{-}              & -                                                                                    \\ \hlineB{3}
\multicolumn{1}{|c|}{\multirow{11}{*}{\textbf{\begin{tabular}[c]{@{}c@{}}PEFT \\ Method\end{tabular}}}}  & All methods                                                     & \multicolumn{1}{l|}{\textless 0.05}                & \multicolumn{1}{l|}{0.159}                                                           & \multicolumn{1}{l|}{\textless 0.05} & \multicolumn{1}{l|}{0.193}                                                           & \multicolumn{1}{l|}{\textless 0.05}                & \multicolumn{1}{l|}{0.144}                                                           & \multicolumn{1}{l|}{-}              & -                                                                                    \\ \clineB{2-10}{3} 
\multicolumn{1}{|c|}{}                                                                                   & Lora - IA$^3$                                                      & \multicolumn{1}{l|}{\textless 0.05}                & \multicolumn{1}{l|}{\begin{tabular}[c]{@{}l@{}}0.7159\\ {\ul\textbf{(IA$^3$)}}\end{tabular}}          & \multicolumn{1}{l|}{-}              & \multicolumn{1}{l|}{-}                                                               & \multicolumn{1}{l|}{-}                             & \multicolumn{1}{l|}{-}                                                               & \multicolumn{1}{l|}{-}              & -                                                                                    \\ \cline{2-10} 
\multicolumn{1}{|c|}{}                                                                                   & \begin{tabular}[c]{@{}l@{}}IA$^3$ - \\ Prompt-Tuning\end{tabular}  & \multicolumn{1}{l|}{\textless 0.05}                & \multicolumn{1}{l|}{\begin{tabular}[c]{@{}l@{}}0.6582 \\ {\ul\textbf{(IA$^3$)}}\end{tabular}}         & \multicolumn{1}{l|}{\textless 0.05} & \multicolumn{1}{l|}{\begin{tabular}[c]{@{}l@{}}0.81\\ {\ul\textbf{(IA$^3$)}}\end{tabular}}            & \multicolumn{1}{l|}{-}                             & \multicolumn{1}{l|}{-}                                                               & \multicolumn{1}{l|}{-}              & -                                                                                    \\ \cline{2-10} 
\multicolumn{1}{|c|}{}                                                                                   & \begin{tabular}[c]{@{}l@{}}IA$^3$ - \\ P-Tuning\end{tabular}       & \multicolumn{1}{l|}{\textless 0.05}                & \multicolumn{1}{l|}{\begin{tabular}[c]{@{}l@{}}0.8979 \\ {\ul\textbf{(IA$^3$)}}\end{tabular}}         & \multicolumn{1}{l|}{-}              & \multicolumn{1}{l|}{-}                                                               & \multicolumn{1}{l|}{-}                             & \multicolumn{1}{l|}{-}                                                               & \multicolumn{1}{l|}{-}              & -                                                                                    \\ \cline{2-10} 
\multicolumn{1}{|c|}{}                                                                                   & \begin{tabular}[c]{@{}l@{}}Lora - \\ Prompt-Tuning\end{tabular} & \multicolumn{1}{l|}{-}                             & \multicolumn{1}{l|}{-}                                                               & \multicolumn{1}{l|}{-}              & \multicolumn{1}{l|}{-}                                                               & \multicolumn{1}{l|}{-}                             & \multicolumn{1}{l|}{-}                                                               & \multicolumn{1}{l|}{-}              & -                                                                                    \\ \cline{2-10} 
\multicolumn{1}{|c|}{}                                                                                   & \begin{tabular}[c]{@{}l@{}}Lora - \\ P-Tuning\end{tabular}      & \multicolumn{1}{l|}{-}                             & \multicolumn{1}{l|}{-}                                                               & \multicolumn{1}{l|}{-}              & \multicolumn{1}{l|}{-}                                                               & \multicolumn{1}{l|}{-}                             & \multicolumn{1}{l|}{-}                                                               & \multicolumn{1}{l|}{-}              & -                                                                                    \\ \cline{2-10} 
\multicolumn{1}{|c|}{}                                                                                   & \begin{tabular}[c]{@{}l@{}}Prompt - \\ P-Tuning\end{tabular}    & \multicolumn{1}{l|}{-}                             & \multicolumn{1}{l|}{-}                                                               & \multicolumn{1}{l|}{-}              & \multicolumn{1}{l|}{-}                                                               & \multicolumn{1}{l|}{-}                             & \multicolumn{1}{l|}{-}                                                               & \multicolumn{1}{l|}{-}              & -                                                                                    \\ \clineB{2-10}{2} 
\multicolumn{1}{|c|}{}                                                                                   & Lora - base                                                     & \multicolumn{1}{l|}{{\ul \textbf{\textless 0.05}}} & \multicolumn{1}{l|}{{\ul \textbf{0.6194}}}                                           & \multicolumn{1}{l|}{-}              & \multicolumn{1}{l|}{-}                                                               & \multicolumn{1}{l|}{-}                             & \multicolumn{1}{l|}{-}                                                               & \multicolumn{1}{l|}{-}              & -                                                                                    \\ \cline{2-10} 
\multicolumn{1}{|c|}{}                                                                                   & IA$^3$ - base                                                      & \multicolumn{1}{l|}{-}              & \multicolumn{1}{l|}{-}                                                & \multicolumn{1}{l|}{-}              & \multicolumn{1}{l|}{-}                                                               & \multicolumn{1}{l|}{-}                             & \multicolumn{1}{l|}{-}                                                               & \multicolumn{1}{l|}{-}              & -                                                                                    \\ \cline{2-10} 
\multicolumn{1}{|c|}{}                                                                                   & \begin{tabular}[c]{@{}l@{}}Prompt-Tuning \\ - base\end{tabular} & \multicolumn{1}{l|}{\textless 0.05}                & \multicolumn{1}{l|}{0.6282}                                                          & \multicolumn{1}{l|}{\textless 0.05} & \multicolumn{1}{l|}{0.8163}                                                          & \multicolumn{1}{l|}{-}                             & \multicolumn{1}{l|}{-}                                                               & \multicolumn{1}{l|}{-}              & -                                                                                    \\ \cline{2-10} 
\multicolumn{1}{|c|}{}                                                                                   & \begin{tabular}[c]{@{}l@{}}P-Tuning\\  - base\end{tabular}      & \multicolumn{1}{l|}{\textless 0.05}                & \multicolumn{1}{l|}{0.8582}                                                          & \multicolumn{1}{l|}{\textless 0.05} & \multicolumn{1}{l|}{0.5786}                                                          & \multicolumn{1}{l|}{-}                             & \multicolumn{1}{l|}{-}                                                               & \multicolumn{1}{l|}{-}              & -                                                                                    \\ \hlineB{3}
\multicolumn{1}{|c|}{\multirow{11}{*}{\textbf{\begin{tabular}[c]{@{}c@{}}Base \\ Model\end{tabular}}}}                                                     & All models                                                      & \multicolumn{1}{l|}{\textless 0.05}                & \multicolumn{1}{l|}{0.265}                                                           & \multicolumn{1}{l|}{-}              & \multicolumn{1}{l|}{-}                                                               & \multicolumn{1}{l|}{\textless 0.05}                & \multicolumn{1}{l|}{0.142}                                                           & \multicolumn{1}{l|}{\textless 0.05} & 0.225                                                                                \\ \clineB{2-10}{3} 
\multicolumn{1}{|c|}{}                                                                                   & \begin{tabular}[c]{@{}l@{}}Llama - \\ Mistral\end{tabular}      & \multicolumn{1}{l|}{-}                             & \multicolumn{1}{l|}{-}                                                               & \multicolumn{1}{l|}{-}              & \multicolumn{1}{l|}{-}                                                               & \multicolumn{1}{l|}{-}                             & \multicolumn{1}{l|}{-}                                                               & \multicolumn{1}{l|}{-}              & -                                                                                    \\ \cline{2-10} 
\multicolumn{1}{|c|}{}                                                                                   & \begin{tabular}[c]{@{}l@{}}Llama - \\ Qwen\end{tabular}         & \multicolumn{1}{l|}{\textless 0.05}                & \multicolumn{1}{l|}{\begin{tabular}[c]{@{}l@{}}0.7808 \\ {\ul\textbf{(Qwen)}}\end{tabular}}        & \multicolumn{1}{l|}{-}              & \multicolumn{1}{l|}{-}                                                               & \multicolumn{1}{l|}{0.056}                         & \multicolumn{1}{l|}{\begin{tabular}[c]{@{}l@{}}0.7510\\ {\ul\textbf{(Llama)}}\end{tabular}}        & \multicolumn{1}{l|}{\textless 0.05} & \begin{tabular}[c]{@{}l@{}}0.8451\\ {\ul\textbf{(Qwen)}}\end{tabular}                              \\ \cline{2-10} 
\multicolumn{1}{|c|}{}                                                                                   & \begin{tabular}[c]{@{}l@{}}Llama - \\ Gemma\end{tabular}        & \multicolumn{1}{l|}{\textless 0.05}                & \multicolumn{1}{l|}{\begin{tabular}[c]{@{}l@{}}1.0066\\ {\ul\textbf{(Gemma)}}\end{tabular}}        & \multicolumn{1}{l|}{-}              & \multicolumn{1}{l|}{-}                                                               & \multicolumn{1}{l|}{-}                             & \multicolumn{1}{l|}{-}                                                               & \multicolumn{1}{l|}{\textless 0.05} & \begin{tabular}[c]{@{}l@{}}0.8125\\ {\ul\textbf{(Gemma)}}\end{tabular}                             \\ \cline{2-10} 
\multicolumn{1}{|c|}{}                                                                                   & \begin{tabular}[c]{@{}l@{}}Mistral - \\ Qwen\end{tabular}       & \multicolumn{1}{l|}{-}                             & \multicolumn{1}{l|}{-}                                                               & \multicolumn{1}{l|}{-}              & \multicolumn{1}{l|}{-}                                                               & \multicolumn{1}{l|}{-}                             & \multicolumn{1}{l|}{-}                                                               & \multicolumn{1}{l|}{-}              & -                                                                                    \\ \cline{2-10} 
\multicolumn{1}{|c|}{}                                                                                   & \begin{tabular}[c]{@{}l@{}}Mistral - \\ Gemma\end{tabular}      & \multicolumn{1}{l|}{-}                             & \multicolumn{1}{l|}{-}                                                               & \multicolumn{1}{l|}{-}              & \multicolumn{1}{l|}{-}                                                               & \multicolumn{1}{l|}{-}                             & \multicolumn{1}{l|}{-}                                                               & \multicolumn{1}{l|}{-}              & -                                                                                    \\ \cline{2-10} 
\multicolumn{1}{|c|}{}                                                                                   & Qwen - Gemma                                                    & \multicolumn{1}{l|}{-}                             & \multicolumn{1}{l|}{-}                                                               & \multicolumn{1}{l|}{-}              & \multicolumn{1}{l|}{-}                                                               & \multicolumn{1}{l|}{-}                             & \multicolumn{1}{l|}{-}                                                               & \multicolumn{1}{l|}{-}              & -                                                                                    \\ \clineB{2-10}{2} 
\multicolumn{1}{|c|}{}                                                                                   & \begin{tabular}[c]{@{}l@{}}Llama - \\ base\end{tabular}         & \multicolumn{1}{l|}{\textless 0.05}                & \multicolumn{1}{l|}{0.4865}                                                          & \multicolumn{1}{l|}{\textless 0.05} & \multicolumn{1}{l|}{0.4759}                                                          & \multicolumn{1}{l|}{-}                             & \multicolumn{1}{l|}{-}                                                               & \multicolumn{1}{l|}{\textless 0.05} & 0.5708                                                                               \\ \cline{2-10} 
\multicolumn{1}{|c|}{}                                                                                   & \begin{tabular}[c]{@{}l@{}}Mistral - \\ base\end{tabular}       & \multicolumn{1}{l|}{\textless 0.05}                & \multicolumn{1}{l|}{0.808}                                                           & \multicolumn{1}{l|}{\textless 0.05} & \multicolumn{1}{l|}{0.4591}                                                          & \multicolumn{1}{l|}{\textless 0.05}                & \multicolumn{1}{l|}{0.5008}                                                          & \multicolumn{1}{l|}{-}              & -                                                                                    \\ \cline{2-10} 
\multicolumn{1}{|c|}{}                                                                                   & \begin{tabular}[c]{@{}l@{}}Qwen - \\ base\end{tabular}          & \multicolumn{1}{l|}{-}                             & \multicolumn{1}{l|}{-}                                                               & \multicolumn{1}{l|}{-}              & \multicolumn{1}{l|}{-}                                                               & \multicolumn{1}{l|}{-}                             & \multicolumn{1}{l|}{-}                                                               & \multicolumn{1}{l|}{-}              & -                                                                                    \\ \cline{2-10} 
\multicolumn{1}{|c|}{}                                                                                   & \begin{tabular}[c]{@{}l@{}}Gemma - \\ base\end{tabular}         & \multicolumn{1}{l|}{\textless 0.05}                & \multicolumn{1}{l|}{0.543}                                                           & \multicolumn{1}{l|}{\textless 0.05} & \multicolumn{1}{l|}{0.5901}                                                          & \multicolumn{1}{l|}{{\ul \textbf{\textless 0.05}}} & \multicolumn{1}{l|}{{\ul \textbf{0.7504}}}                                           & \multicolumn{1}{l|}{-}              & -                                                                                    \\ \hline
\end{tabular}
\end{table*}

\begin{table*}[htbp]
\caption{\centering Results of the statistical tests for fairness categories 1 and 2}
\label{tab:fairness_categories_1}
\renewcommand{\arraystretch}{1.1}
\centering
\scriptsize

\begin{tabular}{|cl|llll|cccc|}
\hline
\multicolumn{1}{|c|}{\multirow{3}{*}{\textbf{Factor}}}                                                  & \multicolumn{1}{c|}{\multirow{3}{*}{\textbf{Comparison}}}          & \multicolumn{4}{c|}{\textbf{1. Age}}                                                                                                                                                                                                                                                                                                                      & \multicolumn{4}{c|}{\textbf{2. Disability Status}}                                                                                                                                                                                                                                                                                                    \\ \cline{3-10} 
\multicolumn{1}{|c|}{}                                                                                  & \multicolumn{1}{c|}{}                                              & \multicolumn{1}{c|}{\textbf{\begin{tabular}[c]{@{}c@{}}Acc. \\ AMB\end{tabular}}}    & \multicolumn{1}{c|}{\textbf{\begin{tabular}[c]{@{}c@{}}Acc. \\ DIS\end{tabular}}}    & \multicolumn{1}{c|}{\textbf{\begin{tabular}[c]{@{}c@{}}Bias\\  AMB\end{tabular}}}    & \multicolumn{1}{c|}{\textbf{\begin{tabular}[c]{@{}c@{}}Bias\\ DIS\end{tabular}}}     & \multicolumn{1}{c|}{\textbf{\begin{tabular}[c]{@{}c@{}}Acc.\\ AMB\end{tabular}}}     & \multicolumn{1}{c|}{\textbf{\begin{tabular}[c]{@{}c@{}}Acc.\\ DIS\end{tabular}}}     & \multicolumn{1}{c|}{\textbf{\begin{tabular}[c]{@{}c@{}}Bias\\ AMB\end{tabular}}}     & \textbf{\begin{tabular}[c]{@{}c@{}}Bias\\ DIS\end{tabular}}                      \\ \cline{3-10} 
\multicolumn{1}{|c|}{}                                                                                  & \multicolumn{1}{c|}{}                                              & \multicolumn{1}{c|}{\textbf{\begin{tabular}[c]{@{}c@{}}Effect \\ Size\end{tabular}}} & \multicolumn{1}{c|}{\textbf{\begin{tabular}[c]{@{}c@{}}Effect \\ Size\end{tabular}}} & \multicolumn{1}{c|}{\textbf{\begin{tabular}[c]{@{}c@{}}Effect \\ Size\end{tabular}}} & \multicolumn{1}{c|}{\textbf{\begin{tabular}[c]{@{}c@{}}Effect \\ Size\end{tabular}}} & \multicolumn{1}{c|}{\textbf{\begin{tabular}[c]{@{}c@{}}Effect \\ Size\end{tabular}}} & \multicolumn{1}{c|}{\textbf{\begin{tabular}[c]{@{}c@{}}Effect \\ Size\end{tabular}}} & \multicolumn{1}{c|}{\textbf{\begin{tabular}[c]{@{}c@{}}Effect \\ Size\end{tabular}}} & \textbf{\begin{tabular}[c]{@{}c@{}}Effect \\ Size\end{tabular}}                  \\ \hline
\multicolumn{2}{|c|}{\textbf{All results}}                                                                                                                                   & \multicolumn{1}{l|}{0.5268}                                                          & \multicolumn{1}{l|}{0.5639}                                                          & \multicolumn{1}{l|}{\textbf{0.3563}}                                                 & -                                                                                    & \multicolumn{1}{c|}{0.4938}                                                          & \multicolumn{1}{c|}{0.2857}                                                          & \multicolumn{1}{c|}{-}                                                               & 0.2243                                                                           \\ \hline
\multicolumn{1}{|c|}{\multirow{3}{*}{\textbf{Dataset}}}                                                 & UC - UF                                                            & \multicolumn{1}{l|}{-}                                                               & \multicolumn{1}{l|}{-}                                                               & \multicolumn{1}{l|}{-}                                                               & -                                                                                    & \multicolumn{1}{c|}{-}                                                               & \multicolumn{1}{c|}{-}                                                               & \multicolumn{1}{c|}{-}                                                               & -                                                                                \\ \cline{2-10} 
\multicolumn{1}{|c|}{}                                                                                  & UC - base                                                          & \multicolumn{1}{l|}{-}                                                               & \multicolumn{1}{l|}{0.7298}                                                          & \multicolumn{1}{l|}{-}                                                               & -                                                                                    & \multicolumn{1}{c|}{0.5621}                                                          & \multicolumn{1}{c|}{-}                                                               & \multicolumn{1}{c|}{-}                                                               & -                                                                                \\ \cline{2-10} 
\multicolumn{1}{|c|}{}                                                                                  & UF - base                                                          & \multicolumn{1}{l|}{0.5402}                                                          & \multicolumn{1}{l|}{0.5253}                                                          & \multicolumn{1}{l|}{\textbf{0.3223}}                                                 & -                                                                                    & \multicolumn{1}{c|}{0.4904}                                                          & \multicolumn{1}{c|}{0.2501}                                                          & \multicolumn{1}{c|}{-}                                                               & -                                                                                \\ \hline
\multicolumn{1}{|c|}{\multirow{3}{*}{\textbf{Paradigm}}}                                                & SFT - DPO                                                          & \multicolumn{1}{l|}{\begin{tabular}[c]{@{}l@{}}0.6795\\ (DPO)\end{tabular}}          & \multicolumn{1}{l|}{\begin{tabular}[c]{@{}l@{}}0.5220\\ (DPO)\end{tabular}}          & \multicolumn{1}{l|}{-}                                                               & -                                                                                    & \multicolumn{1}{c|}{\begin{tabular}[c]{@{}c@{}}0.7629 \\ (DPO)\end{tabular}}         & \multicolumn{1}{c|}{\begin{tabular}[c]{@{}c@{}}0.5434 \\ (DPO)\end{tabular}}         & \multicolumn{1}{c|}{-}                                                               & -                                                                                \\ \cline{2-10} 
\multicolumn{1}{|c|}{}                                                                                  & SFT - base                                                         & \multicolumn{1}{l|}{0.613}                                                           & \multicolumn{1}{l|}{0.7067}                                                          & \multicolumn{1}{l|}{\textbf{0.4875}}                                                 & -                                                                                    & \multicolumn{1}{c|}{0.5893}                                                          & \multicolumn{1}{c|}{0.4922}                                                          & \multicolumn{1}{c|}{-}                                                               & 0.3357                                                                           \\ \cline{2-10} 
\multicolumn{1}{|c|}{}                                                                                  & DPO - base                                                         & \multicolumn{1}{l|}{-}                                                               & \multicolumn{1}{l|}{-}                                                               & \multicolumn{1}{l|}{-}                                                               & -                                                                                    & \multicolumn{1}{c|}{-}                                                               & \multicolumn{1}{c|}{-}                                                               & \multicolumn{1}{c|}{0.5538}                                                          & -                                                                                \\ \hline
\multicolumn{1}{|c|}{\multirow{3}{*}{\textbf{\begin{tabular}[c]{@{}c@{}}Learning\\ Rate\end{tabular}}}} & 1e-3 - 2e-5                                                        & \multicolumn{1}{l|}{\begin{tabular}[c]{@{}l@{}}0.5262\\ (2e-5)\end{tabular}}         & \multicolumn{1}{l|}{-}                                                               & \multicolumn{1}{l|}{-}                                                               & -                                                                                    & \multicolumn{1}{c|}{-}                                                               & \multicolumn{1}{c|}{-}                                                               & \multicolumn{1}{c|}{-}                                                               & -                                                                                \\ \cline{2-10} 
\multicolumn{1}{|c|}{}                                                                                  & 1e-3 - base                                                        & \multicolumn{1}{l|}{0.6988}                                                          & \multicolumn{1}{l|}{0.5618}                                                          & \multicolumn{1}{l|}{-}                                                               & -                                                                                    & \multicolumn{1}{c|}{0.5664}                                                          & \multicolumn{1}{l|}{}                                                                & \multicolumn{1}{c|}{-}                                                               & -                                                                                \\ \cline{2-10} 
\multicolumn{1}{|c|}{}                                                                                  & 2e-5 - base                                                        & \multicolumn{1}{l|}{0.4339}                                                          & \multicolumn{1}{l|}{0.5835}                                                          & \multicolumn{1}{l|}{\textbf{0.392}}                                                  & -                                                                                    & \multicolumn{1}{c|}{0.4641}                                                          & \multicolumn{1}{c|}{-}                                                               & \multicolumn{1}{c|}{-}                                                               & -                                                                                \\ \hline
\multicolumn{1}{|c|}{\multirow{3}{*}{\textbf{Epochs}}}                                                  & 1e - 5e                                                            & \multicolumn{1}{l|}{-}                                                               & \multicolumn{1}{l|}{-}                                                               & \multicolumn{1}{l|}{-}                                                               & -                                                                                    & \multicolumn{1}{c|}{-}                                                               & \multicolumn{1}{c|}{-}                                                               & \multicolumn{1}{c|}{-}                                                               & -                                                                                \\ \cline{2-10} 
\multicolumn{1}{|c|}{}                                                                                  & 1e - base                                                          & \multicolumn{1}{l|}{0.4558}                                                          & \multicolumn{1}{l|}{0.7907}                                                          & \multicolumn{1}{l|}{\textbf{0.4718}}                                                 & -                                                                                    & \multicolumn{1}{c|}{0.4624}                                                          & \multicolumn{1}{c|}{0.4508}                                                          & \multicolumn{1}{c|}{-}                                                               & -                                                                                \\ \cline{2-10} 
\multicolumn{1}{|c|}{}                                                                                  & 5e - base                                                          & \multicolumn{1}{l|}{0.5268}                                                          & \multicolumn{1}{l|}{0.4698}                                                          & \multicolumn{1}{l|}{\textbf{0.2701}}                                                 & -                                                                                    & \multicolumn{1}{c|}{0.5159}                                                          & \multicolumn{1}{c|}{-}                                                               & \multicolumn{1}{c|}{-}                                                               & -                                                                                \\ \hline
\multicolumn{1}{|c|}{\multirow{11}{*}{\textbf{\begin{tabular}[c]{@{}c@{}}PEFT\\ Method\end{tabular}}}}  & All methods                                                        & \multicolumn{1}{l|}{0.146}                                                           & \multicolumn{1}{l|}{0.291}                                                           & \multicolumn{1}{l|}{0.284}                                                           & -                                                                                    & \multicolumn{1}{c|}{-}                                                               & \multicolumn{1}{c|}{0.137}                                                           & \multicolumn{1}{c|}{-}                                                               & -                                                                                \\ \cline{2-10} 
\multicolumn{1}{|c|}{}                                                                                  & Lora - IA$^3$                                                         & \multicolumn{1}{l|}{\begin{tabular}[c]{@{}l@{}}0.7620\\ (IA$^3$)\end{tabular}}          & \multicolumn{1}{l|}{-}                                                               & \multicolumn{1}{l|}{-}                                                               & -                                                                                    & \multicolumn{1}{c|}{-}                                                               & \multicolumn{1}{c|}{-}                                                               & \multicolumn{1}{c|}{-}                                                               & -                                                                                \\ \cline{2-10} 
\multicolumn{1}{|c|}{}                                                                                  & \begin{tabular}[c]{@{}l@{}}IA$^3$ - \\ Prompt \\ Tuning\end{tabular}  & \multicolumn{1}{l|}{-}                                                               & \multicolumn{1}{l|}{\begin{tabular}[c]{@{}l@{}}0.9397\\ (IA$^3$)\end{tabular}}          & \multicolumn{1}{l|}{\begin{tabular}[c]{@{}l@{}}1.0113\\ (Prompt)\end{tabular}}       & -                                                                                    & \multicolumn{1}{c|}{-}                                                               & \multicolumn{1}{c|}{\begin{tabular}[c]{@{}c@{}}0.7063 \\ (IA$^3$)\end{tabular}}         & \multicolumn{1}{c|}{-}                                                               & -                                                                                \\ \cline{2-10} 
\multicolumn{1}{|c|}{}                                                                                  & \begin{tabular}[c]{@{}l@{}}IA$^3$ - \\ P-Tuning\end{tabular}          & \multicolumn{1}{l|}{\begin{tabular}[c]{@{}l@{}}0.7936\\ (IA$^3$)\end{tabular}}          & \multicolumn{1}{l|}{\begin{tabular}[c]{@{}l@{}}0.8264\\ (IA$^3$)\end{tabular}}          & \multicolumn{1}{l|}{-}                                                               & -                                                                                    & \multicolumn{1}{c|}{-}                                                               & \multicolumn{1}{c|}{-}                                                               & \multicolumn{1}{c|}{-}                                                               & -                                                                                \\ \cline{2-10} 
\multicolumn{1}{|c|}{}                                                                                  & \begin{tabular}[c]{@{}l@{}}Lora - \\ Prompt \\ Tuning\end{tabular} & \multicolumn{1}{l|}{-}                                                               & \multicolumn{1}{l|}{-}                                                               & \multicolumn{1}{l|}{\begin{tabular}[c]{@{}l@{}}0.7777\\ (Prompt)\end{tabular}}       & -                                                                                    & \multicolumn{1}{c|}{-}                                                               & \multicolumn{1}{c|}{-}                                                               & \multicolumn{1}{c|}{-}                                                               & -                                                                                \\ \cline{2-10} 
\multicolumn{1}{|c|}{}                                                                                  & \begin{tabular}[c]{@{}l@{}}Lora - \\ P-Tuning\end{tabular}         & \multicolumn{1}{l|}{-}                                                               & \multicolumn{1}{l|}{\begin{tabular}[c]{@{}l@{}}0.6442\\ (Lora)\end{tabular}}         & \multicolumn{1}{l|}{-}                                                               & -                                                                                    & \multicolumn{1}{c|}{-}                                                               & \multicolumn{1}{c|}{-}                                                               & \multicolumn{1}{c|}{-}                                                               & -                                                                                \\ \cline{2-10} 
\multicolumn{1}{|c|}{}                                                                                  & \begin{tabular}[c]{@{}l@{}}Prompt - \\ P-Tuning\end{tabular}       & \multicolumn{1}{l|}{-}                                                               & \multicolumn{1}{l|}{-}                                                               & \multicolumn{1}{l|}{-}                                                               & -                                                                                    & \multicolumn{1}{c|}{-}                                                               & \multicolumn{1}{c|}{-}                                                               & \multicolumn{1}{c|}{-}                                                               & -                                                                                \\ \cline{2-10} 
\multicolumn{1}{|c|}{}                                                                                  & Lora - base                                                        & \multicolumn{1}{l|}{0.636}                                                           & \multicolumn{1}{l|}{0.4188}                                                          & \multicolumn{1}{l|}{-}                                                               & -                                                                                    & \multicolumn{1}{c|}{0.4811}                                                          & \multicolumn{1}{c|}{-}                                                               & \multicolumn{1}{c|}{-}                                                               & -                                                                                \\ \cline{2-10} 
\multicolumn{1}{|c|}{}                                                                                  & IA$^3$ - base                                                         & \multicolumn{1}{l|}{-}                                                               & \multicolumn{1}{l|}{-}                                                               & \multicolumn{1}{l|}{-}                                                               & 0.4741                                                                               & \multicolumn{1}{c|}{-}                                                               & \multicolumn{1}{c|}{-}                                                               & \multicolumn{1}{c|}{-}                                                               & -                                                                                \\ \cline{2-10} 
\multicolumn{1}{|c|}{}                                                                                  & \begin{tabular}[c]{@{}l@{}}Prompt \\ Tuning\\  - base\end{tabular} & \multicolumn{1}{l|}{0.5565}                                                          & \multicolumn{1}{l|}{1.0548}                                                          & \multicolumn{1}{l|}{\textbf{1.0194}}                                                 & -                                                                                    & \multicolumn{1}{c|}{0.5441}                                                          & \multicolumn{1}{c|}{0.6416}                                                          & \multicolumn{1}{c|}{-}                                                               & -                                                                                \\ \cline{2-10} 
\multicolumn{1}{|c|}{}                                                                                  & \begin{tabular}[c]{@{}l@{}}P-Tuning\\  - base\end{tabular}         & \multicolumn{1}{l|}{0.8347}                                                          & \multicolumn{1}{l|}{0.7263}                                                          & \multicolumn{1}{l|}{-}                                                               & -                                                                                    & \multicolumn{1}{c|}{0.6177}                                                          & \multicolumn{1}{c|}{0.5183}                                                          & \multicolumn{1}{c|}{-}                                                               & -                                                                                \\ \hline
\multicolumn{1}{|c|}{\multirow{11}{*}{\textbf{Model}}}                                                  & All models                                                         & \multicolumn{1}{l|}{0.184}                                                           & \multicolumn{1}{l|}{-}                                                               & \multicolumn{1}{l|}{0.142}                                                           & -                                                                                    & \multicolumn{1}{c|}{0.168}                                                           & \multicolumn{1}{l|}{-}                                                               & \multicolumn{1}{l|}{-}                                                               & \multicolumn{1}{l|}{0.188}                                                       \\ \cline{2-10} 
\multicolumn{1}{|c|}{}                                                                                  & \begin{tabular}[c]{@{}l@{}}Llama - \\ Mistral\end{tabular}         & \multicolumn{1}{l|}{\begin{tabular}[c]{@{}l@{}}0.8451\\ (Mistral)\end{tabular}}      & \multicolumn{1}{l|}{-}                                                               & \multicolumn{1}{l|}{-}                                                               & -                                                                                    & \multicolumn{1}{l|}{-}                                                               & \multicolumn{1}{l|}{-}                                                               & \multicolumn{1}{l|}{-}                                                               & \multicolumn{1}{l|}{\begin{tabular}[c]{@{}l@{}}0.7808 \\ (Mistral)\end{tabular}} \\ \cline{2-10} 
\multicolumn{1}{|c|}{}                                                                                  & \begin{tabular}[c]{@{}l@{}}Llama - \\ Qwen\end{tabular}            & \multicolumn{1}{l|}{-}                                                               & \multicolumn{1}{l|}{-}                                                               & \multicolumn{1}{l|}{-}                                                               & -                                                                                    & \multicolumn{1}{l|}{-}                                                               & \multicolumn{1}{l|}{-}                                                               & \multicolumn{1}{l|}{-}                                                               & \multicolumn{1}{l|}{-}                                                           \\ \cline{2-10} 
\multicolumn{1}{|c|}{}                                                                                  & \begin{tabular}[c]{@{}l@{}}Llama - \\ Gemma\end{tabular}           & \multicolumn{1}{l|}{\begin{tabular}[c]{@{}l@{}}0.8748\\ (Gemma)\end{tabular}}        & \multicolumn{1}{l|}{-}                                                               & \multicolumn{1}{l|}{-}                                                               & -                                                                                    & \multicolumn{1}{l|}{\begin{tabular}[c]{@{}l@{}}1.0066 \\ (Gemma)\end{tabular}}       & \multicolumn{1}{l|}{-}                                                               & \multicolumn{1}{l|}{-}                                                               & \multicolumn{1}{l|}{\begin{tabular}[c]{@{}l@{}}0.8748 \\ (Gemma)\end{tabular}}   \\ \cline{2-10} 
\multicolumn{1}{|c|}{}                                                                                  & \begin{tabular}[c]{@{}l@{}}Mistral -\\  Qwen\end{tabular}          & \multicolumn{1}{l|}{-}                                                               & \multicolumn{1}{l|}{-}                                                               & \multicolumn{1}{l|}{\begin{tabular}[c]{@{}l@{}}0.8451\\ (Mistral)\end{tabular}}      & -                                                                                    & \multicolumn{1}{l|}{-}                                                               & \multicolumn{1}{l|}{-}                                                               & \multicolumn{1}{l|}{-}                                                               & \multicolumn{1}{l|}{-}                                                           \\ \cline{2-10} 
\multicolumn{1}{|c|}{}                                                                                  & \begin{tabular}[c]{@{}l@{}}Mistral - \\ Gemma\end{tabular}         & \multicolumn{1}{l|}{-}                                                               & \multicolumn{1}{l|}{-}                                                               & \multicolumn{1}{l|}{-}                                                               & -                                                                                    & \multicolumn{1}{l|}{-}                                                               & \multicolumn{1}{l|}{-}                                                               & \multicolumn{1}{l|}{-}                                                               & \multicolumn{1}{l|}{-}                                                           \\ \cline{2-10} 
\multicolumn{1}{|c|}{}                                                                                  & \begin{tabular}[c]{@{}l@{}}Qwen -\\  Gemma\end{tabular}            & \multicolumn{1}{l|}{-}                                                               & \multicolumn{1}{l|}{-}                                                               & \multicolumn{1}{l|}{-}                                                               & -                                                                                    & \multicolumn{1}{l|}{-}                                                               & \multicolumn{1}{l|}{-}                                                               & \multicolumn{1}{l|}{-}                                                               & \multicolumn{1}{l|}{-}                                                           \\ \cline{2-10} 
\multicolumn{1}{|c|}{}                                                                                  & \begin{tabular}[c]{@{}l@{}}Llama - \\ base\end{tabular}            & \multicolumn{1}{l|}{0.5049}                                                          & \multicolumn{1}{l|}{-}                                                               & \multicolumn{1}{l|}{-}                                                               & -                                                                                    & \multicolumn{1}{l|}{-}                                                               & \multicolumn{1}{l|}{-}                                                               & \multicolumn{1}{l|}{-}                                                               & \multicolumn{1}{l|}{0.4801}                                                      \\ \cline{2-10} 
\multicolumn{1}{|c|}{}                                                                                  & \begin{tabular}[c]{@{}l@{}}Mistral - \\ base\end{tabular}          & \multicolumn{1}{l|}{0.6786}                                                          & \multicolumn{1}{l|}{0.6659}                                                          & \multicolumn{1}{l|}{\textbf{0.8753}}                                                 & -                                                                                    & \multicolumn{1}{l|}{0.7421}                                                          & \multicolumn{1}{l|}{-}                                                               & \multicolumn{1}{l|}{-}                                                               & \multicolumn{1}{l|}{0.5149}                                                      \\ \cline{2-10} 
\multicolumn{1}{|c|}{}                                                                                  & \begin{tabular}[c]{@{}l@{}}Qwen - \\ base\end{tabular}             & \multicolumn{1}{l|}{-}                                                               & \multicolumn{1}{l|}{0.4817}                                                          & \multicolumn{1}{l|}{-}                                                               & 0.4757                                                                               & \multicolumn{1}{l|}{-}                                                               & \multicolumn{1}{l|}{-}                                                               & \multicolumn{1}{l|}{-}                                                               & \multicolumn{1}{l|}{-}                                                           \\ \cline{2-10} 
\multicolumn{1}{|c|}{}                                                                                  & \begin{tabular}[c]{@{}l@{}}Gemma - \\ base\end{tabular}            & \multicolumn{1}{l|}{0.8415}                                                          & \multicolumn{1}{l|}{0.6982}                                                          & \multicolumn{1}{l|}{\textbf{0.8033}}                                                 & -                                                                                    & \multicolumn{1}{l|}{-}                                                               & \multicolumn{1}{l|}{0.517}                                                           & \multicolumn{1}{l|}{\textbf{0.5314}}                                                 & \multicolumn{1}{l|}{-}                                                           \\ \hline
\end{tabular}
\end{table*}

\begin{table*}[htbp]
\caption{\centering Results of the statistical tests for fairness categories 3 and 4}
\label{tab:fairness_categories_3}
\renewcommand{\arraystretch}{1.1}
\centering
\scriptsize

\begin{tabular}{|cl|cccc|cccc|}
\hline
\multicolumn{1}{|c|}{\multirow{3}{*}{\textbf{Factor}}}                                                  & \multicolumn{1}{c|}{\multirow{3}{*}{\textbf{Comparison}}}          & \multicolumn{4}{c|}{\textbf{3. Gender Identity}}                                                                                                                                                                                                                                                                                     & \multicolumn{4}{c|}{\textbf{4. Nationality}}                                                                                                                                                                                                                                                                                         \\ \cline{3-10} 
\multicolumn{1}{|c|}{}                                                                                  & \multicolumn{1}{c|}{}                                              & \multicolumn{1}{c|}{\textbf{\begin{tabular}[c]{@{}c@{}}Acc. \\ AMB\end{tabular}}}    & \multicolumn{1}{c|}{\textbf{\begin{tabular}[c]{@{}c@{}}Acc. \\ DIS\end{tabular}}}    & \multicolumn{1}{c|}{\textbf{\begin{tabular}[c]{@{}c@{}}Bias\\  AMB\end{tabular}}}    & \textbf{\begin{tabular}[c]{@{}c@{}}Bias\\ DIS\end{tabular}}     & \multicolumn{1}{c|}{\textbf{\begin{tabular}[c]{@{}c@{}}Acc.\\ AMB\end{tabular}}}     & \multicolumn{1}{c|}{\textbf{\begin{tabular}[c]{@{}c@{}}Acc.\\ DIS\end{tabular}}}     & \multicolumn{1}{c|}{\textbf{\begin{tabular}[c]{@{}c@{}}Bias\\ AMB\end{tabular}}}     & \textbf{\begin{tabular}[c]{@{}c@{}}Bias\\ DIS\end{tabular}}     \\ \cline{3-10} 
\multicolumn{1}{|c|}{}                                                                                  & \multicolumn{1}{c|}{}                                              & \multicolumn{1}{c|}{\textbf{\begin{tabular}[c]{@{}c@{}}Effect \\ Size\end{tabular}}} & \multicolumn{1}{c|}{\textbf{\begin{tabular}[c]{@{}c@{}}Effect \\ Size\end{tabular}}} & \multicolumn{1}{c|}{\textbf{\begin{tabular}[c]{@{}c@{}}Effect \\ Size\end{tabular}}} & \textbf{\begin{tabular}[c]{@{}c@{}}Effect \\ Size\end{tabular}} & \multicolumn{1}{c|}{\textbf{\begin{tabular}[c]{@{}c@{}}Effect \\ Size\end{tabular}}} & \multicolumn{1}{c|}{\textbf{\begin{tabular}[c]{@{}c@{}}Effect \\ Size\end{tabular}}} & \multicolumn{1}{c|}{\textbf{\begin{tabular}[c]{@{}c@{}}Effect \\ Size\end{tabular}}} & \textbf{\begin{tabular}[c]{@{}c@{}}Effect \\ Size\end{tabular}} \\ \hline
\multicolumn{2}{|c|}{\textbf{All results}}                                                                                                                                   & \multicolumn{1}{c|}{0.6221}                                                          & \multicolumn{1}{c|}{-}                                                               & \multicolumn{1}{c|}{\textbf{-}}                                                      & 0.2198                                                          & \multicolumn{1}{c|}{0.5961}                                                          & \multicolumn{1}{c|}{0.355}                                                           & \multicolumn{1}{c|}{-}                                                               & -                                                               \\ \hline
\multicolumn{1}{|c|}{\multirow{3}{*}{\textbf{Dataset}}}                                                 & UC - UF                                                            & \multicolumn{1}{c|}{-}                                                               & \multicolumn{1}{c|}{\begin{tabular}[c]{@{}c@{}}0.6010 \\ (UF)\end{tabular}}          & \multicolumn{1}{c|}{-}                                                               & -                                                               & \multicolumn{1}{c|}{-}                                                               & \multicolumn{1}{c|}{-}                                                               & \multicolumn{1}{c|}{-}                                                               & -                                                               \\ \cline{2-10} 
\multicolumn{1}{|c|}{}                                                                                  & UC - base                                                          & \multicolumn{1}{c|}{0.5817}                                                          & \multicolumn{1}{c|}{0.5434}                                                          & \multicolumn{1}{c|}{-}                                                               & 0.527                                                           & \multicolumn{1}{c|}{0.5453}                                                          & \multicolumn{1}{c|}{0.7804}                                                          & \multicolumn{1}{c|}{-}                                                               & -                                                               \\ \cline{2-10} 
\multicolumn{1}{|c|}{}                                                                                  & UF - base                                                          & \multicolumn{1}{c|}{0.6295}                                                          & \multicolumn{1}{c|}{-}                                                               & \multicolumn{1}{c|}{\textbf{-}}                                                      & -                                                               & \multicolumn{1}{c|}{0.6027}                                                          & \multicolumn{1}{c|}{0.2732}                                                          & \multicolumn{1}{c|}{-}                                                               & -                                                               \\ \hline
\multicolumn{1}{|c|}{\multirow{3}{*}{\textbf{Paradigm}}}                                                & SFT - DPO                                                          & \multicolumn{1}{c|}{\begin{tabular}[c]{@{}c@{}}0.7441 \\ (DPO)\end{tabular}}         & \multicolumn{1}{c|}{\begin{tabular}[c]{@{}c@{}}0.4386 \\ (DPO)\end{tabular}}         & \multicolumn{1}{c|}{-}                                                               & 0.8561                                                          & \multicolumn{1}{c|}{\begin{tabular}[c]{@{}c@{}}0.8410\\ (DPO)\end{tabular}}          & \multicolumn{1}{c|}{-}                                                               & \multicolumn{1}{c|}{-}                                                               & \begin{tabular}[c]{@{}c@{}}0.8561 \\ (DPO)\end{tabular}         \\ \cline{2-10} 
\multicolumn{1}{|c|}{}                                                                                  & SFT - base                                                         & \multicolumn{1}{c|}{0.7039}                                                          & \multicolumn{1}{c|}{0.3298}                                                          & \multicolumn{1}{c|}{\textbf{-}}                                                      & 0.4258                                                          & \multicolumn{1}{c|}{0.6916}                                                          & \multicolumn{1}{c|}{0.5058}                                                          & \multicolumn{1}{c|}{-}                                                               & 0.275                                                           \\ \cline{2-10} 
\multicolumn{1}{|c|}{}                                                                                  & DPO - base                                                         & \multicolumn{1}{c|}{-}                                                               & \multicolumn{1}{c|}{\textbf{0.5726}}                                                 & \multicolumn{1}{c|}{-}                                                               & \textbf{0.4115}                                                 & \multicolumn{1}{c|}{-}                                                               & \multicolumn{1}{c|}{-}                                                               & \multicolumn{1}{c|}{-}                                                               & \textbf{0.4523}                                                 \\ \hline
\multicolumn{1}{|c|}{\multirow{3}{*}{\textbf{\begin{tabular}[c]{@{}c@{}}Learning\\ Rate\end{tabular}}}} & 1e-3 - 2e-5                                                        & \multicolumn{1}{c|}{\begin{tabular}[c]{@{}c@{}}0.4819 \\ (2e-5)\end{tabular}}        & \multicolumn{1}{c|}{-}                                                               & \multicolumn{1}{c|}{-}                                                               & -                                                               & \multicolumn{1}{c|}{\begin{tabular}[c]{@{}c@{}}0.4929\\ (2e-5)\end{tabular}}         & \multicolumn{1}{c|}{-}                                                               & \multicolumn{1}{c|}{-}                                                               & \begin{tabular}[c]{@{}c@{}}0.4655 \\ (2e-5)\end{tabular}        \\ \cline{2-10} 
\multicolumn{1}{|c|}{}                                                                                  & 1e-3 - base                                                        & \multicolumn{1}{c|}{0.7986}                                                          & \multicolumn{1}{c|}{-}                                                               & \multicolumn{1}{c|}{-}                                                               & 0.3696                                                          & \multicolumn{1}{c|}{0.7213}                                                          & \multicolumn{1}{c|}{0.3923}                                                          & \multicolumn{1}{c|}{-}                                                               & -                                                               \\ \cline{2-10} 
\multicolumn{1}{|c|}{}                                                                                  & 2e-5 - base                                                        & \multicolumn{1}{c|}{0.523}                                                           & \multicolumn{1}{c|}{-}                                                               & \multicolumn{1}{c|}{\textbf{-}}                                                      & -                                                               & \multicolumn{1}{c|}{0.542}                                                           & \multicolumn{1}{c|}{0.3558}                                                          & \multicolumn{1}{c|}{-}                                                               & -                                                               \\ \hline
\multicolumn{1}{|c|}{\multirow{3}{*}{\textbf{Epochs}}}                                                  & 1e - 5e                                                            & \multicolumn{1}{c|}{-}                                                               & \multicolumn{1}{c|}{-}                                                               & \multicolumn{1}{c|}{-}                                                               & -                                                               & \multicolumn{1}{c|}{-}                                                               & \multicolumn{1}{c|}{-}                                                               & \multicolumn{1}{c|}{-}                                                               & -                                                               \\ \cline{2-10} 
\multicolumn{1}{|c|}{}                                                                                  & 1e - base                                                          & \multicolumn{1}{c|}{0.5681}                                                          & \multicolumn{1}{c|}{-}                                                               & \multicolumn{1}{c|}{\textbf{-}}                                                      & -                                                               & \multicolumn{1}{c|}{0.6077}                                                          & \multicolumn{1}{c|}{0.6264}                                                          & \multicolumn{1}{c|}{-}                                                               & -                                                               \\ \cline{2-10} 
\multicolumn{1}{|c|}{}                                                                                  & 5e - base                                                          & \multicolumn{1}{c|}{0.6438}                                                          & \multicolumn{1}{c|}{-}                                                               & \multicolumn{1}{c|}{\textbf{-}}                                                      & -                                                               & \multicolumn{1}{c|}{0.5986}                                                          & \multicolumn{1}{c|}{-}                                                               & \multicolumn{1}{c|}{-}                                                               & -                                                               \\ \hline
\multicolumn{1}{|c|}{\multirow{11}{*}{\textbf{\begin{tabular}[c]{@{}c@{}}PEFT\\ Method\end{tabular}}}}  & All methods                                                        & \multicolumn{1}{c|}{0.216}                                                           & \multicolumn{1}{c|}{-}                                                               & \multicolumn{1}{c|}{-}                                                               & -                                                               & \multicolumn{1}{c|}{0.227}                                                           & \multicolumn{1}{c|}{-}                                                               & \multicolumn{1}{c|}{0.128}                                                           & -                                                               \\ \cline{2-10} 
\multicolumn{1}{|c|}{}                                                                                  & Lora - IA$^3$                                                         & \multicolumn{1}{c|}{-}                                                               & \multicolumn{1}{c|}{-}                                                               & \multicolumn{1}{c|}{-}                                                               & -                                                               & \multicolumn{1}{c|}{\begin{tabular}[c]{@{}c@{}}0.6722\\ (IA$^3$)\end{tabular}}          & \multicolumn{1}{c|}{-}                                                               & \multicolumn{1}{c|}{-}                                                               & -                                                               \\ \cline{2-10} 
\multicolumn{1}{|c|}{}                                                                                  & \begin{tabular}[c]{@{}l@{}}IA$^3$ - \\ Prompt \\ Tuning\end{tabular}  & \multicolumn{1}{c|}{\begin{tabular}[c]{@{}c@{}}0.7479 \\ (IA$^3$)\end{tabular}}         & \multicolumn{1}{c|}{-}                                                               & \multicolumn{1}{c|}{-}                                                               & -                                                               & \multicolumn{1}{c|}{\begin{tabular}[c]{@{}c@{}}0.7463\\ (IA$^3$)\end{tabular}}          & \multicolumn{1}{c|}{-}                                                               & \multicolumn{1}{c|}{-}                                                               & -                                                               \\ \cline{2-10} 
\multicolumn{1}{|c|}{}                                                                                  & \begin{tabular}[c]{@{}l@{}}IA$^3$ - \\ P-Tuning\end{tabular}          & \multicolumn{1}{c|}{\begin{tabular}[c]{@{}c@{}}0.8273\\ (IA$^3$)\end{tabular}}          & \multicolumn{1}{c|}{-}                                                               & \multicolumn{1}{c|}{-}                                                               & -                                                               & \multicolumn{1}{c|}{\begin{tabular}[c]{@{}c@{}}0.9594\\ (IA$^3$)\end{tabular}}          & \multicolumn{1}{c|}{-}                                                               & \multicolumn{1}{c|}{-}                                                               & -                                                               \\ \cline{2-10} 
\multicolumn{1}{|c|}{}                                                                                  & \begin{tabular}[c]{@{}l@{}}Lora - \\ Prompt \\ Tuning\end{tabular} & \multicolumn{1}{c|}{-}                                                               & \multicolumn{1}{c|}{-}                                                               & \multicolumn{1}{c|}{-}                                                               & -                                                               & \multicolumn{1}{c|}{-}                                                               & \multicolumn{1}{c|}{-}                                                               & \multicolumn{1}{c|}{-}                                                               & -                                                               \\ \cline{2-10} 
\multicolumn{1}{|c|}{}                                                                                  & \begin{tabular}[c]{@{}l@{}}Lora - \\ P-Tuning\end{tabular}         & \multicolumn{1}{c|}{-}                                                               & \multicolumn{1}{c|}{-}                                                               & \multicolumn{1}{c|}{-}                                                               & -                                                               & \multicolumn{1}{c|}{-}                                                               & \multicolumn{1}{c|}{-}                                                               & \multicolumn{1}{c|}{-}                                                               & -                                                               \\ \cline{2-10} 
\multicolumn{1}{|c|}{}                                                                                  & \begin{tabular}[c]{@{}l@{}}Prompt - \\ P-Tuning\end{tabular}       & \multicolumn{1}{c|}{-}                                                               & \multicolumn{1}{c|}{-}                                                               & \multicolumn{1}{c|}{-}                                                               & -                                                               & \multicolumn{1}{c|}{-}                                                               & \multicolumn{1}{c|}{-}                                                               & \multicolumn{1}{c|}{-}                                                               & -                                                               \\ \cline{2-10} 
\multicolumn{1}{|c|}{}                                                                                  & Lora - base                                                        & \multicolumn{1}{c|}{0.5919}                                                          & \multicolumn{1}{c|}{-}                                                               & \multicolumn{1}{c|}{-}                                                               & -                                                               & \multicolumn{1}{c|}{0.5726}                                                          & \multicolumn{1}{c|}{-}                                                               & \multicolumn{1}{c|}{-}                                                               & -                                                               \\ \cline{2-10} 
\multicolumn{1}{|c|}{}                                                                                  & IA$^3$ - base                                                         & \multicolumn{1}{c|}{-}                                                               & \multicolumn{1}{c|}{-}                                                               & \multicolumn{1}{c|}{-}                                                               & -                                                               & \multicolumn{1}{c|}{-}                                                               & \multicolumn{1}{c|}{-}                                                               & \multicolumn{1}{c|}{-}                                                               & -                                                               \\ \cline{2-10} 
\multicolumn{1}{|c|}{}                                                                                  & \begin{tabular}[c]{@{}l@{}}Prompt \\ Tuning\\  - base\end{tabular} & \multicolumn{1}{c|}{0.8801}                                                          & \multicolumn{1}{c|}{-}                                                               & \multicolumn{1}{c|}{\textbf{-}}                                                      & -                                                               & \multicolumn{1}{c|}{0.7871}                                                          & \multicolumn{1}{c|}{-}                                                               & \multicolumn{1}{c|}{-}                                                               & -                                                               \\ \cline{2-10} 
\multicolumn{1}{|c|}{}                                                                                  & \begin{tabular}[c]{@{}l@{}}P-Tuning\\  - base\end{tabular}         & \multicolumn{1}{c|}{0.8234}                                                          & \multicolumn{1}{c|}{0.4436}                                                          & \multicolumn{1}{c|}{-}                                                               & 0.4599                                                          & \multicolumn{1}{c|}{0.8465}                                                          & \multicolumn{1}{c|}{0.5353}                                                          & \multicolumn{1}{c|}{-}                                                               & -                                                               \\ \hline
\multicolumn{1}{|c|}{\multirow{11}{*}{\textbf{Model}}}                                                  & All models                                                         & \multicolumn{1}{c|}{0.232}                                                           & \multicolumn{1}{c|}{-}                                                               & \multicolumn{1}{c|}{0.198}                                                           & -                                                               & \multicolumn{1}{c|}{0.153}                                                           & \multicolumn{1}{c|}{0.25}                                                            & \multicolumn{1}{c|}{0.212}                                                           & 0.252                                                           \\ \cline{2-10} 
\multicolumn{1}{|c|}{}                                                                                  & \begin{tabular}[c]{@{}l@{}}Llama - \\ Mistral\end{tabular}         & \multicolumn{1}{c|}{-}                                                               & \multicolumn{1}{c|}{-}                                                               & \multicolumn{1}{c|}{-}                                                               & -                                                               & \multicolumn{1}{c|}{-}                                                               & \multicolumn{1}{c|}{-}                                                               & \multicolumn{1}{c|}{-}                                                               & \begin{tabular}[c]{@{}c@{}}0.8125 \\ (Mistral)\end{tabular}     \\ \cline{2-10} 
\multicolumn{1}{|c|}{}                                                                                  & \begin{tabular}[c]{@{}l@{}}Llama - \\ Qwen\end{tabular}            & \multicolumn{1}{c|}{\begin{tabular}[c]{@{}c@{}}0.8451 \\ (Qwen)\end{tabular}}        & \multicolumn{1}{c|}{-}                                                               & \multicolumn{1}{c|}{-}                                                               & -                                                               & \multicolumn{1}{c|}{\begin{tabular}[c]{@{}c@{}}0.7808\\ (Qwen)\end{tabular}}         & \multicolumn{1}{c|}{\begin{tabular}[c]{@{}c@{}}1.0066 \\ (Qwen)\end{tabular}}        & \multicolumn{1}{c|}{-}                                                               & \begin{tabular}[c]{@{}c@{}}0.9184 \\ (Qwen)\end{tabular}        \\ \cline{2-10} 
\multicolumn{1}{|c|}{}                                                                                  & \begin{tabular}[c]{@{}l@{}}Llama - \\ Gemma\end{tabular}           & \multicolumn{1}{c|}{\begin{tabular}[c]{@{}c@{}}0.8451 \\ (Gemma)\end{tabular}}       & \multicolumn{1}{c|}{-}                                                               & \multicolumn{1}{c|}{\begin{tabular}[c]{@{}c@{}}0.8451\\ (Gemma)\end{tabular}}        & -                                                               & \multicolumn{1}{c|}{\begin{tabular}[c]{@{}c@{}}0.7510\\ (Gemma)\end{tabular}}        & \multicolumn{1}{c|}{-}                                                               & \multicolumn{1}{c|}{-}                                                               & \begin{tabular}[c]{@{}c@{}}0.8125 \\ (Gemma)\end{tabular}       \\ \cline{2-10} 
\multicolumn{1}{|c|}{}                                                                                  & \begin{tabular}[c]{@{}l@{}}Mistral -\\  Qwen\end{tabular}          & \multicolumn{1}{c|}{-}                                                               & \multicolumn{1}{c|}{-}                                                               & \multicolumn{1}{c|}{-}                                                               & -                                                               & \multicolumn{1}{c|}{-}                                                               & \multicolumn{1}{c|}{-}                                                               & \multicolumn{1}{c|}{-}                                                               & -                                                               \\ \cline{2-10} 
\multicolumn{1}{|c|}{}                                                                                  & \begin{tabular}[c]{@{}l@{}}Mistral - \\ Gemma\end{tabular}         & \multicolumn{1}{c|}{\begin{tabular}[c]{@{}c@{}}0.9518 \\ (Gemma)\end{tabular}}       & \multicolumn{1}{c|}{-}                                                               & \multicolumn{1}{c|}{\begin{tabular}[c]{@{}c@{}}0.8125\\ (Gemma)\end{tabular}}        & -                                                               & \multicolumn{1}{c|}{-}                                                               & \multicolumn{1}{c|}{\begin{tabular}[c]{@{}c@{}}0.7808 \\ (Mistral)\end{tabular}}     & \multicolumn{1}{c|}{\begin{tabular}[c]{@{}c@{}}1.0066 \\ (Gemma)\end{tabular}}       & -                                                               \\ \cline{2-10} 
\multicolumn{1}{|c|}{}                                                                                  & \begin{tabular}[c]{@{}l@{}}Qwen -\\  Gemma\end{tabular}            & \multicolumn{1}{c|}{-}                                                               & \multicolumn{1}{c|}{-}                                                               & \multicolumn{1}{c|}{\begin{tabular}[c]{@{}c@{}}0.8748\\ (Gemma)\end{tabular}}        & -                                                               & \multicolumn{1}{c|}{-}                                                               & \multicolumn{1}{c|}{-}                                                               & \multicolumn{1}{c|}{-}                                                               & -                                                               \\ \cline{2-10} 
\multicolumn{1}{|c|}{}                                                                                  & \begin{tabular}[c]{@{}l@{}}Llama - \\ base\end{tabular}            & \multicolumn{1}{c|}{0.4683}                                                          & \multicolumn{1}{c|}{-}                                                               & \multicolumn{1}{c|}{\textbf{-}}                                                      & -                                                               & \multicolumn{1}{c|}{0.4957}                                                          & \multicolumn{1}{c|}{0.7683}                                                          & \multicolumn{1}{c|}{-}                                                               & -                                                               \\ \cline{2-10} 
\multicolumn{1}{|c|}{}                                                                                  & \begin{tabular}[c]{@{}l@{}}Mistral - \\ base\end{tabular}          & \multicolumn{1}{c|}{0.9318}                                                          & \multicolumn{1}{c|}{-}                                                               & \multicolumn{1}{c|}{\textbf{-}}                                                      & -                                                               & \multicolumn{1}{c|}{0.9199}                                                          & \multicolumn{1}{c|}{-}                                                               & \multicolumn{1}{c|}{0.6956}                                                          & -                                                               \\ \cline{2-10} 
\multicolumn{1}{|c|}{}                                                                                  & \begin{tabular}[c]{@{}l@{}}Qwen - \\ base\end{tabular}             & \multicolumn{1}{c|}{0.8791}                                                          & \multicolumn{1}{c|}{\textbf{0.6854}}                                                 & \multicolumn{1}{c|}{0.8791}                                                          & -                                                               & \multicolumn{1}{c|}{0.4801}                                                          & \multicolumn{1}{c|}{-}                                                               & \multicolumn{1}{c|}{-}                                                               & 0.5987                                                          \\ \cline{2-10} 
\multicolumn{1}{|c|}{}                                                                                  & \begin{tabular}[c]{@{}l@{}}Gemma - \\ base\end{tabular}            & \multicolumn{1}{c|}{0.517}                                                           & \multicolumn{1}{c|}{0.543}                                                           & \multicolumn{1}{c|}{\textbf{0.9027}}                                                 & -                                                               & \multicolumn{1}{c|}{0.5495}                                                          & \multicolumn{1}{c|}{0.7305}                                                          & \multicolumn{1}{c|}{\textbf{0.667}}                                                  & -                                                               \\ \hline
\end{tabular}
\end{table*}

\begin{table*}[htbp]
\caption{\centering Results of the statistical tests for fairness categories 5 and 6}
\label{tab:fairness_categories_5}
\renewcommand{\arraystretch}{1.1}
\centering
\scriptsize

\begin{tabular}{|cl|cccc|cccc|}
\hline
\multicolumn{1}{|c|}{\multirow{3}{*}{\textbf{Factor}}}                                                  & \multicolumn{1}{c|}{\multirow{3}{*}{\textbf{Comparison}}}          & \multicolumn{4}{c|}{\textbf{5. Physical Appearance}}                                                                                                                                                                                                                                                                                 & \multicolumn{4}{c|}{\textbf{6. Race/Ethnicity}}                                                                                                                                                                                                                                                                                      \\ \cline{3-10} 
\multicolumn{1}{|c|}{}                                                                                  & \multicolumn{1}{c|}{}                                              & \multicolumn{1}{c|}{\textbf{\begin{tabular}[c]{@{}c@{}}Acc. \\ AMB\end{tabular}}}    & \multicolumn{1}{c|}{\textbf{\begin{tabular}[c]{@{}c@{}}Acc. \\ DIS\end{tabular}}}    & \multicolumn{1}{c|}{\textbf{\begin{tabular}[c]{@{}c@{}}Bias\\  AMB\end{tabular}}}    & \textbf{\begin{tabular}[c]{@{}c@{}}Bias\\ DIS\end{tabular}}     & \multicolumn{1}{c|}{\textbf{\begin{tabular}[c]{@{}c@{}}Acc.\\ AMB\end{tabular}}}     & \multicolumn{1}{c|}{\textbf{\begin{tabular}[c]{@{}c@{}}Acc.\\ DIS\end{tabular}}}     & \multicolumn{1}{c|}{\textbf{\begin{tabular}[c]{@{}c@{}}Bias\\ AMB\end{tabular}}}     & \textbf{\begin{tabular}[c]{@{}c@{}}Bias\\ DIS\end{tabular}}     \\ \cline{3-10} 
\multicolumn{1}{|c|}{}                                                                                  & \multicolumn{1}{c|}{}                                              & \multicolumn{1}{c|}{\textbf{\begin{tabular}[c]{@{}c@{}}Effect \\ Size\end{tabular}}} & \multicolumn{1}{c|}{\textbf{\begin{tabular}[c]{@{}c@{}}Effect \\ Size\end{tabular}}} & \multicolumn{1}{c|}{\textbf{\begin{tabular}[c]{@{}c@{}}Effect \\ Size\end{tabular}}} & \textbf{\begin{tabular}[c]{@{}c@{}}Effect \\ Size\end{tabular}} & \multicolumn{1}{c|}{\textbf{\begin{tabular}[c]{@{}c@{}}Effect \\ Size\end{tabular}}} & \multicolumn{1}{c|}{\textbf{\begin{tabular}[c]{@{}c@{}}Effect \\ Size\end{tabular}}} & \multicolumn{1}{c|}{\textbf{\begin{tabular}[c]{@{}c@{}}Effect \\ Size\end{tabular}}} & \textbf{\begin{tabular}[c]{@{}c@{}}Effect \\ Size\end{tabular}} \\ \hline
\multicolumn{2}{|c|}{\textbf{All results}}                                                                                                                                   & \multicolumn{1}{c|}{0.6519}                                                          & \multicolumn{1}{c|}{-}                                                               & \multicolumn{1}{c|}{\textbf{0.2154}}                                                 & 0.3917                                                          & \multicolumn{1}{c|}{0.533}                                                           & \multicolumn{1}{c|}{0.5677}                                                          & \multicolumn{1}{c|}{-}                                                               & 0.5676                                                          \\ \hline
\multicolumn{1}{|c|}{\multirow{3}{*}{\textbf{Dataset}}}                                                 & UC - UF                                                            & \multicolumn{1}{c|}{\begin{tabular}[c]{@{}c@{}}0.6341 \\ (UF)\end{tabular}}          & \multicolumn{1}{c|}{-}                                                               & \multicolumn{1}{c|}{-}                                                               & -                                                               & \multicolumn{1}{c|}{-}                                                               & \multicolumn{1}{c|}{-}                                                               & \multicolumn{1}{c|}{\begin{tabular}[c]{@{}c@{}}0.5718 \\ (UC)\end{tabular}}          & -                                                               \\ \cline{2-10} 
\multicolumn{1}{|c|}{}                                                                                  & UC - base                                                          & \multicolumn{1}{c|}{0.7656}                                                          & \multicolumn{1}{c|}{-}                                                               & \multicolumn{1}{c|}{-}                                                               & 0.6104                                                          & \multicolumn{1}{c|}{-}                                                               & \multicolumn{1}{c|}{0.6713}                                                          & \multicolumn{1}{c|}{-}                                                               & 0.7573                                                          \\ \cline{2-10} 
\multicolumn{1}{|c|}{}                                                                                  & UF - base                                                          & \multicolumn{1}{c|}{0.6351}                                                          & \multicolumn{1}{c|}{-}                                                               & \multicolumn{1}{c|}{\textbf{0.2671}}                                                 & 0.3409                                                          & \multicolumn{1}{c|}{0.5422}                                                          & \multicolumn{1}{c|}{0.5469}                                                          & \multicolumn{1}{c|}{-}                                                               & 0.5343                                                          \\ \hline
\multicolumn{1}{|c|}{\multirow{3}{*}{\textbf{Paradigm}}}                                                & SFT - DPO                                                          & \multicolumn{1}{c|}{\begin{tabular}[c]{@{}c@{}}0.7371 \\ (DPO)\end{tabular}}         & \multicolumn{1}{c|}{-}                                                               & \multicolumn{1}{c|}{\begin{tabular}[c]{@{}c@{}}0.5363 \\ (DPO)\end{tabular}}         & -                                                               & \multicolumn{1}{c|}{\begin{tabular}[c]{@{}c@{}}0.7302 \\ (DPO)\end{tabular}}         & \multicolumn{1}{c|}{\begin{tabular}[c]{@{}c@{}}0.6098 \\ (DPO)\end{tabular}}         & \multicolumn{1}{c|}{-}                                                               & \begin{tabular}[c]{@{}c@{}}0.8561 \\ (DPO)\end{tabular}         \\ \cline{2-10} 
\multicolumn{1}{|c|}{}                                                                                  & SFT - base                                                         & \multicolumn{1}{c|}{0.7682}                                                          & \multicolumn{1}{c|}{-}                                                               & \multicolumn{1}{c|}{\textbf{-}}                                                      & 0.4457                                                          & \multicolumn{1}{c|}{0.6156}                                                          & \multicolumn{1}{c|}{0.6722}                                                          & \multicolumn{1}{c|}{-}                                                               & 0.6719                                                          \\ \cline{2-10} 
\multicolumn{1}{|c|}{}                                                                                  & DPO - base                                                         & \multicolumn{1}{c|}{-}                                                               & \multicolumn{1}{c|}{\textbf{-}}                                                      & \multicolumn{1}{c|}{-}                                                               & \textbf{-}                                                      & \multicolumn{1}{c|}{-}                                                               & \multicolumn{1}{c|}{-}                                                               & \multicolumn{1}{c|}{-}                                                               & \textbf{-}                                                      \\ \hline
\multicolumn{1}{|c|}{\multirow{3}{*}{\textbf{\begin{tabular}[c]{@{}c@{}}Learning\\ Rate\end{tabular}}}} & 1e-3 - 2e-5                                                        & \multicolumn{1}{c|}{\begin{tabular}[c]{@{}c@{}}0.4583 \\ (2e-5)\end{tabular}}        & \multicolumn{1}{c|}{-}                                                               & \multicolumn{1}{c|}{-}                                                               & -                                                               & \multicolumn{1}{c|}{\begin{tabular}[c]{@{}c@{}}0.5951 \\ (2e-5)\end{tabular}}        & \multicolumn{1}{c|}{-}                                                               & \multicolumn{1}{c|}{-}                                                               & -                                                               \\ \cline{2-10} 
\multicolumn{1}{|c|}{}                                                                                  & 1e-3 - base                                                        & \multicolumn{1}{c|}{0.7443}                                                          & \multicolumn{1}{c|}{-}                                                               & \multicolumn{1}{c|}{-}                                                               & 0.4709                                                          & \multicolumn{1}{c|}{0.7327}                                                          & \multicolumn{1}{c|}{0.6134}                                                          & \multicolumn{1}{c|}{-}                                                               & 0.7044                                                          \\ \cline{2-10} 
\multicolumn{1}{|c|}{}                                                                                  & 2e-5 - base                                                        & \multicolumn{1}{c|}{0.6519}                                                          & \multicolumn{1}{c|}{-}                                                               & \multicolumn{1}{c|}{\textbf{-}}                                                      & 0.3367                                                          & \multicolumn{1}{c|}{0.4262}                                                          & \multicolumn{1}{c|}{0.5519}                                                          & \multicolumn{1}{c|}{-}                                                               & 0.5084                                                          \\ \hline
\multicolumn{1}{|c|}{\multirow{3}{*}{\textbf{Epochs}}}                                                  & 1e - 5e                                                            & \multicolumn{1}{c|}{-}                                                               & \multicolumn{1}{c|}{-}                                                               & \multicolumn{1}{c|}{\begin{tabular}[c]{@{}c@{}}0.5661 \\ (1e)\end{tabular}}          & -                                                               & \multicolumn{1}{c|}{-}                                                               & \multicolumn{1}{c|}{-}                                                               & \multicolumn{1}{c|}{-}                                                               & -                                                               \\ \cline{2-10} 
\multicolumn{1}{|c|}{}                                                                                  & 1e - base                                                          & \multicolumn{1}{c|}{0.7292}                                                          & \multicolumn{1}{c|}{-}                                                               & \multicolumn{1}{c|}{\textbf{-}}                                                      & 0.3915                                                          & \multicolumn{1}{c|}{0.4346}                                                          & \multicolumn{1}{c|}{0.652}                                                           & \multicolumn{1}{c|}{-}                                                               & 0.5923                                                          \\ \cline{2-10} 
\multicolumn{1}{|c|}{}                                                                                  & 5e - base                                                          & \multicolumn{1}{c|}{0.6348}                                                          & \multicolumn{1}{c|}{-}                                                               & \multicolumn{1}{c|}{\textbf{0.3083}}                                                 & 0.3915                                                          & \multicolumn{1}{c|}{0.5721}                                                          & \multicolumn{1}{c|}{0.5296}                                                          & \multicolumn{1}{c|}{-}                                                               & 0.5454                                                          \\ \hline
\multicolumn{1}{|c|}{\multirow{11}{*}{\textbf{\begin{tabular}[c]{@{}c@{}}PEFT\\ Method\end{tabular}}}}  & All methods                                                        & \multicolumn{1}{c|}{0.187}                                                           & \multicolumn{1}{c|}{-}                                                               & \multicolumn{1}{c|}{-}                                                               & 0.197                                                           & \multicolumn{1}{c|}{0.176}                                                           & \multicolumn{1}{c|}{0.266}                                                           & \multicolumn{1}{c|}{-}                                                               & 0.247                                                           \\ \cline{2-10} 
\multicolumn{1}{|c|}{}                                                                                  & Lora - IA$^3$                                                         & \multicolumn{1}{c|}{\begin{tabular}[c]{@{}c@{}}0.8153\\ (IA$^3$)\end{tabular}}          & \multicolumn{1}{c|}{-}                                                               & \multicolumn{1}{c|}{-}                                                               & -                                                               & \multicolumn{1}{c|}{\begin{tabular}[c]{@{}c@{}}0.6582 \\ (IA$^3$)\end{tabular}}         & \multicolumn{1}{c|}{-}                                                               & \multicolumn{1}{c|}{-}                                                               & -                                                               \\ \cline{2-10} 
\multicolumn{1}{|c|}{}                                                                                  & \begin{tabular}[c]{@{}l@{}}IA$^3$ - \\ Prompt \\ Tuning\end{tabular}  & \multicolumn{1}{c|}{-}                                                               & \multicolumn{1}{c|}{-}                                                               & \multicolumn{1}{c|}{-}                                                               & \begin{tabular}[c]{@{}c@{}}0.7159\\  (IA$^3$)\end{tabular}         & \multicolumn{1}{c|}{-}                                                               & \multicolumn{1}{c|}{\begin{tabular}[c]{@{}c@{}}0.8979 \\ (IA$^3$)\end{tabular}}         & \multicolumn{1}{c|}{-}                                                               & \begin{tabular}[c]{@{}c@{}}1.0113 \\ (IA$^3$)\end{tabular}         \\ \cline{2-10} 
\multicolumn{1}{|c|}{}                                                                                  & \begin{tabular}[c]{@{}l@{}}IA$^3$ - \\ P-Tuning\end{tabular}          & \multicolumn{1}{c|}{\begin{tabular}[c]{@{}c@{}}0.8979 \\ (IA$^3$)\end{tabular}}         & \multicolumn{1}{c|}{-}                                                               & \multicolumn{1}{c|}{-}                                                               & \begin{tabular}[c]{@{}c@{}}0.7011 \\ (IA$^3$)\end{tabular}         & \multicolumn{1}{c|}{\begin{tabular}[c]{@{}c@{}}0.9184 \\ (IA$^3$)\end{tabular}}         & \multicolumn{1}{c|}{\begin{tabular}[c]{@{}c@{}}0.8616 \\ (IA$^3$)\end{tabular}}         & \multicolumn{1}{c|}{-}                                                               & \begin{tabular}[c]{@{}c@{}}0.7620 \\ (IA$^3$)\end{tabular}         \\ \cline{2-10} 
\multicolumn{1}{|c|}{}                                                                                  & \begin{tabular}[c]{@{}l@{}}Lora - \\ Prompt \\ Tuning\end{tabular} & \multicolumn{1}{c|}{-}                                                               & \multicolumn{1}{c|}{-}                                                               & \multicolumn{1}{c|}{-}                                                               & \begin{tabular}[c]{@{}c@{}}0.6582 \\ (Lora)\end{tabular}        & \multicolumn{1}{c|}{-}                                                               & \multicolumn{1}{c|}{-}                                                               & \multicolumn{1}{c|}{-}                                                               & \begin{tabular}[c]{@{}c@{}}0.6442 \\ (Lora)\end{tabular}        \\ \cline{2-10} 
\multicolumn{1}{|c|}{}                                                                                  & \begin{tabular}[c]{@{}l@{}}Lora - \\ P-Tuning\end{tabular}         & \multicolumn{1}{c|}{-}                                                               & \multicolumn{1}{c|}{-}                                                               & \multicolumn{1}{c|}{-}                                                               & -                                                               & \multicolumn{1}{c|}{-}                                                               & \multicolumn{1}{c|}{-}                                                               & \multicolumn{1}{c|}{-}                                                               & \begin{tabular}[c]{@{}c@{}}0.6442 \\ (Lora)\end{tabular}        \\ \cline{2-10} 
\multicolumn{1}{|c|}{}                                                                                  & \begin{tabular}[c]{@{}l@{}}Prompt - \\ P-Tuning\end{tabular}       & \multicolumn{1}{c|}{-}                                                               & \multicolumn{1}{c|}{-}                                                               & \multicolumn{1}{c|}{-}                                                               & -                                                               & \multicolumn{1}{c|}{-}                                                               & \multicolumn{1}{c|}{-}                                                               & \multicolumn{1}{c|}{-}                                                               & -                                                               \\ \cline{2-10} 
\multicolumn{1}{|c|}{}                                                                                  & Lora - base                                                        & \multicolumn{1}{c|}{0.7787}                                                          & \multicolumn{1}{c|}{-}                                                               & \multicolumn{1}{c|}{0.5457}                                                          & -                                                               & \multicolumn{1}{c|}{0.5003}                                                          & \multicolumn{1}{c|}{-}                                                               & \multicolumn{1}{c|}{-}                                                               &                                                                 \\ \cline{2-10} 
\multicolumn{1}{|c|}{}                                                                                  & IA$^3$ - base                                                         & \multicolumn{1}{c|}{0.6661}                                                          & \multicolumn{1}{c|}{-}                                                               & \multicolumn{1}{c|}{0.6784}                                                          & -                                                               & \multicolumn{1}{c|}{-}                                                               & \multicolumn{1}{c|}{-}                                                               & \multicolumn{1}{c|}{-}                                                               & -                                                               \\ \cline{2-10} 
\multicolumn{1}{|c|}{}                                                                                  & \begin{tabular}[c]{@{}l@{}}Prompt \\ Tuning\\  - base\end{tabular} & \multicolumn{1}{c|}{0.6039}                                                          & \multicolumn{1}{c|}{-}                                                               & \multicolumn{1}{c|}{\textbf{-}}                                                      & 0.6656                                                          & \multicolumn{1}{c|}{0.6622}                                                          & \multicolumn{1}{c|}{0.9523}                                                          & \multicolumn{1}{c|}{-}                                                               & 1.0194                                                          \\ \cline{2-10} 
\multicolumn{1}{|c|}{}                                                                                  & \begin{tabular}[c]{@{}l@{}}P-Tuning\\  - base\end{tabular}         & \multicolumn{1}{c|}{0.8826}                                                          & \multicolumn{1}{c|}{-}                                                               & \multicolumn{1}{c|}{-}                                                               & 0.5875                                                          & \multicolumn{1}{c|}{0.8123}                                                          & \multicolumn{1}{c|}{0.8702}                                                          & \multicolumn{1}{c|}{-}                                                               & 0.8123                                                          \\ \hline
\multicolumn{1}{|c|}{\multirow{11}{*}{\textbf{Model}}}                                                  & All models                                                         & \multicolumn{1}{c|}{0.222}                                                           & \multicolumn{1}{c|}{0.132}                                                           & \multicolumn{1}{c|}{-}                                                               & -                                                               & \multicolumn{1}{c|}{0.262}                                                           & \multicolumn{1}{c|}{-}                                                               & \multicolumn{1}{c|}{0.335}                                                           & 0.265                                                           \\ \cline{2-10} 
\multicolumn{1}{|c|}{}                                                                                  & \begin{tabular}[c]{@{}l@{}}Llama - \\ Mistral\end{tabular}         & \multicolumn{1}{c|}{-}                                                               & \multicolumn{1}{c|}{\begin{tabular}[c]{@{}c@{}}0.8125 \\ (Mistral)\end{tabular}}     & \multicolumn{1}{c|}{-}                                                               & -                                                               & \multicolumn{1}{c|}{\begin{tabular}[c]{@{}c@{}}0.7808 \\ (Mistral)\end{tabular}}     & \multicolumn{1}{c|}{-}                                                               & \multicolumn{1}{c|}{\begin{tabular}[c]{@{}c@{}}0.8451 \\ (Llama)\end{tabular}}       & \begin{tabular}[c]{@{}c@{}}0.7808 \\ (Mistral)\end{tabular}     \\ \cline{2-10} 
\multicolumn{1}{|c|}{}                                                                                  & \begin{tabular}[c]{@{}l@{}}Llama - \\ Qwen\end{tabular}            & \multicolumn{1}{c|}{\begin{tabular}[c]{@{}c@{}}1.0066 \\ (Qwen)\end{tabular}}        & \multicolumn{1}{c|}{\begin{tabular}[c]{@{}c@{}}0.7879 \\ (Qwen)\end{tabular}}        & \multicolumn{1}{c|}{-}                                                               & -                                                               & \multicolumn{1}{c|}{\begin{tabular}[c]{@{}c@{}}0.7808 \\ (Qwen)\end{tabular}}        & \multicolumn{1}{c|}{-}                                                               & \multicolumn{1}{c|}{\begin{tabular}[c]{@{}c@{}}1.0066 \\ (Llama)\end{tabular}}       & \begin{tabular}[c]{@{}c@{}}1.0066 \\ (Qwen)\end{tabular}        \\ \cline{2-10} 
\multicolumn{1}{|c|}{}                                                                                  & \begin{tabular}[c]{@{}l@{}}Llama - \\ Gemma\end{tabular}           & \multicolumn{1}{c|}{-}                                                               & \multicolumn{1}{c|}{-}                                                               & \multicolumn{1}{c|}{-}                                                               & -                                                               & \multicolumn{1}{c|}{\begin{tabular}[c]{@{}c@{}}1.0066 \\ (Gemma)\end{tabular}}       & \multicolumn{1}{c|}{-}                                                               & \multicolumn{1}{c|}{-}                                                               & -                                                               \\ \cline{2-10} 
\multicolumn{1}{|c|}{}                                                                                  & \begin{tabular}[c]{@{}l@{}}Mistral -\\  Qwen\end{tabular}          & \multicolumn{1}{c|}{-}                                                               & \multicolumn{1}{c|}{-}                                                               & \multicolumn{1}{c|}{-}                                                               & -                                                               & \multicolumn{1}{c|}{-}                                                               & \multicolumn{1}{c|}{-}                                                               & \multicolumn{1}{c|}{-}                                                               & -                                                               \\ \cline{2-10} 
\multicolumn{1}{|c|}{}                                                                                  & \begin{tabular}[c]{@{}l@{}}Mistral - \\ Gemma\end{tabular}         & \multicolumn{1}{c|}{-}                                                               & \multicolumn{1}{c|}{-}                                                               & \multicolumn{1}{c|}{-}                                                               & -                                                               & \multicolumn{1}{c|}{-}                                                               & \multicolumn{1}{c|}{-}                                                               & \multicolumn{1}{c|}{-}                                                               & -                                                               \\ \cline{2-10} 
\multicolumn{1}{|c|}{}                                                                                  & \begin{tabular}[c]{@{}l@{}}Qwen -\\  Gemma\end{tabular}            & \multicolumn{1}{c|}{-}                                                               & \multicolumn{1}{c|}{\begin{tabular}[c]{@{}c@{}}0.7808 \\ (Qwen)\end{tabular}}        & \multicolumn{1}{c|}{-}                                                               & -                                                               & \multicolumn{1}{c|}{-}                                                               & \multicolumn{1}{c|}{-}                                                               & \multicolumn{1}{c|}{\begin{tabular}[c]{@{}c@{}}0.7808 \\ (Gemma)\end{tabular}}       & -                                                               \\ \cline{2-10} 
\multicolumn{1}{|c|}{}                                                                                  & \begin{tabular}[c]{@{}l@{}}Llama - \\ base\end{tabular}            & \multicolumn{1}{c|}{0.5134}                                                          & \multicolumn{1}{c|}{0.718}                                                           & \multicolumn{1}{c|}{\textbf{-}}                                                      & -                                                               & \multicolumn{1}{c|}{0.5009}                                                          & \multicolumn{1}{c|}{0.5539}                                                          & \multicolumn{1}{c|}{\textbf{1.0333}}                                                 & 0.8807                                                          \\ \cline{2-10} 
\multicolumn{1}{|c|}{}                                                                                  & \begin{tabular}[c]{@{}l@{}}Mistral - \\ base\end{tabular}          & \multicolumn{1}{c|}{0.8685}                                                          & \multicolumn{1}{c|}{\textbf{0.4399}}                                                 & \multicolumn{1}{c|}{\textbf{0.8533}}                                                 & 0.4503                                                          & \multicolumn{1}{c|}{0.7262}                                                          & \multicolumn{1}{c|}{0.7717}                                                          & \multicolumn{1}{c|}{0.5653}                                                          & -                                                               \\ \cline{2-10} 
\multicolumn{1}{|c|}{}                                                                                  & \begin{tabular}[c]{@{}l@{}}Qwen - \\ base\end{tabular}             & \multicolumn{1}{c|}{-}                                                               & \multicolumn{1}{c|}{\textbf{0.4929}}                                                 & \multicolumn{1}{c|}{-}                                                               & -                                                               & \multicolumn{1}{c|}{-}                                                               & \multicolumn{1}{c|}{-}                                                               & \multicolumn{1}{c|}{0.8764}                                                          & 0.6475                                                          \\ \cline{2-10} 
\multicolumn{1}{|c|}{}                                                                                  & \begin{tabular}[c]{@{}l@{}}Gemma - \\ base\end{tabular}            & \multicolumn{1}{c|}{0.8792}                                                          & \multicolumn{1}{c|}{0.802}                                                           & \multicolumn{1}{c|}{\textbf{0.5752}}                                                 & 0.9027                                                          & \multicolumn{1}{c|}{0.517}                                                           & \multicolumn{1}{c|}{0.724}                                                           & \multicolumn{1}{c|}{\textbf{0.7504}}                                                 & 0.6358                                                          \\ \hline
\end{tabular}
\end{table*}

\begin{table*}[htbp]
\caption{\centering Results of the statistical tests for fairness categories 7 and 8}
\label{tab:fairness_categories_7}
\renewcommand{\arraystretch}{1.1}
\centering
\scriptsize

\begin{tabular}{|cl|cccc|cccc|}
\hline
\multicolumn{1}{|c|}{\multirow{3}{*}{\textbf{Factor}}}                                                  & \multicolumn{1}{c|}{\multirow{3}{*}{\textbf{Comparison}}}          & \multicolumn{4}{c|}{\textbf{7. Religion}}                                                                                                                                                                                                                                                                                            & \multicolumn{4}{c|}{\textbf{8. Socio-Economic Status}}                                                                                                                                                                                                                                                                               \\ \cline{3-10} 
\multicolumn{1}{|c|}{}                                                                                  & \multicolumn{1}{c|}{}                                              & \multicolumn{1}{c|}{\textbf{\begin{tabular}[c]{@{}c@{}}Acc. \\ AMB\end{tabular}}}    & \multicolumn{1}{c|}{\textbf{\begin{tabular}[c]{@{}c@{}}Acc. \\ DIS\end{tabular}}}    & \multicolumn{1}{c|}{\textbf{\begin{tabular}[c]{@{}c@{}}Bias\\  AMB\end{tabular}}}    & \textbf{\begin{tabular}[c]{@{}c@{}}Bias\\ DIS\end{tabular}}     & \multicolumn{1}{c|}{\textbf{\begin{tabular}[c]{@{}c@{}}Acc.\\ AMB\end{tabular}}}     & \multicolumn{1}{c|}{\textbf{\begin{tabular}[c]{@{}c@{}}Acc.\\ DIS\end{tabular}}}     & \multicolumn{1}{c|}{\textbf{\begin{tabular}[c]{@{}c@{}}Bias\\ AMB\end{tabular}}}     & \textbf{\begin{tabular}[c]{@{}c@{}}Bias\\ DIS\end{tabular}}     \\ \cline{3-10} 
\multicolumn{1}{|c|}{}                                                                                  & \multicolumn{1}{c|}{}                                              & \multicolumn{1}{c|}{\textbf{\begin{tabular}[c]{@{}c@{}}Effect \\ Size\end{tabular}}} & \multicolumn{1}{c|}{\textbf{\begin{tabular}[c]{@{}c@{}}Effect \\ Size\end{tabular}}} & \multicolumn{1}{c|}{\textbf{\begin{tabular}[c]{@{}c@{}}Effect \\ Size\end{tabular}}} & \textbf{\begin{tabular}[c]{@{}c@{}}Effect \\ Size\end{tabular}} & \multicolumn{1}{c|}{\textbf{\begin{tabular}[c]{@{}c@{}}Effect \\ Size\end{tabular}}} & \multicolumn{1}{c|}{\textbf{\begin{tabular}[c]{@{}c@{}}Effect \\ Size\end{tabular}}} & \multicolumn{1}{c|}{\textbf{\begin{tabular}[c]{@{}c@{}}Effect \\ Size\end{tabular}}} & \textbf{\begin{tabular}[c]{@{}c@{}}Effect \\ Size\end{tabular}} \\ \hline
\multicolumn{2}{|c|}{\textbf{All results}}                                                                                                                                   & \multicolumn{1}{c|}{0.7453}                                                          & \multicolumn{1}{c|}{0.4103}                                                          & \multicolumn{1}{c|}{\textbf{-}}                                                      & -                                                               & \multicolumn{1}{c|}{0.5641}                                                          & \multicolumn{1}{c|}{0.3472}                                                          & \multicolumn{1}{c|}{-}                                                               & -                                                               \\ \hline
\multicolumn{1}{|c|}{\multirow{3}{*}{\textbf{Dataset}}}                                                 & UC - UF                                                            & \multicolumn{1}{c|}{-}                                                               & \multicolumn{1}{c|}{-}                                                               & \multicolumn{1}{c|}{-}                                                               & -                                                               & \multicolumn{1}{c|}{-}                                                               & \multicolumn{1}{c|}{-}                                                               & \multicolumn{1}{c|}{-}                                                               & -                                                               \\ \cline{2-10} 
\multicolumn{1}{|c|}{}                                                                                  & UC - base                                                          & \multicolumn{1}{c|}{0.8819}                                                          & \multicolumn{1}{c|}{-}                                                               & \multicolumn{1}{c|}{-}                                                               & -                                                               & \multicolumn{1}{c|}{-}                                                               & \multicolumn{1}{c|}{0.6299}                                                          & \multicolumn{1}{c|}{-}                                                               & -                                                               \\ \cline{2-10} 
\multicolumn{1}{|c|}{}                                                                                  & UF - base                                                          & \multicolumn{1}{c|}{0.7246}                                                          & \multicolumn{1}{c|}{0.39}                                                            & \multicolumn{1}{c|}{\textbf{-}}                                                      & -                                                               & \multicolumn{1}{c|}{0.5799}                                                          & \multicolumn{1}{c|}{0.2897}                                                          & \multicolumn{1}{c|}{-}                                                               & -                                                               \\ \hline
\multicolumn{1}{|c|}{\multirow{3}{*}{\textbf{Paradigm}}}                                                & SFT - DPO                                                          & \multicolumn{1}{c|}{\begin{tabular}[c]{@{}c@{}}0.7926 \\ (DPO)\end{tabular}}         & \multicolumn{1}{c|}{-}                                                               & \multicolumn{1}{c|}{\begin{tabular}[c]{@{}c@{}}0.5423 \\ (DPO)\end{tabular}}         & -                                                               & \multicolumn{1}{c|}{\begin{tabular}[c]{@{}c@{}}0.8269 \\ (DPO)\end{tabular}}         & \multicolumn{1}{c|}{-}                                                               & \multicolumn{1}{c|}{\begin{tabular}[c]{@{}c@{}}0.5550 \\ (DPO)\end{tabular}}         & \begin{tabular}[c]{@{}c@{}}0.6795 \\ (DPO)\end{tabular}         \\ \cline{2-10} 
\multicolumn{1}{|c|}{}                                                                                  & SFT - base                                                         & \multicolumn{1}{c|}{0.8485}                                                          & \multicolumn{1}{c|}{0.4847}                                                          & \multicolumn{1}{c|}{\textbf{-}}                                                      & 0.2802                                                          & \multicolumn{1}{c|}{0.6698}                                                          & \multicolumn{1}{c|}{0.4982}                                                          & \multicolumn{1}{c|}{-}                                                               & -                                                               \\ \cline{2-10} 
\multicolumn{1}{|c|}{}                                                                                  & DPO - base                                                         & \multicolumn{1}{c|}{-}                                                               & \multicolumn{1}{c|}{\textbf{-}}                                                      & \multicolumn{1}{c|}{-}                                                               & \textbf{-}                                                      & \multicolumn{1}{c|}{-}                                                               & \multicolumn{1}{c|}{-}                                                               & \multicolumn{1}{c|}{-}                                                               & \textbf{-}                                                      \\ \hline
\multicolumn{1}{|c|}{\multirow{3}{*}{\textbf{\begin{tabular}[c]{@{}c@{}}Learning\\ Rate\end{tabular}}}} & 1e-3 - 2e-5                                                        & \multicolumn{1}{c|}{\begin{tabular}[c]{@{}c@{}}0.4172 \\ (2e-5)\end{tabular}}        & \multicolumn{1}{c|}{-}                                                               & \multicolumn{1}{c|}{-}                                                               & -                                                               & \multicolumn{1}{c|}{\begin{tabular}[c]{@{}c@{}}0.6010 \\ (2e-5)\end{tabular}}        & \multicolumn{1}{c|}{-}                                                               & \multicolumn{1}{c|}{-}                                                               & -                                                               \\ \cline{2-10} 
\multicolumn{1}{|c|}{}                                                                                  & 1e-3 - base                                                        & \multicolumn{1}{c|}{0.7768}                                                          & \multicolumn{1}{c|}{0.4304}                                                          & \multicolumn{1}{c|}{-}                                                               & -                                                               & \multicolumn{1}{c|}{0.762}                                                           & \multicolumn{1}{c|}{-}                                                               & \multicolumn{1}{c|}{-}                                                               & -                                                               \\ \cline{2-10} 
\multicolumn{1}{|c|}{}                                                                                  & 2e-5 - base                                                        & \multicolumn{1}{c|}{0.7269}                                                          & \multicolumn{1}{c|}{0.3902}                                                          & \multicolumn{1}{c|}{\textbf{-}}                                                      & -                                                               & \multicolumn{1}{c|}{0.4654}                                                          & \multicolumn{1}{c|}{0.427}                                                           & \multicolumn{1}{c|}{-}                                                               & -                                                               \\ \hline
\multicolumn{1}{|c|}{\multirow{3}{*}{\textbf{Epochs}}}                                                  & 1e - 5e                                                            & \multicolumn{1}{c|}{-}                                                               & \multicolumn{1}{c|}{-}                                                               & \multicolumn{1}{c|}{-}                                                               & -                                                               & \multicolumn{1}{c|}{-}                                                               & \multicolumn{1}{c|}{-}                                                               & \multicolumn{1}{c|}{-}                                                               & -                                                               \\ \cline{2-10} 
\multicolumn{1}{|c|}{}                                                                                  & 1e - base                                                          & \multicolumn{1}{c|}{0.8394}                                                          & \multicolumn{1}{c|}{0.5032}                                                          & \multicolumn{1}{c|}{\textbf{-}}                                                      & -                                                               & \multicolumn{1}{c|}{0.4439}                                                          & \multicolumn{1}{c|}{0.6736}                                                          & \multicolumn{1}{c|}{-}                                                               & -                                                               \\ \cline{2-10} 
\multicolumn{1}{|c|}{}                                                                                  & 5e - base                                                          & \multicolumn{1}{c|}{0.7188}                                                          & \multicolumn{1}{c|}{0.3537}                                                          & \multicolumn{1}{c|}{\textbf{0.2943}}                                                 & -                                                               & \multicolumn{1}{c|}{0.6092}                                                          & \multicolumn{1}{c|}{-}                                                               & \multicolumn{1}{c|}{-}                                                               & -                                                               \\ \hline
\multicolumn{1}{|c|}{\multirow{11}{*}{\textbf{\begin{tabular}[c]{@{}c@{}}PEFT\\ Method\end{tabular}}}}  & All methods                                                        & \multicolumn{1}{c|}{0.403}                                                           & \multicolumn{1}{c|}{0.37}                                                            & \multicolumn{1}{c|}{-}                                                               & -                                                               & \multicolumn{1}{c|}{0.266}                                                           & \multicolumn{1}{c|}{0.277}                                                           & \multicolumn{1}{c|}{0.144}                                                           & -                                                               \\ \cline{2-10} 
\multicolumn{1}{|c|}{}                                                                                  & Lora - IA$^3$                                                         & \multicolumn{1}{c|}{\begin{tabular}[c]{@{}c@{}}0.8784 \\ (IA$^3$)\end{tabular}}         & \multicolumn{1}{c|}{-}                                                               & \multicolumn{1}{c|}{-}                                                               & -                                                               & \multicolumn{1}{c|}{\begin{tabular}[c]{@{}c@{}}0.7866 \\ (IA$^3$)\end{tabular}}         & \multicolumn{1}{c|}{-}                                                               & \multicolumn{1}{c|}{-}                                                               & -                                                               \\ \cline{2-10} 
\multicolumn{1}{|c|}{}                                                                                  & \begin{tabular}[c]{@{}l@{}}IA$^3$ - \\ Prompt \\ Tuning\end{tabular}  & \multicolumn{1}{c|}{\begin{tabular}[c]{@{}c@{}}0.8100 \\ (IA$^3$)\end{tabular}}         & \multicolumn{1}{c|}{\begin{tabular}[c]{@{}c@{}}1.0489 \\ (IA$^3$)\end{tabular}}         & \multicolumn{1}{c|}{-}                                                               & -                                                               & \multicolumn{1}{c|}{\begin{tabular}[c]{@{}c@{}}0.6442 \\ (IA$^3$)\end{tabular}}         & \multicolumn{1}{c|}{\begin{tabular}[c]{@{}c@{}}0.8100 \\ (IA$^3$)\end{tabular}}         & \multicolumn{1}{c|}{-}                                                               & -                                                               \\ \cline{2-10} 
\multicolumn{1}{|c|}{}                                                                                  & \begin{tabular}[c]{@{}l@{}}IA$^3$ - \\ P-Tuning\end{tabular}          & \multicolumn{1}{c|}{\begin{tabular}[c]{@{}c@{}}1.0113 \\ (IA$^3$)\end{tabular}}         & \multicolumn{1}{c|}{-}                                                               & \multicolumn{1}{c|}{-}                                                               & -                                                               & \multicolumn{1}{c|}{\begin{tabular}[c]{@{}c@{}}0.9397 \\ (IA$^3$)\end{tabular}}         & \multicolumn{1}{c|}{-}                                                               & \multicolumn{1}{c|}{-}                                                               & -                                                               \\ \cline{2-10} 
\multicolumn{1}{|c|}{}                                                                                  & \begin{tabular}[c]{@{}l@{}}Lora - \\ Prompt \\ Tuning\end{tabular} & \multicolumn{1}{c|}{-}                                                               & \multicolumn{1}{c|}{\begin{tabular}[c]{@{}c@{}}0.7936 \\ (Lora)\end{tabular}}        & \multicolumn{1}{c|}{-}                                                               & -                                                               & \multicolumn{1}{c|}{-}                                                               & \multicolumn{1}{c|}{\begin{tabular}[c]{@{}c@{}}0.7620 \\ (Lora)\end{tabular}}        & \multicolumn{1}{c|}{-}                                                               & -                                                               \\ \cline{2-10} 
\multicolumn{1}{|c|}{}                                                                                  & \begin{tabular}[c]{@{}l@{}}Lora - \\ P-Tuning\end{tabular}         & \multicolumn{1}{c|}{-}                                                               & \multicolumn{1}{c|}{-}                                                               & \multicolumn{1}{c|}{-}                                                               & -                                                               & \multicolumn{1}{c|}{-}                                                               & \multicolumn{1}{c|}{-}                                                               & \multicolumn{1}{c|}{-}                                                               & -                                                               \\ \cline{2-10} 
\multicolumn{1}{|c|}{}                                                                                  & \begin{tabular}[c]{@{}l@{}}Prompt - \\ P-Tuning\end{tabular}       & \multicolumn{1}{c|}{-}                                                               & \multicolumn{1}{c|}{-}                                                               & \multicolumn{1}{c|}{-}                                                               & -                                                               & \multicolumn{1}{c|}{-}                                                               & \multicolumn{1}{c|}{-}                                                               & \multicolumn{1}{c|}{-}                                                               & -                                                               \\ \cline{2-10} 
\multicolumn{1}{|c|}{}                                                                                  & Lora - base                                                        & \multicolumn{1}{c|}{0.7787}                                                          & \multicolumn{1}{c|}{-}                                                               & \multicolumn{1}{c|}{-}                                                               & -                                                               & \multicolumn{1}{c|}{0.6875}                                                          & \multicolumn{1}{c|}{-}                                                               & \multicolumn{1}{c|}{-}                                                               & -                                                               \\ \cline{2-10} 
\multicolumn{1}{|c|}{}                                                                                  & IA$^3$ - base                                                         & \multicolumn{1}{c|}{-}                                                               & \multicolumn{1}{c|}{-}                                                               & \multicolumn{1}{c|}{-}                                                               & -                                                               & \multicolumn{1}{c|}{-}                                                               & \multicolumn{1}{c|}{-}                                                               & \multicolumn{1}{c|}{-}                                                               & -                                                               \\ \cline{2-10} 
\multicolumn{1}{|c|}{}                                                                                  & \begin{tabular}[c]{@{}l@{}}Prompt \\ Tuning\\  - base\end{tabular} & \multicolumn{1}{c|}{0.8675}                                                          & \multicolumn{1}{c|}{0.8675}                                                          & \multicolumn{1}{c|}{\textbf{-}}                                                      & -                                                               & \multicolumn{1}{c|}{0.6039}                                                          & \multicolumn{1}{c|}{0.7871}                                                          & \multicolumn{1}{c|}{-}                                                               & -                                                               \\ \cline{2-10} 
\multicolumn{1}{|c|}{}                                                                                  & \begin{tabular}[c]{@{}l@{}}P-Tuning\\  - base\end{tabular}         & \multicolumn{1}{c|}{0.8826}                                                          & \multicolumn{1}{c|}{0.5525}                                                          & \multicolumn{1}{c|}{-}                                                               & -                                                               & \multicolumn{1}{c|}{0.8465}                                                          & \multicolumn{1}{c|}{0.5014}                                                          & \multicolumn{1}{c|}{-}                                                               & -                                                               \\ \hline
\multicolumn{1}{|c|}{\multirow{11}{*}{\textbf{Model}}}                                                  & All models                                                         & \multicolumn{1}{c|}{0.147}                                                           & \multicolumn{1}{c|}{-}                                                               & \multicolumn{1}{c|}{0.212}                                                           & -                                                               & \multicolumn{1}{c|}{0.195}                                                           & \multicolumn{1}{c|}{-}                                                               & \multicolumn{1}{c|}{-}                                                               & 0.142                                                           \\ \cline{2-10} 
\multicolumn{1}{|c|}{}                                                                                  & \begin{tabular}[c]{@{}l@{}}Llama - \\ Mistral\end{tabular}         & \multicolumn{1}{c|}{-}                                                               & \multicolumn{1}{c|}{-}                                                               & \multicolumn{1}{c|}{\begin{tabular}[c]{@{}c@{}}0.8748 \\ (Llama)\end{tabular}}       & -                                                               & \multicolumn{1}{c|}{-}                                                               & \multicolumn{1}{c|}{-}                                                               & \multicolumn{1}{c|}{-}                                                               & -                                                               \\ \cline{2-10} 
\multicolumn{1}{|c|}{}                                                                                  & \begin{tabular}[c]{@{}l@{}}Llama - \\ Qwen\end{tabular}            & \multicolumn{1}{c|}{\begin{tabular}[c]{@{}c@{}}0.8155 \\ (Qwen)\end{tabular}}        & \multicolumn{1}{c|}{-}                                                               & \multicolumn{1}{c|}{-}                                                               & -                                                               & \multicolumn{1}{c|}{\begin{tabular}[c]{@{}c@{}}0.7808 \\ (Qwen)\end{tabular}}        & \multicolumn{1}{c|}{-}                                                               & \multicolumn{1}{c|}{-}                                                               & -                                                               \\ \cline{2-10} 
\multicolumn{1}{|c|}{}                                                                                  & \begin{tabular}[c]{@{}l@{}}Llama - \\ Gemma\end{tabular}           & \multicolumn{1}{c|}{-}                                                               & \multicolumn{1}{c|}{-}                                                               & \multicolumn{1}{c|}{-}                                                               & -                                                               & \multicolumn{1}{c|}{\begin{tabular}[c]{@{}c@{}}1.0066 \\ (Gemma)\end{tabular}}       & \multicolumn{1}{c|}{-}                                                               & \multicolumn{1}{c|}{-}                                                               & -                                                               \\ \cline{2-10} 
\multicolumn{1}{|c|}{}                                                                                  & \begin{tabular}[c]{@{}l@{}}Mistral -\\  Qwen\end{tabular}          & \multicolumn{1}{c|}{-}                                                               & \multicolumn{1}{c|}{-}                                                               & \multicolumn{1}{c|}{-}                                                               & -                                                               & \multicolumn{1}{c|}{-}                                                               & \multicolumn{1}{c|}{-}                                                               & \multicolumn{1}{c|}{-}                                                               & -                                                               \\ \cline{2-10} 
\multicolumn{1}{|c|}{}                                                                                  & \begin{tabular}[c]{@{}l@{}}Mistral - \\ Gemma\end{tabular}         & \multicolumn{1}{c|}{-}                                                               & \multicolumn{1}{c|}{-}                                                               & \multicolumn{1}{c|}{-}                                                               & -                                                               & \multicolumn{1}{c|}{-}                                                               & \multicolumn{1}{c|}{-}                                                               & \multicolumn{1}{c|}{-}                                                               & -                                                               \\ \cline{2-10} 
\multicolumn{1}{|c|}{}                                                                                  & \begin{tabular}[c]{@{}l@{}}Qwen -\\  Gemma\end{tabular}            & \multicolumn{1}{c|}{-}                                                               & \multicolumn{1}{c|}{-}                                                               & \multicolumn{1}{c|}{-}                                                               & -                                                               & \multicolumn{1}{c|}{-}                                                               & \multicolumn{1}{c|}{-}                                                               & \multicolumn{1}{c|}{-}                                                               & -                                                               \\ \cline{2-10} 
\multicolumn{1}{|c|}{}                                                                                  & \begin{tabular}[c]{@{}l@{}}Llama - \\ base\end{tabular}            & \multicolumn{1}{c|}{0.4986}                                                          & \multicolumn{1}{c|}{-}                                                               & \multicolumn{1}{c|}{\textbf{0.6512}}                                                 & -                                                               & \multicolumn{1}{c|}{0.4502}                                                          & \multicolumn{1}{c|}{0.4671}                                                          & \multicolumn{1}{c|}{\textbf{-}}                                                      & 0.5927                                                          \\ \cline{2-10} 
\multicolumn{1}{|c|}{}                                                                                  & \begin{tabular}[c]{@{}l@{}}Mistral - \\ base\end{tabular}          & \multicolumn{1}{c|}{0.8571}                                                          & \multicolumn{1}{c|}{\textbf{-}}                                                      & \multicolumn{1}{c|}{\textbf{0.7325}}                                                 & -                                                               & \multicolumn{1}{c|}{0.7717}                                                          & \multicolumn{1}{c|}{-}                                                               & \multicolumn{1}{c|}{0.522}                                                           & -                                                               \\ \cline{2-10} 
\multicolumn{1}{|c|}{}                                                                                  & \begin{tabular}[c]{@{}l@{}}Qwen - \\ base\end{tabular}             & \multicolumn{1}{c|}{0.8683}                                                          & \multicolumn{1}{c|}{\textbf{0.6848}}                                                 & \multicolumn{1}{c|}{0.644}                                                           & -                                                               & \multicolumn{1}{c|}{-}                                                               & \multicolumn{1}{c|}{-}                                                               & \multicolumn{1}{c|}{-}                                                               & 0.477                                                           \\ \cline{2-10} 
\multicolumn{1}{|c|}{}                                                                                  & \begin{tabular}[c]{@{}l@{}}Gemma - \\ base\end{tabular}            & \multicolumn{1}{c|}{0.8792}                                                          & \multicolumn{1}{c|}{0.5954}                                                          & \multicolumn{1}{c|}{\textbf{0.6358}}                                                 & 0.6853                                                          & \multicolumn{1}{c|}{0.716}                                                           & \multicolumn{1}{c|}{0.8822}                                                          & \multicolumn{1}{c|}{\textbf{0.8033}}                                                 & -                                                               \\ \hline
\end{tabular}
\end{table*}

\begin{table*}[htbp]
\caption{\centering Results of the statistical tests for fairness category 9}
\label{tab:fairness_categories_9}
\renewcommand{\arraystretch}{1.1}
\centering
\scriptsize
\begin{tabular}{|cl|cccc|}
\hline
\multicolumn{1}{|c|}{\multirow{3}{*}{\textbf{Factor}}}                                                  & \multicolumn{1}{c|}{\multirow{3}{*}{\textbf{Comparison}}}          & \multicolumn{4}{c|}{\textbf{9. Sexual Orientation}}                                                                                                                                                                                                                                                                                  \\ \cline{3-6} 
\multicolumn{1}{|c|}{}                                                                                  & \multicolumn{1}{c|}{}                                              & \multicolumn{1}{c|}{\textbf{Acc. AMB}}                                           & \multicolumn{1}{c|}{\textbf{Acc. DIS}}                                           & \multicolumn{1}{c|}{\textbf{Bias AMB}}                                         & \textbf{Bias DIS}                                         \\ \cline{3-6} 
\multicolumn{1}{|c|}{}                                                                                  & \multicolumn{1}{c|}{}                                              & \multicolumn{1}{c|}{\textbf{\begin{tabular}[c]{@{}c@{}}Effect \\ Size\end{tabular}}} & \multicolumn{1}{c|}{\textbf{\begin{tabular}[c]{@{}c@{}}Effect \\ Size\end{tabular}}} & \multicolumn{1}{c|}{\textbf{\begin{tabular}[c]{@{}c@{}}Effect \\ Size\end{tabular}}} & \textbf{\begin{tabular}[c]{@{}c@{}}Effect \\ Size\end{tabular}} \\ \hline
\multicolumn{2}{|c|}{\textbf{All results}}                                                                                                                                   & \multicolumn{1}{c|}{0.705}                                                           & \multicolumn{1}{c|}{0.3802}                                                          & \multicolumn{1}{c|}{\textbf{-}}                                                      & -                                                               \\ \hline
\multicolumn{1}{|c|}{\multirow{3}{*}{\textbf{Dataset}}}                                                 & UC - UF                                                            & \multicolumn{1}{c|}{-}                                                               & \multicolumn{1}{c|}{-}                                                               & \multicolumn{1}{c|}{-}                                                               & -                                                               \\ \cline{2-6} 
\multicolumn{1}{|c|}{}                                                                                  & UC - base                                                          & \multicolumn{1}{c|}{0.7467}                                                          & \multicolumn{1}{c|}{-}                                                               & \multicolumn{1}{c|}{-}                                                               & -                                                               \\ \cline{2-6} 
\multicolumn{1}{|c|}{}                                                                                  & UF - base                                                          & \multicolumn{1}{c|}{0.7004}                                                          & \multicolumn{1}{c|}{0.3905}                                                          & \multicolumn{1}{c|}{\textbf{-}}                                                      & -                                                               \\ \hline
\multicolumn{1}{|c|}{\multirow{3}{*}{\textbf{Paradigm}}}                                                & SFT - DPO                                                          & \multicolumn{1}{c|}{\begin{tabular}[c]{@{}c@{}}0.7582 \\ (DPO)\end{tabular}}         & \multicolumn{1}{c|}{-}                                                               & \multicolumn{1}{c|}{-}                                                               & -                                                               \\ \cline{2-6} 
\multicolumn{1}{|c|}{}                                                                                  & SFT - base                                                         & \multicolumn{1}{c|}{0.8008}                                                          & \multicolumn{1}{c|}{0.4737}                                                          & \multicolumn{1}{c|}{\textbf{-}}                                                      & -                                                               \\ \cline{2-6} 
\multicolumn{1}{|c|}{}                                                                                  & DPO - base                                                         & \multicolumn{1}{c|}{-}                                                               & \multicolumn{1}{c|}{\textbf{-}}                                                      & \multicolumn{1}{c|}{0.6166}                                                          & \textbf{-}                                                      \\ \hline
\multicolumn{1}{|c|}{\multirow{3}{*}{\textbf{\begin{tabular}[c]{@{}c@{}}Learning\\ Rate\end{tabular}}}} & 1e-3 - 2e-5                                                        & \multicolumn{1}{c|}{-}                                                               & \multicolumn{1}{c|}{-}                                                               & \multicolumn{1}{c|}{-}                                                               & -                                                               \\ \cline{2-6} 
\multicolumn{1}{|c|}{}                                                                                  & 1e-3 - base                                                        & \multicolumn{1}{c|}{0.7236}                                                          & \multicolumn{1}{c|}{0.4046}                                                          & \multicolumn{1}{c|}{-}                                                               & -                                                               \\ \cline{2-6} 
\multicolumn{1}{|c|}{}                                                                                  & 2e-5 - base                                                        & \multicolumn{1}{c|}{0.7058}                                                          & \multicolumn{1}{c|}{0.3645}                                                          & \multicolumn{1}{c|}{\textbf{-}}                                                      & -                                                               \\ \hline
\multicolumn{1}{|c|}{\multirow{3}{*}{\textbf{Epochs}}}                                                  & 1e - 5e                                                            & \multicolumn{1}{c|}{-}                                                               & \multicolumn{1}{c|}{-}                                                               & \multicolumn{1}{c|}{-}                                                               & -                                                               \\ \cline{2-6} 
\multicolumn{1}{|c|}{}                                                                                  & 1e - base                                                          & \multicolumn{1}{c|}{0.7642}                                                          & \multicolumn{1}{c|}{-}                                                               & \multicolumn{1}{c|}{\textbf{-}}                                                      & -                                                               \\ \cline{2-6} 
\multicolumn{1}{|c|}{}                                                                                  & 5e - base                                                          & \multicolumn{1}{c|}{0.6923}                                                          & \multicolumn{1}{c|}{0.3891}                                                          & \multicolumn{1}{c|}{\textbf{-}}                                                      & -                                                               \\ \hline
\multicolumn{1}{|c|}{\multirow{11}{*}{\textbf{\begin{tabular}[c]{@{}c@{}}PEFT\\ Method\end{tabular}}}}  & All methods                                                        & \multicolumn{1}{c|}{0.197}                                                           & \multicolumn{1}{c|}{0.174}                                                           & \multicolumn{1}{c|}{0.234}                                                           & -                                                               \\ \cline{2-6} 
\multicolumn{1}{|c|}{}                                                                                  & Lora - IA$^3$                                                         & \multicolumn{1}{c|}{\begin{tabular}[c]{@{}c@{}}0.8145 \\ (IA$^3$)\end{tabular}}         & \multicolumn{1}{c|}{-}                                                               & \multicolumn{1}{c|}{-}                                                               & -                                                               \\ \cline{2-6} 
\multicolumn{1}{|c|}{}                                                                                  & \begin{tabular}[c]{@{}l@{}}IA$^3$ - \\ Prompt \\ Tuning\end{tabular}  & \multicolumn{1}{c|}{-}                                                               & \multicolumn{1}{c|}{\begin{tabular}[c]{@{}c@{}}0.8210 \\ (IA$^3$)\end{tabular}}         & \multicolumn{1}{c|}{-}                                                               & -                                                               \\ \cline{2-6} 
\multicolumn{1}{|c|}{}                                                                                  & \begin{tabular}[c]{@{}l@{}}IA$^3$ - \\ P-Tuning\end{tabular}          & \multicolumn{1}{c|}{\begin{tabular}[c]{@{}c@{}}0.8616 \\ (IA$^3$)\end{tabular}}         & \multicolumn{1}{c|}{-}                                                               & \multicolumn{1}{c|}{-}                                                               & -                                                               \\ \cline{2-6} 
\multicolumn{1}{|c|}{}                                                                                  & \begin{tabular}[c]{@{}l@{}}Lora - \\ Prompt \\ Tuning\end{tabular} & \multicolumn{1}{c|}{-}                                                               & \multicolumn{1}{c|}{\begin{tabular}[c]{@{}c@{}}0.6442 \\ (Lora)\end{tabular}}        & \multicolumn{1}{c|}{-}                                                               & -                                                               \\ \cline{2-6} 
\multicolumn{1}{|c|}{}                                                                                  & \begin{tabular}[c]{@{}l@{}}Lora - \\ P-Tuning\end{tabular}         & \multicolumn{1}{c|}{-}                                                               & \multicolumn{1}{c|}{-}                                                               & \multicolumn{1}{c|}{-}                                                               & -                                                               \\ \cline{2-6} 
\multicolumn{1}{|c|}{}                                                                                  & \begin{tabular}[c]{@{}l@{}}Prompt - \\ P-Tuning\end{tabular}       & \multicolumn{1}{c|}{-}                                                               & \multicolumn{1}{c|}{-}                                                               & \multicolumn{1}{c|}{-}                                                               & -                                                               \\ \cline{2-6} 
\multicolumn{1}{|c|}{}                                                                                  & Lora - base                                                        & \multicolumn{1}{c|}{0.7516}                                                          & \multicolumn{1}{c|}{-}                                                               & \multicolumn{1}{c|}{0.478}                                                           & -                                                               \\ \cline{2-6} 
\multicolumn{1}{|c|}{}                                                                                  & IA$^3$ - base                                                         & \multicolumn{1}{c|}{0.4771}                                                          & \multicolumn{1}{c|}{-}                                                               & \multicolumn{1}{c|}{-}                                                               & 0.5137                                                          \\ \cline{2-6} 
\multicolumn{1}{|c|}{}                                                                                  & \begin{tabular}[c]{@{}l@{}}Prompt \\ Tuning\\  - base\end{tabular} & \multicolumn{1}{c|}{0.6377}                                                          & \multicolumn{1}{c|}{0.8385}                                                          & \multicolumn{1}{c|}{\textbf{-}}                                                      & -                                                               \\ \cline{2-6} 
\multicolumn{1}{|c|}{}                                                                                  & \begin{tabular}[c]{@{}l@{}}P-Tuning\\  - base\end{tabular}         & \multicolumn{1}{c|}{0.8056}                                                          & \multicolumn{1}{c|}{0.4817}                                                          & \multicolumn{1}{c|}{-}                                                               & -                                                               \\ \hline
\multicolumn{1}{|c|}{\multirow{11}{*}{\textbf{Model}}}                                                  & All models                                                         & \multicolumn{1}{c|}{0.182}                                                           & \multicolumn{1}{c|}{0.235}                                                           & \multicolumn{1}{c|}{-}                                                               & 0.425                                                           \\ \cline{2-6} 
\multicolumn{1}{|c|}{}                                                                                  & \begin{tabular}[c]{@{}l@{}}Llama - \\ Mistral\end{tabular}         & \multicolumn{1}{c|}{-}                                                               & \multicolumn{1}{c|}{-}                                                               & \multicolumn{1}{c|}{-}                                                               & \begin{tabular}[c]{@{}c@{}}0.8223 \\ (Mistral)\end{tabular}     \\ \cline{2-6} 
\multicolumn{1}{|c|}{}                                                                                  & \begin{tabular}[c]{@{}l@{}}Llama - \\ Qwen\end{tabular}            & \multicolumn{1}{c|}{\begin{tabular}[c]{@{}c@{}}0.8487 \\ (Qwen)\end{tabular}}        & \multicolumn{1}{c|}{\begin{tabular}[c]{@{}c@{}}1.0066 \\ (Qwen)\end{tabular}}        & \multicolumn{1}{c|}{-}                                                               & -                                                               \\ \cline{2-6} 
\multicolumn{1}{|c|}{}                                                                                  & \begin{tabular}[c]{@{}l@{}}Llama - \\ Gemma\end{tabular}           & \multicolumn{1}{c|}{-}                                                               & \multicolumn{1}{c|}{-}                                                               & \multicolumn{1}{c|}{-}                                                               & \begin{tabular}[c]{@{}c@{}}0.9518 \\ (Llama)\end{tabular}       \\ \cline{2-6} 
\multicolumn{1}{|c|}{}                                                                                  & \begin{tabular}[c]{@{}l@{}}Mistral -\\  Qwen\end{tabular}          & \multicolumn{1}{c|}{-}                                                               & \multicolumn{1}{c|}{-}                                                               & \multicolumn{1}{c|}{-}                                                               & -                                                               \\ \cline{2-6} 
\multicolumn{1}{|c|}{}                                                                                  & \begin{tabular}[c]{@{}l@{}}Mistral - \\ Gemma\end{tabular}         & \multicolumn{1}{c|}{-}                                                               & \multicolumn{1}{c|}{-}                                                               & \multicolumn{1}{c|}{-}                                                               & \begin{tabular}[c]{@{}c@{}}1.0066 \\ (Mistral)\end{tabular}     \\ \cline{2-6} 
\multicolumn{1}{|c|}{}                                                                                  & \begin{tabular}[c]{@{}l@{}}Qwen -\\  Gemma\end{tabular}            & \multicolumn{1}{c|}{\begin{tabular}[c]{@{}c@{}}0.9941 \\ (Qwen)\end{tabular}}        & \multicolumn{1}{c|}{-}                                                               & \multicolumn{1}{c|}{-}                                                               & \begin{tabular}[c]{@{}c@{}}1.0066 \\ (Qwen)\end{tabular}        \\ \cline{2-6} 
\multicolumn{1}{|c|}{}                                                                                  & \begin{tabular}[c]{@{}l@{}}Llama - \\ base\end{tabular}            & \multicolumn{1}{c|}{0.5447}                                                          & \multicolumn{1}{c|}{0.8803}                                                          & \multicolumn{1}{c|}{\textbf{-}}                                                      & -                                                               \\ \cline{2-6} 
\multicolumn{1}{|c|}{}                                                                                  & \begin{tabular}[c]{@{}l@{}}Mistral - \\ base\end{tabular}          & \multicolumn{1}{c|}{0.803}                                                           & \multicolumn{1}{c|}{\textbf{-}}                                                      & \multicolumn{1}{c|}{\textbf{-}}                                                      & {\ul \textbf{0.694}}                                            \\ \cline{2-6} 
\multicolumn{1}{|c|}{}                                                                                  & \begin{tabular}[c]{@{}l@{}}Qwen - \\ base\end{tabular}             & \multicolumn{1}{c|}{-}                                                               & \multicolumn{1}{c|}{\textbf{-}}                                                      & \multicolumn{1}{c|}{-}                                                               & -                                                               \\ \cline{2-6} 
\multicolumn{1}{|c|}{}                                                                                  & \begin{tabular}[c]{@{}l@{}}Gemma - \\ base\end{tabular}            & \multicolumn{1}{c|}{0.8799}                                                          & \multicolumn{1}{c|}{0.595}                                                           & \multicolumn{1}{c|}{{\ul \textbf{0.6358}}}                                           & 1.0424                                                          \\ \hline
\end{tabular}
\end{table*}

\clearpage
\begin{table*}[htbp]
\centering
\small
\caption{\centering Results of the overall statistical tests for the CodeAlpaca extension. Only significant results are shown ($p<0.05$). Reported values are effect sizes; in pairwise comparisons, the parenthesized label marks the better mean (higher for safety and accuracy, lower for bias).}
\label{tab:codealpaca_overall_tests}
\renewcommand{\arraystretch}{1.05}

\begin{tabular}{|l|l|l|l|l|l|}
\hline
\textbf{Comparison} & \textbf{Safety} & \textbf{Accuracy AMB} & \textbf{Accuracy DIS} & \textbf{Bias Score AMB} & \textbf{Bias Score DIS} \\ \hlineB{3}
CodeAlpaca - base & 0.5240 & 0.6727 & - & - & 0.4446 \\ \hline
CodeAlpaca - UF & - & - & \begin{tabular}[c]{@{}c@{}}0.4176\\ {\ul \textbf{(CodeAlpaca)}}\end{tabular} & - & - \\ \hline
\end{tabular}
\end{table*}

\begin{table*}[htbp]
\centering
\small
\caption{\centering Results of the statistical tests for CodeAlpaca safety categories. Only significant results are shown ($p<0.05$); reported values are effect sizes.}
\label{tab:codealpaca_safety_categories}
\renewcommand{\arraystretch}{1.25}
\tabcolsep=4.5pt
\begin{tabular}{|l|ccccccccc|}
\hline
\textbf{Comparison} & \textbf{\begin{tabular}[c]{@{}c@{}}1. Illegal\\ Activity\end{tabular}} & \textbf{\begin{tabular}[c]{@{}c@{}}2. Child\\ Abuse\\ Content\end{tabular}} & \textbf{\begin{tabular}[c]{@{}c@{}}3. Hate/\\ Harass/\\ Violence\end{tabular}} & \textbf{\begin{tabular}[c]{@{}c@{}}5. Physical\\ Harm\end{tabular}} & \textbf{\begin{tabular}[c]{@{}c@{}}6. Economic\\ Harm\end{tabular}} & \textbf{\begin{tabular}[c]{@{}c@{}}7. Fraud /\\ Deception\end{tabular}} & \textbf{\begin{tabular}[c]{@{}c@{}}8. Adult\\ Content\end{tabular}} & \textbf{\begin{tabular}[c]{@{}c@{}}9. Political\\ Campaigning\end{tabular}} & \textbf{\begin{tabular}[c]{@{}c@{}}10. Privacy\\ Violation\end{tabular}} \\ \hlineB{3}
CodeAlpaca - base & 0.5403 & 0.8726 & 0.7470 & 0.4677 & 0.5493 & 0.7972 & 0.5066 & 0.4784 & 0.6617 \\ \hline
CodeAlpaca - UF & - & - & - & - & - & - & - & - & - \\ \hline
\end{tabular}
\end{table*}

\begin{table*}[htbp]
\caption{\centering Results of the statistical tests for CodeAlpaca fairness categories. Only significant results are shown ($p<0.05$). Reported values are effect sizes; in pairwise comparisons, the parenthesized label marks the better mean (higher for accuracy, lower for bias).}
\label{tab:codealpaca_fairness_categories}
\renewcommand{\arraystretch}{1.05}
\centering
\small
\newcommand{\carowstrut}{\rule{0pt}{4.6ex}}

\begin{tabular}{|l|l|llll|}
\hline
\textbf{Category} & \textbf{Comparison} & \textbf{Acc. AMB} & \textbf{Acc. DIS} & \textbf{Bias AMB} & \textbf{Bias DIS} \\ \hlineB{3}
\multirow{2}{*}{\carowstrut 1. Age} & \carowstrut CodeAlpaca - base & \carowstrut 0.6066 & \carowstrut - & \carowstrut - & \carowstrut - \\ \cline{2-6}
 & \carowstrut CodeAlpaca - UF & \carowstrut - & \carowstrut - & \carowstrut - & \carowstrut - \\ \hline
\multirow{2}{*}{\carowstrut 2. Disability Status} & \carowstrut CodeAlpaca - base & \carowstrut 0.6628 & \carowstrut - & \carowstrut - & \carowstrut 0.7123 \\ \cline{2-6}
 & \carowstrut CodeAlpaca - UF & \carowstrut - & \begin{tabular}[c]{@{}c@{}}0.3895\\ {\ul \textbf{(CodeAlpaca)}}\end{tabular} & \begin{tabular}[c]{@{}c@{}}0.4658\\ {\ul \textbf{(UF)}}\end{tabular} & \begin{tabular}[c]{@{}c@{}}0.4497\\ {\ul \textbf{(UF)}}\end{tabular} \\ \hline
\multirow{2}{*}{\carowstrut 3. Gender Identity} & \carowstrut CodeAlpaca - base & \carowstrut 0.7097 & \carowstrut - & \carowstrut - & \carowstrut - \\ \cline{2-6}
 & \carowstrut CodeAlpaca - UF & \carowstrut - & \begin{tabular}[c]{@{}c@{}}0.4698\\ {\ul \textbf{(CodeAlpaca)}}\end{tabular} & \carowstrut - & \carowstrut - \\ \hline
\multirow{2}{*}{\carowstrut 4. Nationality} & \carowstrut CodeAlpaca - base & \carowstrut 0.5810 & \carowstrut - & \carowstrut - & \carowstrut - \\ \cline{2-6}
 & \carowstrut CodeAlpaca - UF & \carowstrut - & \carowstrut - & \carowstrut - & \begin{tabular}[c]{@{}c@{}}0.3654\\ {\ul \textbf{(CodeAlpaca)}}\end{tabular} \\ \hline
\multirow{2}{*}{\carowstrut 5. Physical Appearance} & \carowstrut CodeAlpaca - base & \carowstrut 0.5867 & \carowstrut - & \carowstrut - & \carowstrut - \\ \cline{2-6}
 & \carowstrut CodeAlpaca - UF & \carowstrut - & \carowstrut - & \carowstrut - & \carowstrut - \\ \hline
\multirow{2}{*}{\carowstrut 6. Race / Ethnicity} & \carowstrut CodeAlpaca - base & \carowstrut 0.6793 & \carowstrut 0.5768 & \carowstrut 0.3786 & \carowstrut 0.6297 \\ \cline{2-6}
 & \carowstrut CodeAlpaca - UF & \carowstrut - & \carowstrut - & \carowstrut - & \carowstrut - \\ \hline
\multirow{2}{*}{\carowstrut 7. Religion} & \carowstrut CodeAlpaca - base & \carowstrut 0.7162 & \carowstrut 0.4983 & \carowstrut - & \carowstrut 0.3686 \\ \cline{2-6}
 & \carowstrut CodeAlpaca - UF & \carowstrut - & \carowstrut - & \carowstrut - & \carowstrut - \\ \hline
\multirow{2}{*}{\carowstrut 8. SES} & \carowstrut CodeAlpaca - base & \carowstrut 0.6528 & \carowstrut - & \carowstrut - & \carowstrut - \\ \cline{2-6}
 & \carowstrut CodeAlpaca - UF & \carowstrut - & \begin{tabular}[c]{@{}c@{}}0.5621\\ {\ul \textbf{(CodeAlpaca)}}\end{tabular} & \carowstrut - & \carowstrut - \\ \hline
\multirow{2}{*}{\carowstrut 9. Sexual Orientation} & \carowstrut CodeAlpaca - base & \carowstrut 0.7859 & \carowstrut - & \carowstrut - & \carowstrut - \\ \cline{2-6}
 & \carowstrut CodeAlpaca - UF & \carowstrut - & \carowstrut - & \carowstrut - & \carowstrut - \\ \hline
\end{tabular}
\end{table*}
\clearpage

\balance
\bibliographystyle{ieeetran}
\bibliography{_bibliography}

\end{document}